\documentclass[journal]{IEEEtran}
\usepackage[utf8x]{inputenc}
\usepackage[noadjust]{cite}
\usepackage{graphicx}
\usepackage{epstopdf}
\usepackage{amsfonts}
\usepackage[cmex10]{amsmath}
\ifCLASSOPTIONcompsoc
  \usepackage[caption=false,font=normalsize,labelfont=sf,textfont=sf]{subfig}
\else
  \usepackage[caption=false,font=footnotesize]{subfig}
\fi
\usepackage{tikz}
\usepackage{breqn}
\usepackage{placeins}
\usepackage{float}
\usetikzlibrary{calc,arrows,shapes,backgrounds,petri}
\newcommand{\vecnorm}[1]{\left\|#1\right\|}

\DeclareMathOperator*{\divergence}{div}

\DeclareMathOperator*{\prob}{P}
\DeclareMathOperator*{\expected}{E}
\DeclareMathOperator*{\variance}{Var}
\DeclareMathOperator*{\bernoulli}{Bern}
\DeclareMathOperator*{\binomial}{Bin}
\newtheorem{lemma}{Lemma}
\newtheorem{theorem}{Theorem}
\newtheorem{definition}{Definition}

\begin{document}

\title{Theoretical Analysis of Active Contours on Graphs}

\author{\IEEEauthorblockN{Christos~Sakaridis\IEEEauthorrefmark{1},
        Kimon~Drakopoulos\IEEEauthorrefmark{2},
        and~Petros~Maragos\IEEEauthorrefmark{3}}
        
        \IEEEauthorblockA{\IEEEauthorrefmark{1}ETH~Zürich,
        \IEEEauthorrefmark{2}University~of~Southern~California,
        \IEEEauthorrefmark{3}National~Technical~University~of~Athens}
}

\maketitle

\begin{abstract}
Active contour models based on partial differential equations have proved successful in image segmentation, yet the study of their geometric formulation on arbitrary geometric graphs is still at an early stage. In this paper, we introduce geometric approximations of gradient and curvature, which are used in the geodesic active contour model. We prove convergence in probability of our gradient approximation to the true gradient value and derive an asymptotic upper bound for the error of this approximation for the class of random geometric graphs. Two different approaches for the approximation of curvature are presented and both are also proved to converge in probability in the case of random geometric graphs. We propose neighborhood-based filtering on graphs to improve the accuracy of the aforementioned approximations and define two variants of Gaussian smoothing on graphs which include normalization in order to adapt to graph non-uniformities. The performance of our active contour framework on graphs is demonstrated in the segmentation of regular images and geographical data defined on arbitrary graphs.
\end{abstract}

\begin{IEEEkeywords}
Geodesic active contours, graph segmentation, random geometric graphs, image segmentation, object detection.
\end{IEEEkeywords}

\IEEEpeerreviewmaketitle

\section{Introduction}
\IEEEPARstart{E}{volution} of curves via active contour models has been applied extensively in computer vision for image segmentation and object detection. In the classical image setting which involves a regular grid of pixels, the discretization of PDEs governing the motion of active contours is well-established and ensures proper convergence of the contour to object boundaries. Recently, active contours have been extended to handle more general input in the form of graphs whose vertices are arbitrarily distributed in a two-dimensional Euclidean space. This arbitrary spatial configuration poses a significant challenge to the discrete approximation of continuous operators that are used in active contours. Applications of segmentation of such graphs span not only image processing, but also geographical information systems and generally any field where data can assume the form of a set of pointwise samples of a real-valued function.

Our work focuses mainly on the theoretical study of fundamental geometric terms in active contours, primarily \emph{gradient} and \emph{curvature}, and the introduction of novel, neighborhood-based approximations of them on arbitrary graphs, which improve upon previous approaches. We analyze the exactness of these approximations and prove convergence to the true values in the limit of large-scale input for the class of \emph{random geometric graphs}. Additionally, we derive an asymptotic bound for the error of our gradient approximation with respect to the number of vertices of the graph. Another important contribution is the usage of neighborhood-based smoothing filtering on graphs as an algorithmic heuristic to reduce the error of our approximations for smooth inputs. Last, we propose normalized versions of Gaussian filtering on graphs (which is essential for initialization of active contour schemes), suited to handle non-uniform vertex distributions.

The paper is structured as follows. Section~\ref{sec:background} reviews previous work on active contours, graph-based morphology and PDE-based methods on graphs and provides the necessary background on active contour models. In Section~\ref{sec:gradapprox} we introduce the basic quantities of our framework and present our geometric approximation of gradient on graphs. We provide conditions for convergence in probability of our approximation in the case of random geometric graphs and analyze the asymptotic behavior of approximation error, which enables an advised selection of parameters for graph construction. In Section~\ref{sec:curvapprox} we give two methods to approximate curvature on graphs, both of which rely on gradient approximation, and state theorems about their convergence in probability for random geometric graphs. Section~\ref{sec:smoothing} is dedicated to defining neighborhood-based smoothing filters on graphs, introducing normalized Gaussian filtering and Gaussian derivative filtering on graphs, and demonstrating their use in smoothing synthetic gradient, curvature or image functions. In Section~\ref{sec:results} we apply the geodesic active contour algorithm on graphs constructed synthetically, defined from regular images, or containing geographical data, and compare different methods to create the set of vertices and/or edges of these graphs.

\section{Background and Related Work} \label{sec:background}
Active contour models for curve evolution towards image edges originate from ``snakes'' \cite{KWT88}. These early approaches could not in general handle topological changes of the contour, for instance splitting into two disjoint parts to detect the boundaries of two distinct objects. PDE-based methods using level sets were proposed as an alternative in \cite{CCCD93,CKS97}, where the geometric active contour model was initially introduced and subsequently complemented to establish the geodesic active contour (GAC) framework. The former model involves two forces that govern curve motion: a balloon force that expands or shrinks it, and a curvature-dependent force that maintains its smoothness. The latter model adds an extra spring force that attracts the contour towards salient image edges. Both methods embed the active contour as a \emph{level set} of the function \(u\) involved in the PDE that models curve evolution, allowing the use of a numerical scheme of the type proposed in \cite{OsSe88}.

Graphs have long been connected to image processing, in part through their study in terms of mathematical morphology. The application of morphological transforms on neighborhood graphs was established in \cite{Vinc89}, while a wide variety of graph structures, algorithms for their construction and early applications in computer vision were surveyed in \cite{JaTo92}. The notion of structuring element in classical morphology was extended to graphs in \cite{HNTV92}, where the proposed structuring graph enables a generalization of neighborhood functions on a graph beyond the one induced by its set of edges. Morphological operators on graphs have been studied further in \cite{CNS09}, where the lattice of the subgraphs of a graph is considered in order to define filters that treat the graph as a whole.

Recently, several works, including \cite{TEL11,LET12,LEL14,DrMa12}, have focused on the construction of PDE-based rather than algebraically defined morphological operators on graphs, which are then used to define active contour models on graphs. All these works are based on the definition of a gradient operator on graphs, however, \cite{TEL11,LET12,LEL14} work on weighted graphs and define a discrete gradient vector on vertices whose dimensionality is the same as the cardinality of each vertex's neighborhood, whereas \cite{DrMa12} considers unweighted graphs and approximates the continuous gradient at each vertex. In \cite{LET12,LEL14}, active contours are formulated in a variational framework, while in \cite{DrMa12}, the gradient approximation is leveraged to translate the aforementioned geodesic active contour segmentation framework to 2D graphs with arbitrary structure and vertex configuration. We follow the latter path and carefully treat the geometric quantities involved in the active contour model, such as gradient and curvature. Our aim is to establish graph-based approximations of these quantities that guarantee proper convergence of the contour to object boundaries and that are exact in the limit of large, dense geometric graphs. In particular, to the best of our knowledge, the asymptotic upper bound for the error of our gradient approximation for random geometric graphs is the first of its kind.

A different class of approaches to graph segmentation which has gained a lot of interest in the image processing community is based on graph cuts. These approaches, in contrast to ours, usually operate on a regular image grid and define weighted edges between image pixels based on certain cues like spatial or appearance proximity, in order to find a cut of minimal cost for the resulting weighted graph. The cost of a cut is normalized in \cite{ShMa00} so that balanced partitions are preferred. Approximate solutions to multi-label problems are proposed in \cite{BVZ01}, guaranteeing constant-factor optimality. A link between geodesic active contours and graph cuts is established in \cite{BoKo03}, where the graph is constructed so that the cost of the cut corresponds to the contour's length under the induced anisotropic metric, and this link is extended to the arbitrary graph setting in \cite{DrMa12}. Efficient algorithms for watershed-like segmentation that are formulated as graph cuts are introduced in \cite{CBNC09,CBNC10}. The power watershed framework of \cite{CGNT11} unites and generalizes several graph-based optimization methods for image segmentation by expressing their energies in a common, parametric form.

\section{Gradient Approximation on Graphs} \label{sec:gradapprox}
The first term of the active contour evolution model that needs to be approximated is the gradient of the bivariate embedding function. We thus develop a general method for calculating the gradient of a real-valued, bivariate function that is implicitly defined on a continuous domain, although its values are known only at a sparse, finite set of points, which coincide with the vertices of the graph.

\subsection{Main Idea, Notation and Definitions}
Compared to the proposals of Drakopoulos and Maragos \cite{DrMa12} for gradient approximation, we attempt to incorporate our knowledge about the local spatial configuration of vertices in the approximation. More specifically, we introduce the concept of the \emph{angle} around a vertex which is ``occupied'' by each of its neighbors and use this concept directly in our novel geometric gradient approximation. Our motivation for this approach comes from the following lemma in bivariate calculus.

\begin{lemma} \label{lem:grad}
The gradient of a differentiable function \(u:\mathbb{R}^2\rightarrow\mathbb{R}\) at point \(\mathbf{x}\) is
\begin{equation} \label{eq:grad}
\nabla{u(\mathbf{x})} = \frac{\displaystyle\int_0^{2\pi} D_{\phi}u(\mathbf{x})\,\mathbf{e}_\phi\,d\phi}{\pi},
\end{equation}
where \(\mathbf{e}_\phi\) is the unit vector in direction \(\phi\) and \(D_{\phi}u(\mathbf{x})\) is the directional derivative of \(u\) at \(\mathbf{x}\) in this direction, defined by
\begin{equation*}
D_{\phi}u(\mathbf{x}) = \lim_{h \to 0} \frac{u(\mathbf{x}+h\mathbf{e}_\phi)-u(\mathbf{x})}{h}.
\end{equation*}
\end{lemma}

Based on Lemma \ref{lem:grad}, the goal of this section is to approximate the gradient at a vertex of the graph by substituting the integral
\begin{equation} \label{eq:I}
\mathcal{I} = \int_0^{2\pi} D_{\phi}u(\mathbf{x})\,\mathbf{e}_\phi\,d\phi
\end{equation}
with a sum over all the neighbors of the vertex. To this end, we start by introducing several key concepts.

\begin{figure}
  \centering
  \begin{tikzpicture}

    \coordinate (Origin)   at (0,0);
    \coordinate (XAxisMin) at (-3,0);
    \coordinate (XAxisMax) at (3.5,0);
    \coordinate (BottomLeft) at (-3,-0.5);
    \coordinate (TopRight) at (3.5,4);
    
    \clip (BottomLeft) rectangle (TopRight); 
    
    \draw [thin, gray,-latex] (XAxisMin) -- (XAxisMax); 
    \node [draw,circle,inner sep=2pt,fill,label=above left:$w_{i-1}$] (wimin1) at (3,2) {};
    \node [draw,circle,inner sep=2pt,fill,label=above:$w_i$] (wi) at (-0.5,2.5) {};
    \node [draw,circle,inner sep=2pt,fill,label=above left:$w_{i+1}$] (wiplus1) at (-2,3) {};
    \node[draw,circle,inner sep=2pt,fill,label=below left:$v$] (v) at (0,0) {}
        edge (wi)
        edge (wimin1)
        edge (wiplus1)
        ;
    \draw[dash pattern=on5pt off3pt] (Origin) -- (67.5:4) ;
    \draw[dash pattern=on5pt off3pt] (Origin) -- (112.5:4) ;
    \draw[thick] (0.5,1.2071) arc (67.5:112.5:1.3066);
    \draw[thick] (1.5,0) arc (0:67.5:1.5);
    \draw[thick] (2.35,0) arc (0:101.31:2.35);
    \draw (0.25,0.6036) arc (67.5:101.31:0.6533);
    \draw (0.3,0.7243) arc (67.5:101.31:0.7840);
    \draw (0.7072,0.4715) arc (33.69:67.5:0.85);
    \draw (0.8321,0.5547) arc (33.69:67.5:1);
    \draw[dash pattern=on5pt off3pt] (-0.4118,2.0590) arc (101.31:112.5:2.1);
    \draw[dash pattern=on5pt off3pt] (-0.4413,2.2065) arc (101.31:112.5:2.25);
    \draw[dash pattern=on5pt off3pt] (-1.1481,2.7717) arc (112.5:123.69:3);
    \draw[dash pattern=on5pt off3pt] (-1.2055,2.9103) arc (112.5:123.69:3.15);
    \node[font=\scriptsize] at (93:1.7) {$\Delta\phi(w_i)$};
    \node[font=\scriptsize] at (32:1.85) {$\omega(w_i)$};
    \node[font=\scriptsize] at (55:2.65) {$\phi(w_i)$};
    
  \end{tikzpicture}
  \caption{Angles \(\phi(w_i)\), \(\Delta\phi(w_i)\) and \(\omega(w_i)\).}
  \label{fig:Angles}
\end{figure}
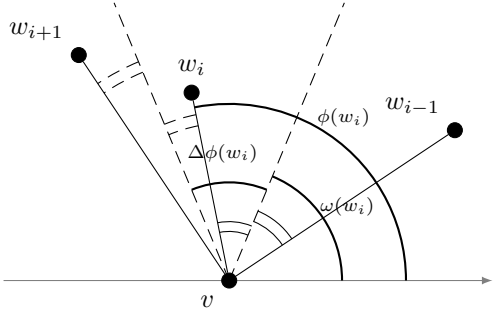

The Euclidean distance between vertices \(v\) and \(w\) of a graph \(\mathcal{G}\) is denoted by \(d(v,w)\) and the unit vector in the direction of the edge \(vw\) starting at \(v\) is denoted by \(\mathbf{e}_{vw}\). We define \(\phi(w) \in [0,2\pi)\) as the angle between the vector \(\mathbf{e}_{vw}\) and the horizontal axis, as in Fig. \ref{fig:Angles}. A vertex \(v\) will be alternatively denoted by \(\mathbf{v}\) to declare its position vector. Moreover, we denote by \(\mathcal{N}(v)\) the set of neighbors of \(v\) in \(\mathcal{G}\), with cardinality \(N(v)\). For the sake of brevity in notation, this cardinality will be written simply as \(N\). We write \(\mathcal{N}(v) = \left\{w_1,\,w_2,\,...,\,w_N\right\}\) so that the angles \(\phi(w_i)\) are in ascending order. Based on this ordering, we define the angle around \(v\) ``occupied'' by \(w_i\), which we call \emph{neighbor angle}, as
\begin{equation} \label{eq:Deltaphi}
\Delta\phi(w_i) = \left\{
{\def\arraystretch{2.2}
\begin{array}{ll}
\displaystyle\frac{\phi\left(w_{i+1}\right)-\left(\phi\left(w_N\right)-2\pi\right)}{2} & \text{if }i=1,\\
\displaystyle\frac{\phi\left(w_1\right)+2\pi-\phi\left(w_{i-1}\right)}{2} & \text{if }i=N,\\
\displaystyle\frac{\phi\left(w_{i+1}\right)-\phi\left(w_{i-1}\right)}{2} & \text{otherwise.}
\end{array}
}
\right.
\end{equation}
In a similar fashion, we define the angle corresponding to the bisector between two consecutive neighbors as
\begin{equation} \label{eq:omega}
\omega(w_i) = \left\{
{\def\arraystretch{2.2}
\begin{array}{ll}
\displaystyle\frac{\phi\left(w_i\right)+\phi\left(w_N\right)-2\pi}{2} & \text{if }i=1,\\
\displaystyle\frac{\phi\left(w_i\right)+\phi\left(w_{i-1}\right)}{2} & \text{otherwise.}
\end{array}
}
\right.
\end{equation}

A visual representation of the neighbor angle is provided in Fig. \ref{fig:Angles}.

Using the above notation, we propose the following  formula as the geometric gradient approximation at \(v\):
\begin{equation} \label{eq:gradgeomapprox}
\nabla{u(v)} \approx \frac{\displaystyle\sum_{i=1}^N \frac{u(w_i)-u(v)}{d(v,w_i)}\,\mathbf{e}_{vw_i}\,\Delta\phi(w_i)}{\pi}.
\end{equation}
The directional derivative term in \eqref{eq:grad} is approximated by the difference quotient of the function along each edge. On the other hand, the angle differential is handled through the neighbor angles, which effectively constitute a Voronoi tessellation of the circle around \(v\), created from its neighbors. The reasoning behind this approach is to use information about the change of \(u\) along each particular direction that comes from the neighbor which is \emph{closest} to this direction.

If we chose not to take the neighbor angles into account, we would place equal importance on all the neighbors of the vertex and we would return to an approximation similar to the weighted sum that was introduced in \cite{DrMa12}:
\begin{equation} \label{eq:gradweighsumapprox}
\nabla{u(v)} \approx \frac{\displaystyle\sum_{i=1}^N \frac{u(w_i)-u(v)}{d(v,w_i)}\,\mathbf{e}_{vw_i}}{N}.
\end{equation}

\subsection{Convergence for Random Geometric Graphs} \label{subsec:gradconvrgg}
In the following, we will mainly focus on a certain type of graphs, \emph{random geometric graphs} defined below, to study the proposed gradient approximation theoretically.

\begin{definition} \label{def:rgg}
A random geometric graph (RGG) \(\mathcal{G}(n,\rho(n))\) is comprised of a set \(\mathcal{V}\) of vertices and a set \(\mathcal{E}\) of edges. The set \(\mathcal{V}\) consists of \(n\) points distributed uniformly and independently in a bounded region \(D \subset \mathbb{R}^2\). The set \(\mathcal{E}\) of edges is defined through the radius \(\rho(n)\) of the graph: an edge connects two vertices \(v\) and \(w\) if and only if their distance is at most \(\rho(n)\), i.e. \(d(v,w) \leq \rho(n)\).
\end{definition}

We show that for this type of graphs, the approximation of \eqref{eq:gradgeomapprox} converges in probability to the true value of the gradient as the number of vertices increases, under some conditions on the radius, which constrain the density of the graph. Before stating the related theorem, we remind the reader of some definitions for the asymptotic notations which are used in the following analysis.

\begin{definition} \label{def:asym}
Let \(f\) and \(g\) be two non-negative functions. Then,
\begin{align*}
f(n) \in O(g(n)) &\Leftrightarrow \exists{}k > 0\;\exists{}n_0\;\forall{}n \geq n_0:\;f(n) \leq kg(n),\\
f(n) \in \Theta(g(n)) &\Leftrightarrow f(n) \in O(g(n)) \wedge g(n) \in O(f(n)),\\
f(n) \in o(g(n)) &\Leftrightarrow \forall{}k > 0\;\exists{}n_0\;\forall{}n \geq n_0:\;f(n) < kg(n),\\
f(n) \in \omega(g(n)) &\Leftrightarrow \forall{}k > 0\;\exists{}n_0\;\forall{}n \geq n_0:\;f(n) > kg(n).
\end{align*}
\end{definition}

\begin{theorem} \label{th:gradapproxconv}
Let \(u: \mathbb{R}^2 \rightarrow \mathbb{R}\) be a differentiable function and \(\mathcal{G}(n,\rho(n))\) an RGG embedded in \([0,1]^2\), with \(\rho(n) \in \omega\left(n^{-1/2}\right)\,\cap\,o\left(1\right)\). For a vertex \(v\) of \(\mathcal{G}\), the approximation of \eqref{eq:gradgeomapprox} converges in probability to \(\nabla{u(v)}\).
\end{theorem}

\begin{IEEEproof}
It suffices to prove that the sum
\begin{equation} \label{eq:gradapproxsum}
\mathcal{S} = \sum_{i=1}^N \frac{u(w_i)-u(v)}{d(v,w_i)}\,\mathbf{e}_{vw_i}\,\Delta\phi(w_i)
\end{equation}
in \eqref{eq:gradgeomapprox} converges in probability to the integral (\ref{eq:I}). Firstly, we show that the norm \(\Lambda_n = \sup_{w \in \mathcal{N}(v)}\{\Delta\phi(w)\}\) of the partition of \([0,2\pi]\) induced by the neighbor angles converges in probability to \(0\) in the limit of large graphs.

\begin{figure}
  \centering
  \begin{tikzpicture}

    \coordinate (Origin)   at (0,0);
    \coordinate (XAxisMin) at (-0.5,0);
    \coordinate (XAxisMax) at (2.5,0);
    \coordinate (BottomLeft) at (-0.5,-0.2);
    \coordinate (TopRight) at (2.5,2);
    
    \clip (BottomLeft) rectangle (TopRight); 
  
    \draw [thin, gray,-latex] (XAxisMin) -- (XAxisMax); 
    \filldraw[fill=gray, fill opacity=0.3, draw=black]
    (Origin) -- (2,1) arc (26.565:60:2.236) -- (Origin);
    \node[draw,circle,inner sep=2pt,fill,label=above left:$v$] (v) at (Origin) {};
    \draw (1,0) arc (0:26.565:1);
    \draw (0.5,0.25) arc (26.565:60:0.559);
    \node[thick,font=\large] at (45:1.5) {$S$};
    \node[font=\footnotesize] at (68:1.3) {$\rho$};
    \node[font=\footnotesize] at (45:0.8) {$\theta$};
    \node[font=\footnotesize] at (13:1.3) {$\theta_0$};

  \end{tikzpicture}
  \caption{Sector \(S(v,\rho,\theta_0,\theta)\).}
  \label{fig:Sector}
\end{figure}
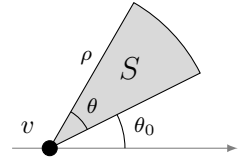

Let \(S(v,\rho,\theta_0,\theta)\) be a circular sector centered at \(v\), with radius \(\rho\), occupying an angle \(\theta > 0\) and whose rightmost radius is in the direction \(\theta_0\) (see Fig. \ref{fig:Sector}). For every vertex \(z_i,\,i = 1, \dots, n\) other than \(v\), we can define a Bernoulli random variable indicating whether this vertex is inside \(S\): \(Z_i \sim \bernoulli\left(\rho^2\,\theta / 2\right)\). The sum of these variables follows a binomial distribution:

\[Z = \sum_{\substack{i=1\\z_i \neq v}}^n{Z_i} \sim \binomial\left(n-1,\frac{\rho^2(n)\,\theta}{2}\right).\]

Therefore, the probability that \(S\) is empty of vertices other than \(v\) is \(\prob(Z = 0) = \left(1 - \rho^2(n)\,\theta / 2\right)^{n-1}\), which converges to \(0\), because \(\rho(n) \in \omega\left(n^{-1/2}\right)\). Let us consider a neighbor \(w\) of \(v\) and the event
\begin{align*}
A =\;&\exists\,y, z \in \mathcal{V}\setminus\{v, w\}:\,y \neq z\\
&{\wedge}\:y \in S\left(v,\rho,\phi(w),\frac{\theta}{2}\right)\\
&{\wedge}\:z \in S\left(v,\rho,\phi(w)-\frac{\theta}{2},\frac{\theta}{2}\right),
\end{align*}
i.e. there is another neighbor of \(v\) ``closer'' than \(\theta / 2\) on each side of \(w\). According to the above analysis, it is straightforward that \(\lim_{n \to +\infty}\prob(A) = 1\). Moreover, \(A\) implies \(\Delta\phi(w) \leq \theta\). Thus, it holds that \(\lim_{n \to +\infty}\prob(\Delta\phi(w) \leq \theta) = 1\) and consequently 
\begin{equation} \label{eq:partitionnormconv}
\lim_{n \to + \infty}\prob(\Lambda_n \leq \theta) = 1\;\forall \theta > 0,
\end{equation}
which concludes the first part of the proof.

Secondly, we show that the sum
\[\mathcal{S}_1 = \sum_{i=1}^N D_{\phi(w_i)}u(v)\,\mathbf{e}_{vw_i}\,\Delta\phi(w_i)\]
converges in probability to \(\mathcal{I}\). \(\mathcal{S}_1\) constitutes a Riemann sum of \(\mathbf{f}(\phi) = D_{\phi}u(v)\,\mathbf{e}_\phi\) over \([0,2\pi]\) with respect to the partition induced by the neighbor angles. Therefore, \eqref{eq:partitionnormconv} directly implies that \(\mathcal{S}_1\) converges in probability to \(\mathcal{I}\).

The third step of the proof is to show that the difference \(\mathcal{S} - \mathcal{S}_1\) converges in probability to \(\mathbf{0}\). Using the triangle inequality, we obtain
\[\vecnorm{\mathcal{S} - \mathcal{S}_1} \leq \sum_{i=1}^N \left|\frac{u(w_i)-u(v)}{d(v,w_i)}-D_{\phi(w_i)}u(v)\right|\Delta\phi(w_i).\]
For every \(w \in \mathcal{N}(v)\), the first order Taylor approximation of \(u\) at \(v\) in the direction \(\phi(w)\) yields
\[\left|u(w)-u(v)-d(v,w)D_{\phi(w)}u(v)\right| \in O\left(d^2(v,w)\right).\]
We divide both sides with \(d(v,w)\) and use the fact that \(0 \leq d(v,w) \leq \rho(n)\) to arrive at
\[\left|\frac{u(w)-u(v)}{d(v,w)}-D_{\phi(w)}u(v)\right| \in O(\rho(n)).\]
The last result holds for every neighbor of \(v\), so we can substitute each term of the sum to get
\[\vecnorm{\mathcal{S} - \mathcal{S}_1} \in \sum_{i=1}^N O(\rho(n))\Delta\phi(w_i) = O(\rho(n))\sum_{i=1}^N \Delta\phi(w_i).\]
The sum of neighbor angles over all neighbors is constant and equals \(2\pi\), which in turn implies that
\begin{equation} \label{eq:E1boundpreliminary}
\vecnorm{\mathcal{S} - \mathcal{S}_1} \in 2\pi{}O(\rho(n)) = O(\rho(n)).
\end{equation}
If we further make use of the fact that \(\rho(n) \in o(1)\), we get that \(\vecnorm{\mathcal{S} - \mathcal{S}_1} \in o(1)\). Consequently, \(\mathcal{S} - \mathcal{S}_1\) converges to \(\mathbf{0}\) almost surely and hence in probability as well.

Finally, we combine the results from the second and third step to conclude the proof. For arbitrary \(\epsilon > 0\), we use the triangle inequality to get
\begin{align*}
\prob(\vecnorm{\mathcal{S} - \mathcal{I}} > \epsilon)\;&{\leq}\;\prob(\vecnorm{\mathcal{S} - \mathcal{S}_1} + \vecnorm{\mathcal{S}_1 - \mathcal{I}} > \epsilon)\\
&{\leq}\;\prob\left(\vecnorm{\mathcal{S} - \mathcal{S}_1} > \epsilon\right) + \prob\left(\vecnorm{\mathcal{S}_1 - \mathcal{I}} > \epsilon\right).
\end{align*}
The above inequalities hold in the limit as well:
\begin{align*}
&\lim_{n \to +\infty}\prob(\vecnorm{\mathcal{S} - \mathcal{I}} > \epsilon)\\
&{\leq}\;\lim_{n \to +\infty}\prob\left(\vecnorm{\mathcal{S} - \mathcal{S}_1} > \epsilon\right) + \prob\left(\vecnorm{\mathcal{S}_1 - \mathcal{I}} > \epsilon\right) = 0,
\end{align*}
where the last equality is due to convergence in probability of \(\mathcal{S} - \mathcal{S}_1\) to \(\mathbf{0}\) and of \(\mathcal{S}_1\) to \(\mathcal{I}\).
\end{IEEEproof}

\subsection{Asymptotic Analysis of Approximation Error}
Going one step further, we decompose the error introduced by the geometric gradient approximation in order to obtain a bound on the \emph{rate} of convergence to the true gradient as the size of the RGG grows large. Since the framework in RGGs is stochastic, our results involve \emph{expectations} for the various quantities.

Let us denote the error in approximating \(\mathcal{I}\) with \(\mathcal{S}\) by \(\mathcal{E} = \mathcal{S} - \mathcal{I}\). Comparing the two expressions, we deduce that the approximation in \(\mathcal{S}\) is threefold:
\begin{enumerate}
\item Directional derivatives along edges are approximated with difference quotients.
\item The approximate value for the directional derivative along each edge is used as a constant estimate for all the directions ``falling into'' the respective neighbor angle.
\item The unit vector in the direction of each edge is also used for all the directions corresponding to the respective neighbor angle.
\end{enumerate}

To isolate the above sources of error, we construct intermediate expressions between \(\mathcal{S}\) and \(\mathcal{I}\) and bound the magnitude of the resulting differences.

\begin{theorem} \label{th:gradapproxerrorbound}
Let \(u: \mathbb{R}^2 \rightarrow \mathbb{R}\) be a differentiable function and \(\mathcal{G}(n,\rho(n))\) an RGG embedded in \([0,1]^2\), with \(\rho(n) \in \omega\left(n^{-1/2}\right)\,\cap\,o\left(1\right)\). For every vertex \(v\) of \(\mathcal{G}\), it holds that
\begin{equation} \label{eq:Ebound}
\expected[\vecnorm{\mathcal{E}}] \in O\left(\rho(n) + \frac{1}{n\,\rho^2(n)}\right).
\end{equation}
\end{theorem}

\begin{IEEEproof}
The first intermediate expression is \(\mathcal{S}_1\), corresponding to the first part of the error
\begin{align*}\mathcal{E}_1 &= \mathcal{S} - \mathcal{S}_1\\
&= \sum_{i=1}^N \left(\frac{u(w_i)-u(v)}{d(v,w_i)} - D_{\phi(w_i)}u(v)\right)\,\mathbf{e}_{vw_i}\,\Delta\phi(w_i).
\end{align*}
We have shown in the proof of Theorem \ref{th:gradapproxconv} that \(\vecnorm{\mathcal{E}_1}\) is bounded asymptotically by the radius of the graph (c.f. \eqref{eq:E1boundpreliminary}), as the difference quotients which are used in \(\mathcal{S}\) are first order approximations of the corresponding directional derivatives:
\begin{equation} \label{eq:E1bound}
\vecnorm{\mathcal{E}_1} \in O(\rho(n)).
\end{equation}

We define the second intermediate expression as
\[\mathcal{S}_2 = \sum_{i=1}^N \int_{\omega(w_i)}^{\omega(w_i) + \Delta\phi(w_i)} D_{\phi}u(v)\,\mathbf{e}_{vw_i}\,d\phi.\]
If we denote the direction of \(\nabla{u(v)}\) by \(\theta\), we can write \(D_{\phi}u(v) = \vecnorm{\nabla{u(v)}}\cos(\theta-\phi)\) and the second part of the error can be expressed as
\begin{align*}
\mathcal{E}_2\;&{=}\;\mathcal{S}_1 - \mathcal{S}_2&\\
&{=}\;\vecnorm{\nabla{u(v)}}\sum_{i=1}^N \mathbf{e}_{vw_i}\int_{\omega(w_i)}^{\omega(w_i)+\Delta\phi(w_i)} &{(}\cos(\theta-\phi(w_i))\\
&&{-}\:\cos(\theta-\phi))\,d\phi.
\end{align*}
After performing some calculations, we obtain
\begin{multline*}
\int_{\omega(w_i)}^{\omega(w_i)+\Delta\phi(w_i)} \cos(\theta-\phi))\,d\phi\\ =2\cos\left(\theta-\omega(w_i)-\frac{\Delta\phi(w_i)}{2}\right)\sin\left(\frac{\Delta\phi(w_i)}{2}\right).
\end{multline*}
The first factor on the right-hand side can be expanded using the identity \(\cos(a-b) = \cos(a)\cos(b)+\sin(a)\sin(b)\) as
\begin{align*}
&\cos\left(\theta-\omega(w_i)-\frac{\Delta\phi(w_i)}{2}\right)\\
&{=}\;\cos(\theta-\phi(w_i))\cos\left(\omega(w_i)+\frac{\Delta\phi(w_i)}{2}-\phi(w_i)\right)\\
&{+}\:\sin(\theta-\phi(w_i))\sin\left(\omega(w_i)+\frac{\Delta\phi(w_i)}{2}-\phi(w_i)\right).
\end{align*}
Furthermore, if we apply the definitions of angles \(\omega(w_i)\) and \(\Delta\phi(w_i)\), we get the following bound:
\begin{align*}
\left|\omega(w_i)+\frac{\Delta\phi(w_i)}{2}-\phi(w_i)\right|\;&{\leq}\;\frac{\Delta\phi(w_i)}{2}\\
\Rightarrow \omega(w_i)+\frac{\Delta\phi(w_i)}{2}-\phi(w_i)\:&{\in}\:O(\Delta\phi(w_i))
\end{align*}
We combine the Taylor expansions of the sine and cosine functions around \(0\)
\[\cos(a) = 1+O\left(a^2\right)\text{ and }\sin(a) = a+O\left(a^3\right) = O(a)\]
with the above bound into
\begin{align*}
&2\cos\left(\theta-\omega(w_i)-\frac{\Delta\phi(w_i)}{2}\right)\sin\left(\frac{\Delta\phi(w_i)}{2}\right)\\
&{=}\;2\left(\cos(\theta-\phi(w_i))\left(1+O\left({\Delta\phi(w_i)}^2\right)\right)\right.\\
&{+}\:\sin(\theta-\phi(w_i))O(\Delta\phi(w_i))\left.\vphantom{\left(1+O\left({\Delta\phi(w_i)}^2\right)\right)}\right)\left(\frac{\Delta\phi(w_i)}{2}+O\left({\Delta\phi(w_i)}^3\right)\right)\\
&{=}\;\cos(\theta-\phi(w_i))\Delta\phi(w_i)+\cos(\theta-\phi(w_i))O\left({\Delta\phi(w_i)}^3\right)\\
&{+}\:\sin(\theta-\phi(w_i))O\left({\Delta\phi(w_i)}^2\right).
\end{align*}
We substitute the above expression into \(\mathcal{E}_2\) and use the triangle inequality and the fact that \(\vecnorm{\mathbf{e}_{vw_i}} = 1\;\forall i \in \{1,\,\dots,\,N\}\) to obtain
\begin{align*}
\vecnorm{\mathcal{E}_2} \in \vecnorm{\nabla{u(v)}}\sum_{i=1}^N &\left|\cos(\theta-\phi(w_i))O\left({\Delta\phi(w_i)}^3\right)\right.\\
&{+}\:\left.\sin(\theta-\phi(w_i))O\left({\Delta\phi(w_i)}^2\right)\right|.
\end{align*}
Additionally, the absolute values of the sine and the cosine in the last result are bounded from above by \(1\). Thus, the second part of the error is bounded by
\begin{align}
\vecnorm{\mathcal{E}_2}\,&{\in}\,\vecnorm{\nabla{u(v)}}\sum_{i=1}^N O\left({\Delta\phi(w_i)}^2\right)\nonumber\\
&{=}\;\vecnorm{\nabla{u(v)}}O\left(\sum_{i=1}^N {\Delta\phi(w_i)}^2\right).\label{eq:E2boundwithoutexpect}
\end{align}

The next step is to take the expectation for both sides of \eqref{eq:E2boundwithoutexpect}. Expectation preserves inequalities and it is straightforward that \(f \in O(g)\) implies \(\expected[f] \in O\left(\expected[g]\right)\) for two sequences of random variables \(f\) and \(g\). As a result, it holds that
\[\expected\left[\vecnorm{\mathcal{E}_2}\right] \in \vecnorm{\nabla{}u(v)}O\left(\expected\left[\sum_{i=1}^N {\Delta\phi(w_i)}^2\right]\right).\]
Since \(N\) is itself a random variable, we employ the law of total expectation and take advantage of the fact that all \(\Delta\phi(w_i)\) are identically distributed to write
\begin{align*}
&{\expected}\left[\sum_{i=1}^N {\Delta\phi(w_i)}^2\right]\\
&{=}\;\expected\left[\expected\left[\left.\sum_{i=1}^N {\Delta\phi(w_i)}^2\right|N\right]\right] = \expected\left[\sum_{i=1}^N \expected\left[\left.{\Delta\phi(w_i)}^2\right|N\right]\right]\\
&{=}\;\expected\left[N\expected\left[\left.{\Delta\phi(w)}^2\right|N\right]\right]\text{ for some }w \in \mathcal{N}(v).
\end{align*}
Let us focus on the term \(\expected\left[\left.{\Delta\phi(w)}^2\right|N\right]\). In order to calculate this conditional expectation, we examine the distribution of the random variable \(\Delta\phi(w)\). The probability that \(\Delta\phi(w) \leq x\) is equal to the probability that at least two neighbors of \(v\) other than \(w\) fall inside the \(2x\) radial interval. Taking into account all possible combinations, it follows that for \(N \geq 3\)
\begin{align*}
\prob(\Delta\phi(w) \leq \pi{}x) &= \sum_{k=2}^{N-1} \binom{N-1}{k}x^k{(1-x)}^{N-1-k}\\
&= 1-(N-1)x{(1-x)}^{N-2}-{(1-x)}^{N-1},
\end{align*}
for \(x \in \left[0,\,1\right]\). The corresponding PDF of the random variable \(\Delta\phi(w)/\pi\) is
\[(N-1)(N-2)x{(1-x)}^{N-3},\,x \in \left[0,\,1\right]\]
and therefore \(\Delta\phi(w)/\pi\) follows a Beta distribution with parameters \(\alpha=2\) and \(\beta=N-2\). We use the formulas for the mean and variance of a Beta distribution with known parameters to write
\begin{align*}
\expected\left[\left.\left(\frac{\Delta\phi(w)}{\pi}\right)^2\right|N\right] &= \variance\left[\frac{\Delta\phi(w)}{\pi}\right]+\left(\expected\left[\frac{\Delta\phi(w)}{\pi}\right]\right)^2\\
&= \frac{2(N-2)}{N^2(N+1)}+\frac{4}{N^2} = \frac{6}{N(N+1)}.
\end{align*}
Due to linearity of expectation, the conditional expectation we are after is
\begin{equation} \label{eq:Deltaphiwsquaredexpected}
\expected\left[\left.{\Delta\phi(w)}^2\right|N\right] = \frac{6\pi^2}{N(N+1)}.
\end{equation}

Using \eqref{eq:Deltaphiwsquaredexpected}, we obtain
\begin{align*}
\expected\left[\vecnorm{\mathcal{E}_2}\right]\,&{\in}\,\vecnorm{\nabla{u(v)}} O\left(\expected\left[N\frac{6\pi^2}{N(N+1)}\right]\right)\\
&{=}\;\vecnorm{\nabla{u(v)}} O\left(\expected\left[\frac{1}{N+1}\right]\right).
\end{align*}
We compute the expectation \(\expected\left[1/(N+1)\right]\) using the binomial distribution of the number of neighbors of \(v\), \(N \sim \binomial\left(n-1,\pi\rho^2\right)\). The definition of this expectation is
\begin{align*}
&{\expected}\left[\frac{1}{N+1}\right]\\
&{=}\;\sum_{k=0}^{n-1} \frac{1}{k+1}\binom{n-1}{k}\left(\pi\rho^2\right)^k\left(1-\pi\rho^2\right)^{n-1-k}\\
&{=}\;\frac{1}{n\pi\rho^2} \sum_{k=0}^{n-1} \binom{n}{k+1}\left(\pi\rho^2\right)^{k+1}\left(1-\pi\rho^2\right)^{n-(k+1)}\\
&{=}\;\frac{1}{n\pi\rho^2} \sum_{k=1}^{n} \binom{n}{k}\left(\pi\rho^2\right)^k\left(1-\pi\rho^2\right)^{n-k}\\
&{=}\;\frac{1}{n\pi\rho^2}\left(1-\binom{n}{0}\left(\pi\rho^2\right)^0\left(1-\pi\rho^2\right)^n\right)\\
&{=}\;\frac{1}{n\pi\rho^2}\left(1-\left(1-\pi\rho^2\right)^n\right).
\end{align*}
Since \(\rho(n) \in \omega\left(n^{-1/2}\right)\,\cap\,o\left(1\right)\), it follows that
\[\left(1-\left(1-\pi\rho^2(n)\right)^n\right) \in \Theta(1).\]
Consequently, the expectation of the second part of the error is bounded through
\begin{equation} \label{eq:E2bound}
\expected[\vecnorm{\mathcal{E}_2}] \in \vecnorm{\nabla{u(v)}}\,O\left(\frac{1}{n\,\rho^2(n)}\right).
\end{equation}

Finally, the third part of the approximation error is
\begin{align*}
\mathcal{E}_3\;&{=}\;\mathcal{S}_2 - \mathcal{I}\\
&{=}\;\vecnorm{\nabla{u(v)}}\sum_{i=1}^N \int\limits_{\omega(w_i)}^{\omega(w_i)+\Delta\phi(w_i)} \cos(\theta-\phi)\left(\mathbf{e}_{vw_i}-\mathbf{e}_{\phi}\right)\,d\phi.
\end{align*}
For the \(i\)-th term of the above sum, it holds that \(|\phi-\phi(w_i)| \leq \Delta\phi(w_i)\), which further implies that
\begin{align*}
\vecnorm{\mathbf{e}_{vw_i}-\mathbf{e}_{\phi}}\;&{\leq}\; \vecnorm{\mathbf{e}_{vw_i}-\mathbf{e}_{\phi(w_i)+\Delta\phi(w_i)}}\\
&{=}\;2\sin\left(\frac{\Delta\phi(w_i)}{2}\right).
\end{align*}
Thus, the magnitude of \(\mathcal{E}_3\) can be bounded using the triangle inequality as follows:
\begin{align*}
\vecnorm{\mathcal{E}_3}\;&{\leq}\;\vecnorm{\nabla{u(v)}}\sum_{i=1}^N \int_{\omega(w_i)}^{\omega(w_i)+\Delta\phi(w_i)} \vecnorm{\mathbf{e}_{vw_i}-\mathbf{e}_{\phi}}\,d\phi\\
&{\leq}\;\vecnorm{\nabla{u(v)}}\sum_{i=1}^N \int_{\omega(w_i)}^{\omega(w_i)+\Delta\phi(w_i)} 2\sin\left(\frac{\Delta\phi(w_i)}{2}\right)\,d\phi\\
&{=}\;\vecnorm{\nabla{u(v)}}\sum_{i=1}^N 2\Delta\phi(w_i)\sin\left(\frac{\Delta\phi(w_i)}{2}\right)\\
&{\in}\,\vecnorm{\nabla{u(v)}}\sum_{i=1}^N O\left({\Delta\phi(w_i)}^2\right).
\end{align*}
The last bound is the same as the one that has been derived in \eqref{eq:E2boundwithoutexpect} for \(\vecnorm{\mathcal{E}_2}\), which yields:
\begin{equation} \label{eq:E3bound}
\expected[\vecnorm{\mathcal{E}_3}] \in \vecnorm{\nabla{u(v)}}\,O\left(\frac{1}{n\,\rho^2(n)}\right).
\end{equation}

The total error is \(\mathcal{E} = \mathcal{E}_1 + \mathcal{E}_2 + \mathcal{E}_3\) and due to the triangle inequality and the fact that expectation preserves inequalities, it follows that
\[\expected[\vecnorm{\mathcal{E}}] \leq \expected[\vecnorm{\mathcal{E}_1}] + \expected[\vecnorm{\mathcal{E}_2}] + \expected[\vecnorm{\mathcal{E}_3}].\]
Based on the asymptotic bounds in \eqref{eq:E1bound}, \eqref{eq:E2bound} and \eqref{eq:E3bound} and the sum property of the \(O\) symbol, we derive the bound of the total error
\[\expected[\vecnorm{\mathcal{E}}] \in O\left(\rho(n) + \frac{1}{n\,\rho^2(n)}\right).\]
\end{IEEEproof}

\begin{figure}
    \centering
    \includegraphics[width = 0.8\linewidth]{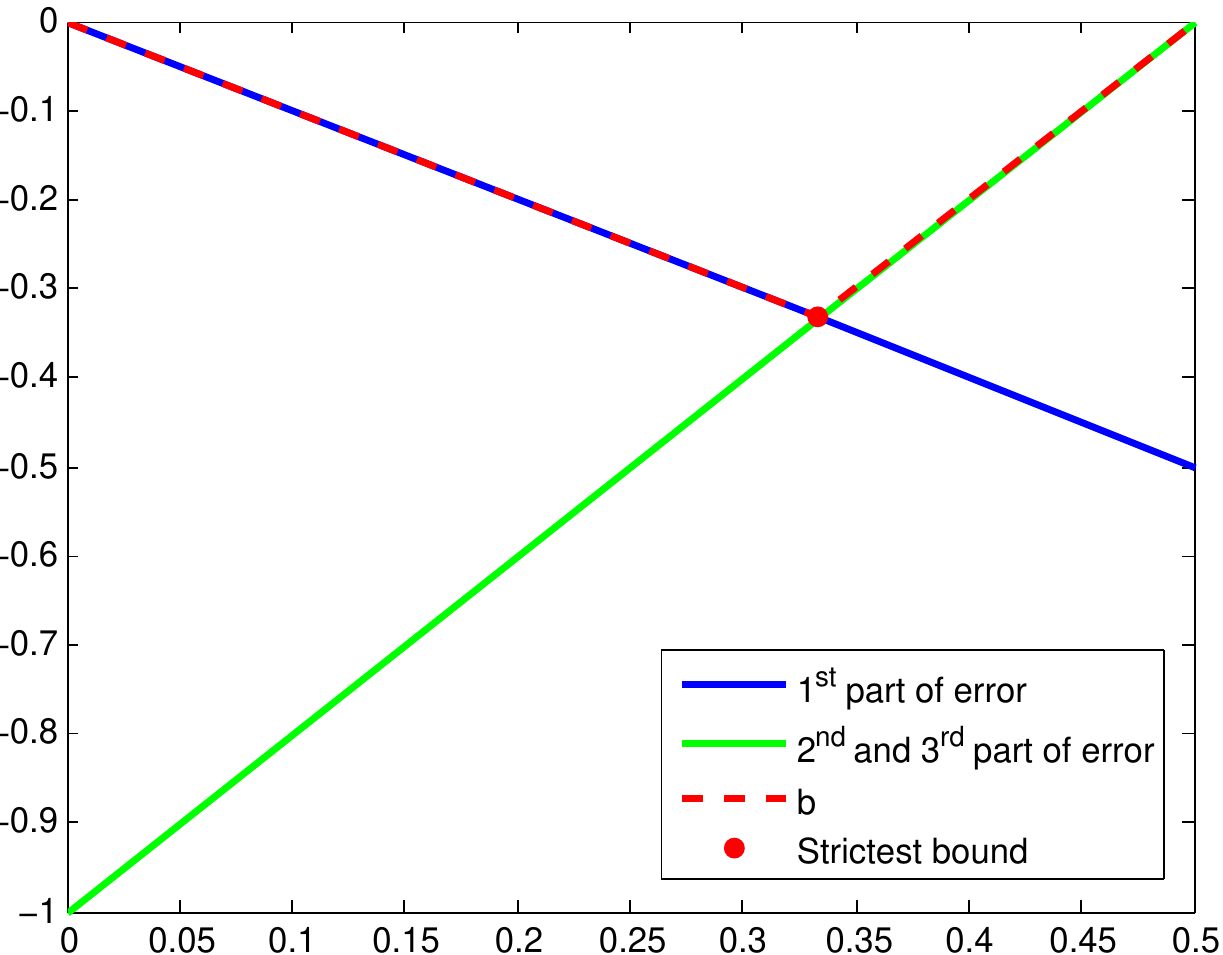}
    \caption{Variation of exponent \(b\) of the asymptotic bound for gradient approximation error, with respect to exponent \(a\) of the radius. Smaller values of \(b\) mean stricter error bounds. The exponents of the individual parts of the error are also presented.}
    \label{fig:ErrorBoundExponent}
\end{figure}

The radius of the graph is effectively the factor that determines the strictness of this bound. To provide better intuition, we study the case when \(\rho(n) \in \Theta\left(n^{-a}\right),\,a \in (0, 1 / 2)\). Substituting in \eqref{eq:Ebound}, we obtain
\begin{equation} \label{eq:Eboundspecialcase}
\expected[\vecnorm{\mathcal{E}}] \in O\left(n^{b}\right),\;b = \left\{
{\def\arraystretch{1.2}
\begin{array}{rl}
-a, & a \in \left(0,\frac{1}{3}\right]. \\
-1 + 2a, & a \in \left(\frac{1}{3},\frac{1}{2}\right).
\end{array}
}
\right\}
\end{equation}
We visualize this expression for the error bound in Fig. \ref{fig:ErrorBoundExponent}. The strictest upper bound is \(O\left(n^{-1/3}\right)\), it is achieved for \(a = 1/3\) and it constitutes a tradeoff between minimizing the first error term, which calls for small radii, and the two other terms, which requires more neighbors and consequently larger radii.

From a practical point of view, \eqref{eq:E2bound} and \eqref{eq:E3bound} indicate that the error in the approximation of \eqref{eq:gradgeomapprox} increases as the magnitude of the true gradient grows large, i.e. when the function exhibits abrupt variations. This does not pose a problem for the calculation of gradient direction, since the latter does not depend on the range of the function's variation around the examined vertex. Utilizing all incident edges in the weighted sum of \eqref{eq:gradgeomapprox} ensures that all available information in the neighborhood of the vertex is used to estimate which direction the gradient points to, as emphasized in \cite{DrMa12}. However, the estimated gradient \emph{magnitude} with our approximation is prone to greater error, as it depends on the range of the function's variation. The use of difference quotients in \eqref{eq:gradgeomapprox} accentuates this effect for dense graphs, where distances between neighboring vertices that appear in the denominator of the quotients approach zero. To circumvent this issue in practice, we adopt the approximation of \cite{DrMa12} for gradient magnitude, namely the maximum absolute difference of values of the function along edges that are incident on \(v\):
\begin{equation} \label{eq:gradmagnmaxabsdiff}
\vecnorm{\nabla{u}(v)} \approx \max_{w \in \mathcal{N}(v)}\{|u(w)-u(v)|\}.
\end{equation}

\section{Curvature Approximation on Graphs} \label{sec:curvapprox}
After having devised an approximation scheme for the gradient of an embedding function, the next step is to use this scheme for estimating the curvature of the level sets of this function. The difference from the gradient case is that the input gradient values for curvature approximation are already approximate themselves, i.e. a \emph{cascaded} approximation is attempted. Therefore, the error in curvature approximation on a graph is expected to accumulate compared to gradient approximation error on the same graph, since the estimated curvature at a vertex inherits the error of the estimated gradients at its neighboring vertices.

\subsection{Geometric Approximation} \label{subsec:curvgeom}
In this type of curvature approximation, we follow an approach similar to \cite{DrMa12}. More specifically, we exploit the expression of curvature as the divergence of the unit gradient field \(\mathbf{F} = \nabla{u}/\vecnorm{\nabla{u}}\) of the embedding function \(u\):
\begin{equation} \label{eq:curvdiv}
\kappa(v) = \divergence\mathbf{F}(v),\;\nabla{u(v)} \neq \mathbf{0}.
\end{equation}

\begin{figure*}
  \centering
  \subfloat[]{
  \begin{tikzpicture}[scale=0.7]
  
    \coordinate (Origin)   at (0,0);
    \coordinate (W1)   at (2,2);
    \coordinate (W2)   at (0,3);
    \coordinate (W3)   at (-1,0);
    
    \fill[fill=gray, fill opacity=0.3] (4.5,-0.5) -- (1,3) -- (-1,3) -- (-1,-1.5);
    \node [draw,circle,inner sep=2pt,fill,label=above right:$w_1$] (w1) at (W1) {};
    \node [draw,circle,inner sep=2pt,fill,label=below left:$w_2$] (w2) at (W2) {};
    \node [draw,circle,inner sep=2pt,fill,label=left:$w_3$] (w3) at (W3) {};
    \node[draw,circle,inner sep=2pt,fill,label=below:$v$] (v) at (Origin) {};
    \draw (Origin) -- (W1);
    \draw (Origin) -- (W2);
    \draw (Origin) -- (W3);
    \draw[dash pattern=on5pt off3pt] (4.5,-0.5) -- (0.5,3.5) ;
    \draw[dash pattern=on5pt off3pt] (-1.5,3) -- (1.5,3) ;
    \draw[dash pattern=on5pt off3pt] (-1,3.5) -- (-1,-1.5) ;
    \node at (3.3,1.5) {$L(w_1)$};
    \node at (-0.3,3.5) {$L(w_2)$};
    \node at (-1.7,1.5) {$L(w_3)$};
    
    \node[thick,font=\large] at (1.5,0) {$S(v)$};
    
  \end{tikzpicture}
  \label{subfig:infiniteS}}
  \hfil
  \subfloat[]{
  \begin{tikzpicture}[scale=0.7]
  
    \coordinate (Origin)   at (0,0);
    \coordinate (W1)   at (1,2);
    \coordinate (W2)   at (-2,1);
    \coordinate (W3)   at (-4,0.5);
    \coordinate (W4)   at (-2,-1);
    \coordinate (W5)   at (1,-1);
    
    \fill[fill=gray, fill opacity=0.3] (3,1) -- (-1,3) -- (-2.5,0) -- (-1,-3);
    \node [draw,circle,inner sep=2pt,fill,label=above right:$w_1$] (w1) at (W1) {};
    \node [draw,circle,inner sep=2pt,fill,label=above left:$w_2$] (w2) at (W2) {};
    \node [draw,circle,inner sep=2pt,fill,label=left:$w_3$] (w3) at (W3) {};
    \node [draw,circle,inner sep=2pt,fill,label=left:$w_4$] (w4) at (W4) {};
    \node [draw,circle,inner sep=2pt,fill,label=below right:$w_5$] (w5) at (W5) {};
    \node[draw,circle,inner sep=2pt,fill,label=below:$v$] (v) at (Origin) {};
    \draw (Origin) -- (W1);
    \draw (Origin) -- (W2);
    \draw (Origin) -- (W3);
    \draw (Origin) -- (W4);
    \draw (Origin) -- (W5);
    \draw[dash pattern=on5pt off3pt] (3.4,0.8) -- (-1.4,3.2) ;
    \draw[dash pattern=on5pt off3pt] (-0.8,3.4) -- (-2.7,-0.4) ;
    \draw[dash pattern=on5pt off3pt] (-3.75,2.5) -- (-4.25,-1.5) ;
    \draw[dash pattern=on5pt off3pt] (-2.7,0.4) -- (-0.8,-3.4) ;
    \draw[dash pattern=on5pt off3pt] (-1.3,-3.3) -- (3.3,1.3) ;
    \node at (0,3) {$L(w_1)$};
    \node at (-2.2,2) {$L(w_2)$};
    \node at (-2.2,-2) {$L(w_4)$};
    \node at (2,-0.8) {$L(w_5)$};
    
    \node[thick,font=\large] at (-0.5,1) {$S(v)$};
    
  \end{tikzpicture}
  \label{subfig:ignoredWi}}
 \hfil
  \subfloat[]{
  \begin{tikzpicture}[scale=0.7]
  
    \coordinate (Origin)   at (0,0);
    \coordinate (W1)   at (3.0642,2.5712);
    \coordinate (W2)   at (-1.8794,0.6840);
    \coordinate (W3)   at (-0.5209,-2.9544);
  
    \filldraw[fill=gray, fill opacity=0.3, draw=black, thick]
      (3.4641,-2) arc (-30:100:4) -- (-0.3473,1.9696) arc (100:210:2) -- (-2.5981,-1.5) arc (210:330:3) -- (3.4641,-2);
  
    \node [draw,circle,inner sep=2pt,fill,label=above right:$w_1$] (w1) at (W1) {};
    \node [draw,circle,inner sep=2pt,fill,label=left:$w_2$] (w2) at (W2) {};
    \node [draw,circle,inner sep=2pt,fill,label=below:$w_3$] (w3) at (W3) {};
    \node [draw,circle,inner sep=2pt,fill,label=right:$v$] (v) at (Origin) {};
    \draw (Origin) -- (W1);
    \draw (Origin) -- (W2);
    \draw (Origin) -- (W3);
  
    \node[thick,font=\large] at (3,1) {$S(v)$};
  
    \draw[dash pattern=on6pt off2pt] (Origin) -- (-0.3473,1.9696) ;
    \draw[dash pattern=on6pt off2pt] (Origin) -- (-1.73205,-1) ;
    \draw[dash pattern=on6pt off2pt] (Origin) -- (2.5981,-1.5) ;
  
    \draw[dash pattern=on6pt off2pt] (0.9526,-0.55) arc (-30:100:1.1);
    \draw[dash pattern=on6pt off2pt] (-0.1216,0.6894) arc (100:210:0.7);
    \draw[dash pattern=on6pt off2pt] (-0.7794,-0.45) arc (210:330:0.9);
    \node[font=\tiny] at (1.2,1) {$\Delta\phi(w_1)$};
    \node[font=\tiny] at (-1,0.5) {$\Delta\phi(w_2)$};
    \node[font=\tiny] at (0.2,-1.1) {$\Delta\phi(w_3)$};
  
  \end{tikzpicture}
  \label{subfig:Scircularsectors}}
  \caption{Deficiencies of original curvature approximation of \cite{DrMa12} in \protect\subref{subfig:infiniteS} and \protect\subref{subfig:ignoredWi} and solution through the new geometric approximation in \protect\subref{subfig:Scircularsectors}. In \protect\subref{subfig:infiniteS} the defined region has infinite area, while in \protect\subref{subfig:ignoredWi} neighbor \(w_3\) causes an unintuitive shape for \(S(v)\). These ill cases are handled properly by defining region \(S(v)\) through the neighbor angles, as done in \protect\subref{subfig:Scircularsectors}.}
  \label{fig:Sregions}
\end{figure*}
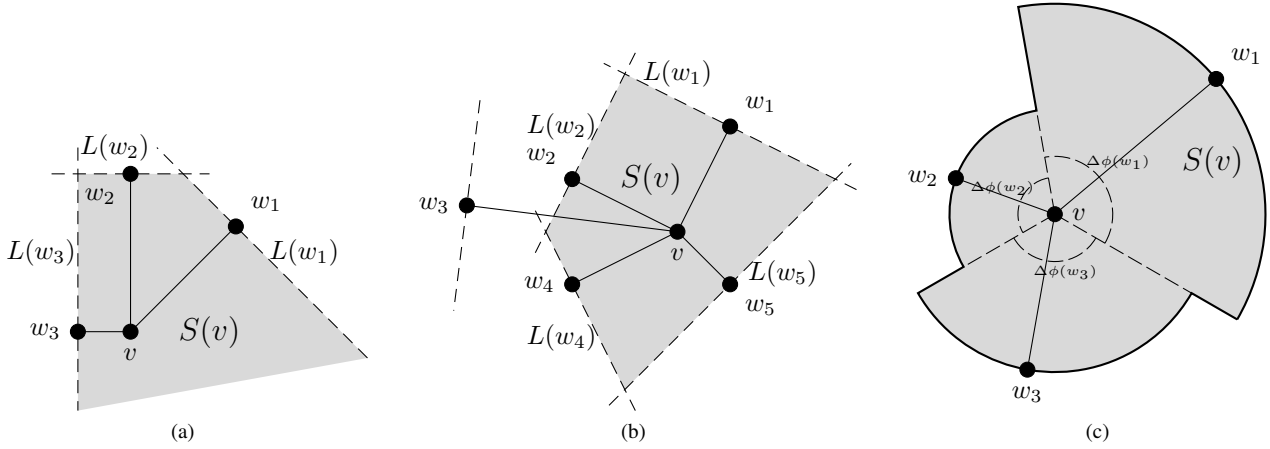

An integral definition of divergence as
\begin{equation} \label{eq:divlim}
\divergence \mathbf{F}(v) = \displaystyle{\lim_{S \to \{\mathbf{v}\}} \frac{\oint\limits_{\Gamma(S)}^{} \mathbf{F}\cdot\mathbf{n}\,d\ell}{\left|S\right|}}
\end{equation}
can then be used as a basis for geometric approximations of curvature, where \(S\) is a region with area \(\left|S\right|\) and boundary \(\Gamma(S)\) and \(\mathbf{n}\) is the outward unit normal to this boundary. In \cite{DrMa12}, the integral in \eqref{eq:divlim} is approximated using a polygonal region to form a finite sum over the neighbors of the vertex (as shown in \cite[Fig. 8]{DrMa12}). However, certain arrangements of the neighbors of the examined vertex, which are shown in Fig. \ref{fig:Sregions}, can lead to regions with ill-defined area, boundary and normals.

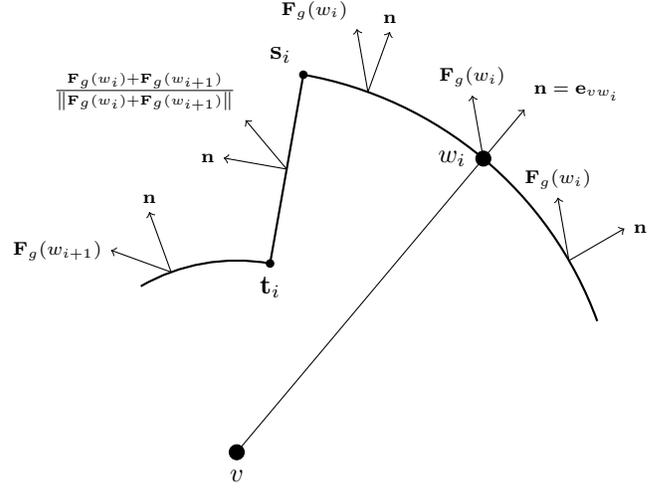
\begin{figure}
  \centering
  \begin{tikzpicture}[scale=0.85]
    \coordinate (Origin)   at (0,0);
    \coordinate (Wi)   at (3.8567,4.5963);
    \coordinate (C1) at (5.1962,3);
    \coordinate (C2) at (2.0521,5.6382);
    \coordinate (C3) at (0.7814,4.4316);
    \coordinate (C4) at (-1.0261,2.8191);
    \coordinate (Si) at (80:6);
    \coordinate (Ti) at (80:3);
    
    \node[draw,circle,inner sep=2pt,fill,label=left:$w_i$] (wi) at (Wi) {};
    \node[draw,circle,inner sep=2pt,fill,label=below:$v$] (v) at (Origin) {};
    \node[draw,circle,inner sep=1pt,fill,label=above left:$\mathbf{s}_i$] (si) at (Si) {};
    \node[draw,circle,inner sep=1pt,fill,label=below:$\mathbf{t}_i$] (ti) at (Ti) {};
    \draw (Origin) -- (Wi);
    
    \draw[->] (Wi) -- (4.4996,5.3623) node [anchor= south west,font=\scriptsize] {$\mathbf{n}=\mathbf{e}_{vw_i}$};
    \draw[->] (C1) -- (6.0622,3.5) node [anchor=west,font=\scriptsize] {$\mathbf{n}$};
    \draw[->] (C2) -- (2.3941,6.5778) node [anchor=south,font=\scriptsize] {$\mathbf{n}$};
    \draw[->] (C3) -- (-0.2034,4.6052) node [anchor=east,font=\scriptsize] {$\mathbf{n}$};
    \draw[->] (C4) -- (-1.3681,3.7588) node [anchor=south,font=\scriptsize] {$\mathbf{n}$};
    \draw[->] (Wi) -- (3.6831,5.5811) node [anchor=south,font=\scriptsize] {$\mathbf{F}_g(w_i)$};
    \draw[->] (C1) -- (5.0226,3.9848) node [anchor=south,font=\scriptsize] {$\mathbf{F}_g(w_i)$};
    \draw[->] (C2) -- (1.8785,6.6230) node [anchor=south east,font=\scriptsize] {$\mathbf{F}_g(w_i)$};
    \draw[->] (C3) -- (0.1386,5.1976) node [anchor=south east,font=\scriptsize] {$\frac{\mathbf{F}_g(w_i)+\mathbf{F}_g(w_{i+1})}{\vecnorm{\mathbf{F}_g(w_i)+\mathbf{F}_g(w_{i+1})}}$};
    \draw[->] (C4) -- (-1.9658,3.1611) node [anchor=east,font=\scriptsize] {$\mathbf{F}_g(w_{i+1})$};
    
    \draw[thick] (5.6382,2.0521) arc (20:80:6) -- (Ti) arc (80:120:3);
    
  \end{tikzpicture}
  \caption{The picture of values of \(\mathbf{F}_g\) at a part of the boundary of \(S(v)\) which corresponds to neighbor \(w_i\) of \(v\).}
  \label{fig:SFvectors}
\end{figure}

To tackle these problems, we employ again the \emph{neighbor angles} that were introduced in Section~\ref{sec:gradapprox}, in order to define the region \(S\) in \eqref{eq:divlim} in a more compact and principled fashion. For vertex \(v\), \(S(v)\) is formed as a union of circular sectors, each of them corresponding to a neighbor of \(v\), as we show in Fig. \ref{fig:Sregions}\subref{subfig:Scircularsectors}. More formally, for each neigbor \(w\) of \(v\), the respective circular sector is \(S(v,d(v,w),\omega(w),\Delta\phi(w))\). The area of \(S(v)\) can then be expressed as
\begin{equation} \label{eq:Sarea}
\left|S(v)\right| = \sum_{i=1}^N \frac{\Delta\phi(w_i)}{2}\,d^2(v,w_i).
\end{equation}
The challenge imposed by our construction of \(S(v)\) is the choice of suitable values for \(\mathbf{F}\) along the boundary of this region, given only its values at the locations of neighbors of \(v\). The resulting boundary consists of arcs, each of which contains a neighbor of \(v\), and line segments which connect these arcs. We fix the value of \(\mathbf{F}\) along each arc at the geometric approximation computed for the corresponding neighbor \(w\) using \eqref{eq:gradgeomapprox}, \(\mathbf{F}_g(w)\). Moreover, for every line segment, we use the normalized mean of the approximate values of \(\mathbf{F}\) along the two neighboring arcs. The concept is again to use information from the closest vertex, which should be more reliable. We visualize the described configuration in Fig. \ref{fig:SFvectors}.

Using the above approximations, we substitute the line integral in \eqref{eq:divlim} with a sum of simple line integrals over single arcs and line segments, which have closed analytical forms. If we denote the integral over the arc \(C_a(w_i)\) containing neighbor \(w_i\) by \(I_a(w_i)\) and the integral over the line segment \(C_l(w_i)\) that connects the arcs \(C_a(w_i)\) and \(C_a(w_{i+1})\) by \(I_l(w_i)\), we obtain 
\begin{align}
I_a(w_i) =\;&\int_{C_a(w_i)} \mathbf{F}_g(w_i)\cdot\mathbf{n}\,d\ell\nonumber\\
=\;&d(v,w_i)\,\mathbf{F}_g(w_i)\cdot(\sin(\omega(w_{i+1}))-\sin(\omega(w_i)),\nonumber\\
&\cos(\omega(w_i))-\cos(\omega(w_{i+1})))\label{eq:Ia}
\end{align}
and
\begin{align}
I_l(w_i) =\;&\int_{C_l(w_i)} \frac{\mathbf{F}_g(w_i)+\mathbf{F}_g(w_{i+1})}{\vecnorm{\mathbf{F}_g(w_i)+\mathbf{F}_g(w_{i+1})}}\cdot\mathbf{n}\,d\ell\nonumber\\
=\;&{(}d(v,w_{i+1})-d(v,w_i))\,\frac{\mathbf{F}_g(w_i)+\mathbf{F}_g(w_{i+1})}{\vecnorm{\mathbf{F}_g(w_i)+\mathbf{F}_g(w_{i+1})}}\nonumber\\
&{\cdot}(\sin(\omega(w_{i+1})),\,-\cos(\omega(w_{i+1}))).\label{eq:Il}
\end{align}
The proposed geometric approximation of curvature is given by
\begin{equation} \label{eq:curvgeomapprox}
\kappa(v) \approx \frac{\displaystyle\sum_{i=1}^N I_a(w_i)+I_l(w_i)}{\left|S(v)\right|}.
\end{equation}
For RGGs, this approximation is exact in the limit of large graphs like in the gradient approximation case, although the conditions are now stronger.

\begin{theorem} \label{th:curvgeomapproxconv}
Let \(\mathcal{G}(n,\rho(n))\) be an RGG embedded in \([0,1]^2\), with \(\rho(n) \in \omega\left(n^{-1/2}\right)\,\cap\,o\left(1\right)\) and \(v\) a vertex of \(\mathcal{G}\). If \(u: \mathbb{R}^2 \rightarrow \mathbb{R}\) is continuously differentiable and \(\nabla{u(v)} \neq \mathbf{0}\), then the approximation of \eqref{eq:curvgeomapprox} converges in probability to \(\kappa(v)\).
\end{theorem}

The full proof of Theorem \ref{th:curvgeomapproxconv} is given in the Appendix. A brief outline of the proof with its key ideas follows. The main task in the proof is to show that
\begin{equation} \label{eq:curvgeomapproxconvnumer}
\displaystyle\sum_{i=1}^N I_a(w_i)+I_l(w_i) - \oint\limits_{\Gamma(S(v))}^{} \mathbf{F}\cdot\mathbf{n}\,d\ell \xrightarrow{\prob} 0.
\end{equation}
We treat each component of \(\Gamma(S(v))\), i.e. each arc and line segment, separately and prove the following convergence in probability results:
\[I_a(w_i)-\int_{C_a(w_i)} \mathbf{F}\cdot\mathbf{n}\,d\ell \xrightarrow{\prob} 0\]
and
\[I_l(w_i)-\int_{C_l(w_i)} \mathbf{F}\cdot\mathbf{n}\,d\ell \xrightarrow{\prob} 0.\]
Afterwards, these convergence results are combined through the sum property of convergence in probability to obtain \eqref{eq:curvgeomapproxconvnumer}. To prove the above results, we make use of the continuity of both \(\mathbf{F}\) and \(\nabla{u}\) at \(v\), which is ensured by the conditions on \(u\) in Theorem \ref{th:curvgeomapproxconv}. In addition, we use the law of total probability to prove that certain probabilities vanish in the limit, by expanding the examined probability with respect to the mutually exclusive events of \(\mathbf{F}\) being continuous or discontinuous at a point (or on a curve), and exploiting the continuity of \(\mathbf{F}\) at \(v\) to show that the probability of discontinuity vanishes. The last step relies on the fact that \(\rho(n) \in o(1)\), which implies that the distance between the aforementioned point (or curve) and \(v\) converges in probability to \(0\).

\subsection{Gradient-based Approximation}
We have seen that the divergence of a vector field can be used to compute the curvature. An alternative way to approximate this divergence is through its differential definition, which avoids handling the geometric quantities of Section~\ref{subsec:curvgeom}. More specifically, the unit gradient field can be expressed through its components as \(\mathbf{F} = (F_1,\,F_2)\), so that its divergence is
\begin{equation} \label{eq:divpd}
\divergence\mathbf{F} = \frac{\partial{F_1}}{\partial{x}} + \frac{\partial{F_2}}{\partial{y}}.
\end{equation}

As a result, a second application of the gradient approximation of \eqref{eq:gradgeomapprox}, this time on the components of the approximate unit gradient field \(\mathbf{F}_g = \left(F_{1,g},\,F_{2,g}\right)\), is adequate for calculating the curvature. The full expression for this approximation is
\begin{align} \label{eq:curvgradapprox}
\kappa(v) \approx &\frac{\displaystyle\sum_{i=1}^N \frac{F_{1,g}(w_i)-F_{1,g}(v)}{d(v,w_i)}\cos(\phi(w_i))\Delta\phi(w_i)}{\pi}\nonumber\\
&{+}\:\frac{\displaystyle\sum_{i=1}^N \frac{F_{2,g}(w_i)-F_{2,g}(v)}{d(v,w_i)}\sin(\phi(w_i))\Delta\phi(w_i)}{\pi}.
\end{align}
This gradient-based curvature approximation also converges in probability for RGGs, under slightly stricter conditions than the geometric curvature approximation.

\begin{theorem} \label{th:curvgradapproxconv}
Let \(\mathcal{G}(n,\rho(n))\) be an RGG embedded in \([0,1]^2\), with \(\rho(n) \in \omega\left(n^{-1/2}\right)\,\cap\,o\left(1\right)\) and \(v\) a vertex of \(\mathcal{G}\). Let \(u: \mathbb{R}^2 \rightarrow \mathbb{R}\) be twice differentiable and \(\nabla{u(v)} \neq \mathbf{0}\). Then, the approximation of \eqref{eq:curvgradapprox} converges in probability to \(\kappa(v)\).
\end{theorem}

\begin{IEEEproof}
With the same argument as in the proof of Theorem \ref{th:curvgeomapproxconv}, it can be shown that \[\mathbf{F}_g(v)-\mathbf{F}(v) \xrightarrow{\prob} \mathbf{0}\]
and
\[\mathbf{F}_g(w_i)-\mathbf{F}(w_i) \xrightarrow{\prob} \mathbf{0}\;\forall i \in \{1,\,\dots,\,N\}.\]
These results can be combined into
\[\mathbf{F}_g(w_i)-\mathbf{F}_g(v)-\left(\mathbf{F}(w_i)-\mathbf{F}(v)\right) \xrightarrow{\prob} \mathbf{0}\;\forall{} i \in \{1,\,\dots,\,N\}.\]
Using the above convergence and the product and sum properties of convergence in probability, it follows that
\begin{align*}
&\displaystyle\sum_{i=1}^N \frac{F_{1,g}(w_i)-F_{1,g}(v)}{d(v,w_i)}\,\mathbf{e}_{vw_i}\,\Delta\phi(w_i)\\
&{-}\:\displaystyle\sum_{i=1}^N \frac{F_1(w_i)-F_1(v)}{d(v,w_i)}\,\mathbf{e}_{vw_i}\,\Delta\phi(w_i) \xrightarrow{\prob} \mathbf{0}.
\end{align*}

Furthermore, since \(u\) is twice differentiable and \(\nabla{u(v)} \neq \mathbf{0}\), \(\mathbf{F}\) is differentiable at \(v\), as the quotient of differentiable functions with nonzero denominator. As a result, \(F_1\) is also differentiable at \(v\) and Theorem \ref{th:gradapproxconv} applies:
\[\frac{\displaystyle\sum_{i=1}^N \frac{F_1(w_i)-F_1(v)}{d(v,w_i)}\,\mathbf{e}_{vw_i}\,\Delta\phi(w_i)}{\pi} \xrightarrow{\prob} \nabla{F_1}(v).\]
This result can be combined with the previous one through the sum property of convergence in probability to obtain
\[\frac{\displaystyle\sum_{i=1}^N \frac{F_{1,g}(w_i)-F_{1,g}(v)}{d(v,w_i)}\,\mathbf{e}_{vw_i}\,\Delta\phi(w_i)}{\pi} \xrightarrow{\prob} \nabla{F_1}(v).\]
An identical analysis to the one above leads to the same result for \(F_2\).

The last step of the proof is to isolate from the last convergence results the \(x\)-component of \(\nabla{F_1}(v)\) and the \(y\)-component of \(\nabla{F_2}(v)\), which appear in \eqref{eq:divpd}, and use the sum property of convergence in probability to show that the expression on the right-hand side of \eqref{eq:curvgradapprox} converges in probability to \(\partial{F_1}(v)/\partial{x}+\partial{F_2}(v)/\partial{y}\).
\end{IEEEproof}

\section{Smoothing Filtering on Graphs} \label{sec:smoothing}
In the last two sections, we developed certain approximations to compute quantities that are essential for the update equation of the GAC model on arbitrary graphs. Despite the convergence of these approximations to the true values of the quantities in the limit of large RGGs, there is a non-negligible error in practice, due to the discrete nature of the approximations. This error is propagated to the embedding function after each update and therefore it may be accumulated after several iterations in a way that leads to instabilities.

Consequently, it is very beneficial to apply smoothing filtering across the graph, in order to increase the robustness of approximations against spatial non-uniformities in vertex locations. Moreover, smoothing filtering is necessary in the initialization stage of the GAC algorithm, where the original intensity function is simplified to distinguish predominant ``edges'' from small-scale variations. Of course, filtering functions across the graph induces a considerable computational burden, especially when done at every iteration, but we are willing to trade a little speed for stability.

\subsection{Neighborhood-based Average/Median Filtering}
To compute a smoother version of a function defined on a graph, one option is to operate in the same neighborhood-based framework that we presented in the previous sections, and apply a simple filter on the original version of the function. This filter can be either an average or a median filter, receiving as input the set of function values at the vertex itself and all its neighbors. In the curvature and embedding function cases this is straightforward, while for the gradient, we filter each of the two vector dimensions separately.

To validate the benign effects of smoothing filtering of both the gradient and curvature component of the GAC model, we experiment with certain analytical functions defined on RGGs. For each graph, we compute the function's gradient and the curvature of its level sets using the proposed approximations and afterwards we filter the results with an average or median filter. Deriving the analytical expressions of the function's gradient and curvature, we are able to compare them with our estimates. To enable a quantitative assessment of our approximations and smoothing filters, we define a suitable error metric which we call \emph{relative error} and denote by \(e_r\). If the approximation error at each vertex is defined as the difference between the approximate value of the function at that vertex and its true, analytical value, then the relative error on the whole graph is simply the ratio of the error's energy and the function's energy:
\begin{equation} \label{eq:relerror}
e_r = \frac{E_{\text{error}}}{E_{\text{analytical}}}.
\end{equation}

\begin{figure*}
  \centering
  \subfloat[]{\includegraphics[width=\columnwidth]{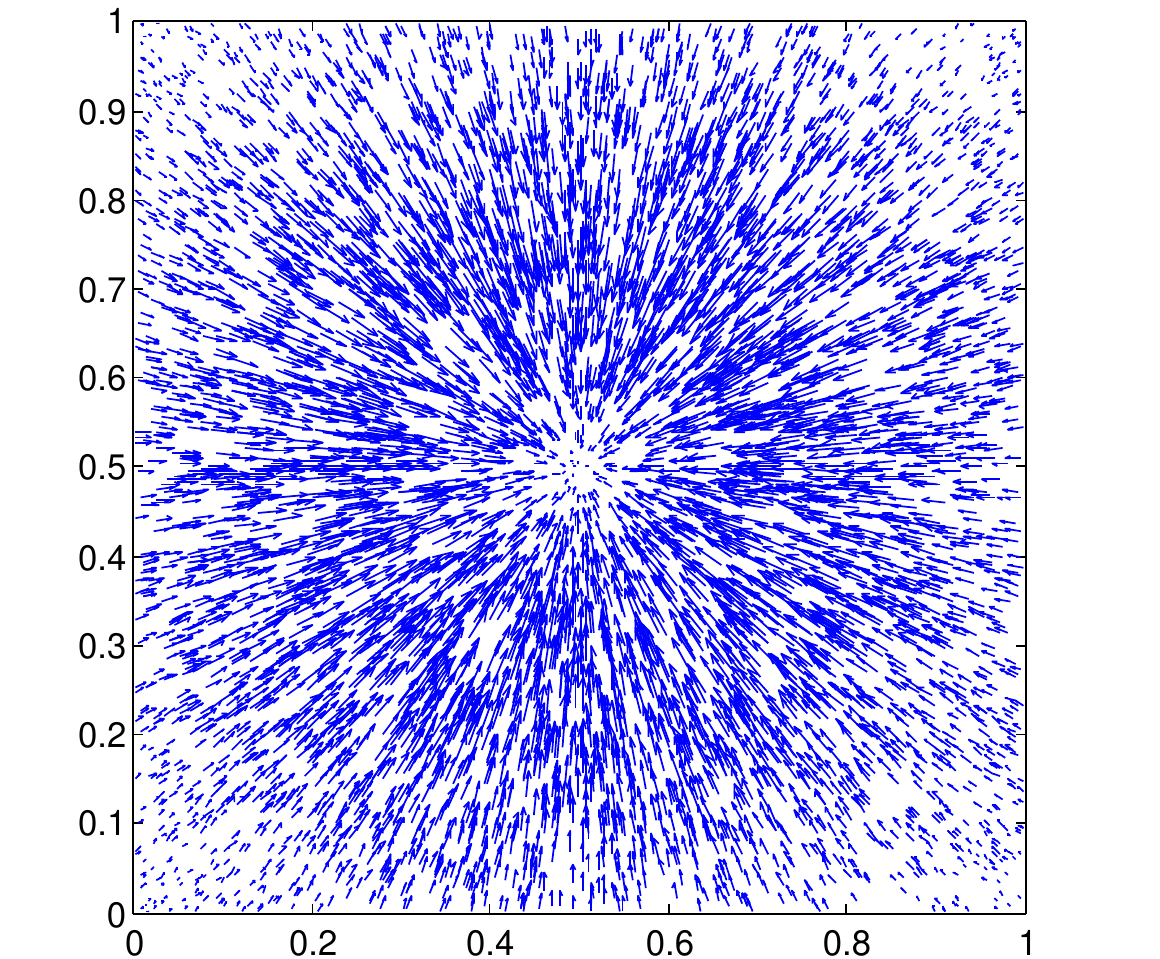}
  \label{subfig:Gaussgradtrue}}
  \hfil
  \subfloat[]{\includegraphics[width=\columnwidth]{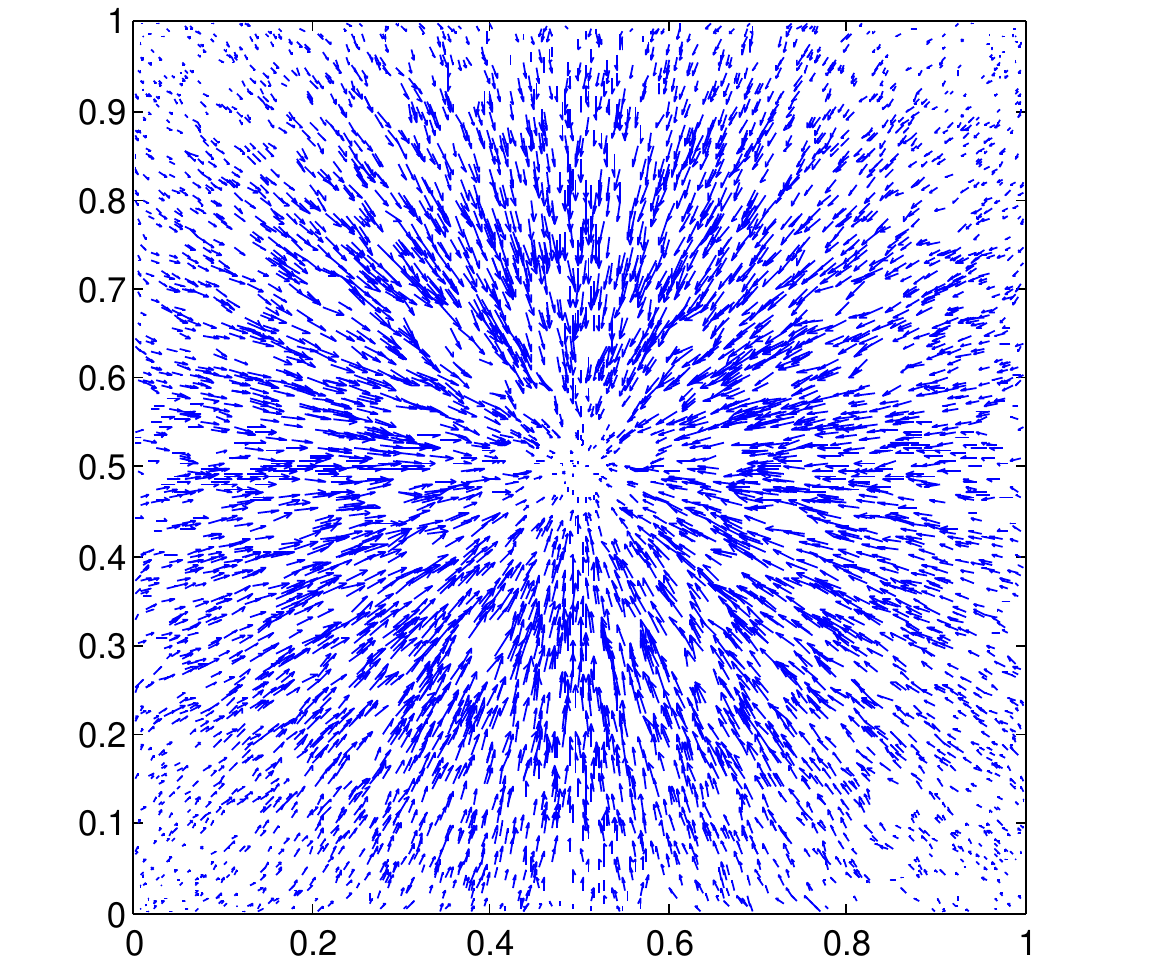}
  \label{subfig:Gaussgradplaingeom}}\\
  \subfloat[]{\includegraphics[width=\columnwidth]{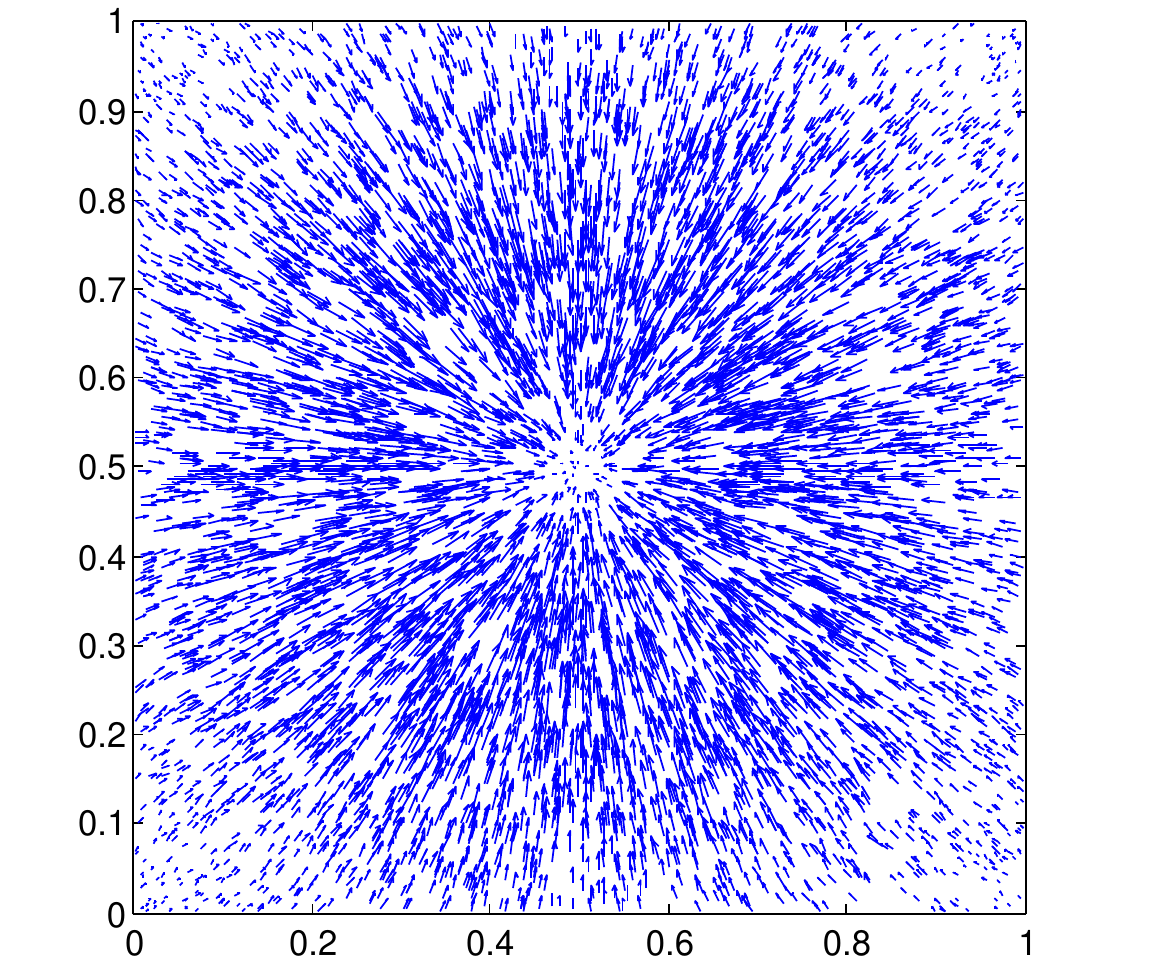}
  \label{subfig:Gaussgradaverage}}
  \hfil
  \subfloat[]{\includegraphics[width=\columnwidth]{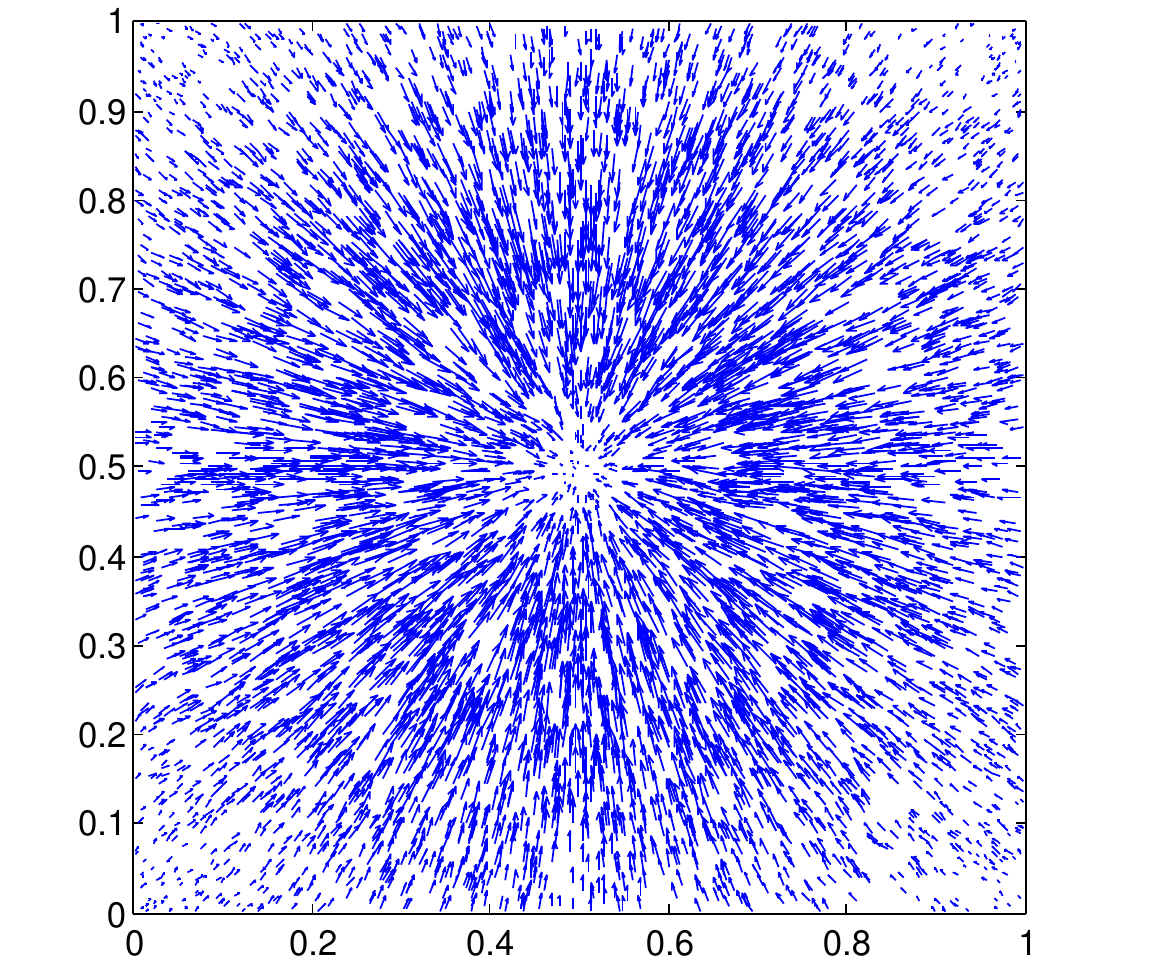}
  \label{subfig:Gaussgradmedian}}
  \caption{Comparison of gradient approximations for a Gaussian function defined on an RGG. The true gradient field is shown in \protect\subref{subfig:Gaussgradtrue}. The result of the geometric approximation is shown in \protect\subref{subfig:Gaussgradplaingeom}. In \protect\subref{subfig:Gaussgradaverage} and \protect\subref{subfig:Gaussgradmedian}, average and median filtering have been used respectively to improve \protect\subref{subfig:Gaussgradplaingeom}. All vector fields have been scaled by a factor of 2 to aid visualization.}
  \label{fig:Gaussgradcomparison}
\end{figure*}

In the experiments that follow in the rest of the paper, the radius of an RGG is computed as
\begin{equation} \label{eq:radius}
\rho(n) = Cn^{-1/3},
\end{equation}
in order to achieve the strictest asymptotic bound for gradient approximation error according to the results of Section~\ref{sec:gradapprox}. Unless otherwise specified, we fix \(C = 0.6\) in the rest of the paper. In the rest of this section, all RGGs are embedded in \([0,1]^2\). In Fig. \ref{fig:Gaussgradcomparison}, we show the results of gradient approximation for an isotropic Gaussian on an RGG with \(n = 5500\) vertices. The analytical form of the Gaussian is
\begin{equation} \label{eq:Gaussian}
\exp\left(-\left((x-x_0)^2+(y-y_0)^2\right)\,/\,2\sigma^2\right),
\end{equation}
with \(\sigma = 0.25\) and \(x_0 = y_0 = 0.5\). The small, local deviations of the geometric approximation (Fig. \ref{fig:Gaussgradcomparison}\subref{subfig:Gaussgradplaingeom}) from the true gradient vector field (Fig. \ref{fig:Gaussgradcomparison}\subref{subfig:Gaussgradtrue}) are almost completely smoothed out with average or median filtering (Fig. \ref{fig:Gaussgradcomparison}\subref{subfig:Gaussgradaverage} and \ref{fig:Gaussgradcomparison}\subref{subfig:Gaussgradmedian} respectively).

\begin{figure}
  \centering
  \includegraphics[width=\columnwidth]{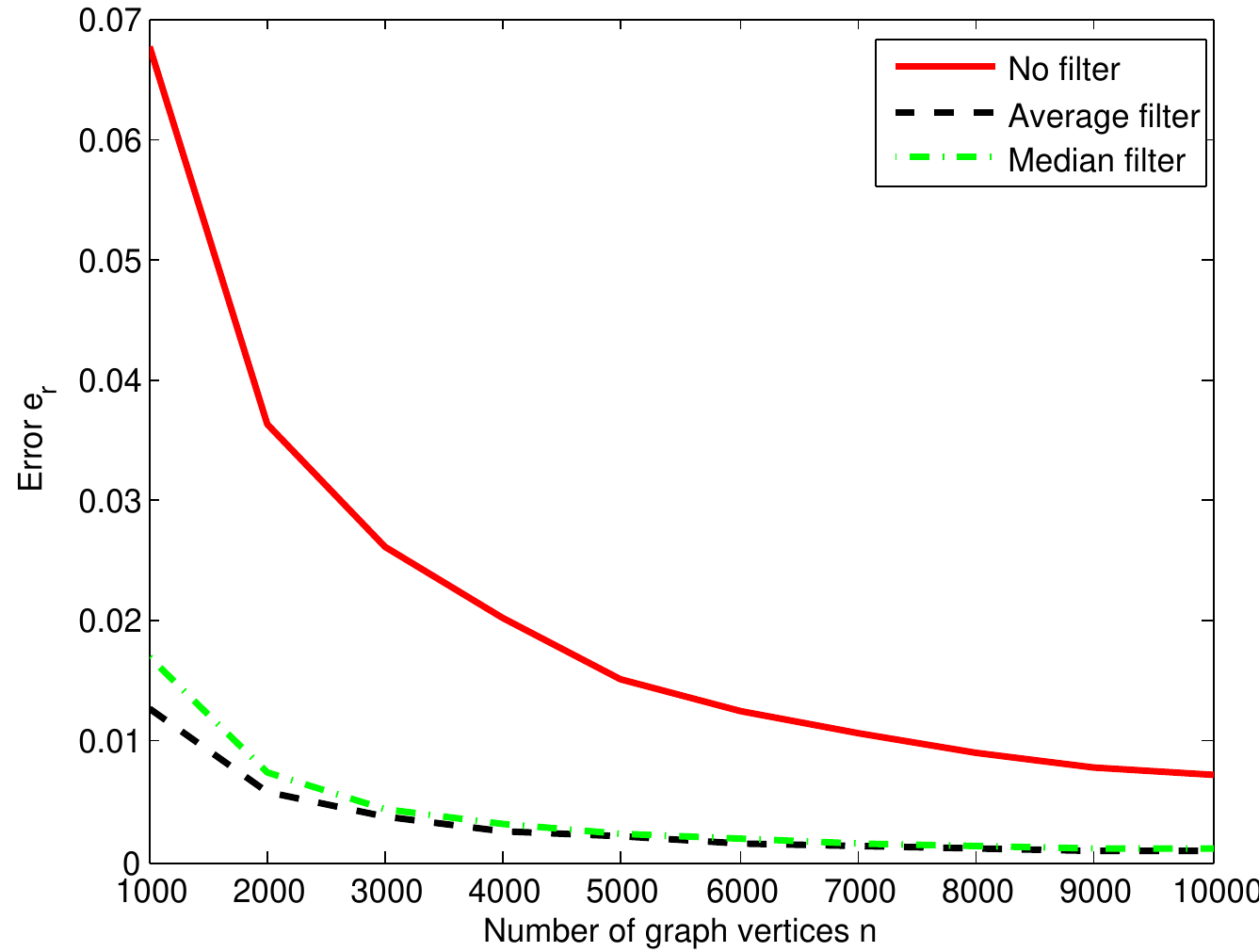}
  \caption{Relative error of gradient approximations for a Gaussian function defined on RGGs of increasing size. The geometric approximation with no filtering is compared to its filtered versions with an average or median filter.}
  \label{fig:Gaussgradrelerrorcomparison}
\end{figure}

To verify the enhancement of gradient approximation with smoothing filtering quantitatively, we evaluate the relative error for the Gaussian function in \eqref{eq:Gaussian} on RGGs whose size ranges from 1000 to 10000 vertices. Fig. \ref{fig:Gaussgradrelerrorcomparison} shows average values of \(e_r\) over 10 different graphs for each size, which leads to a reduced variance in the estimation. Using either smoothing filter reduces relative error substantially irrespective of size. This leads us to apply smoothing on gradient and feed the smoothed version to curvature computation.

\begin{figure*}
  \centering
  \subfloat[]{\includegraphics[width=0.66\columnwidth]{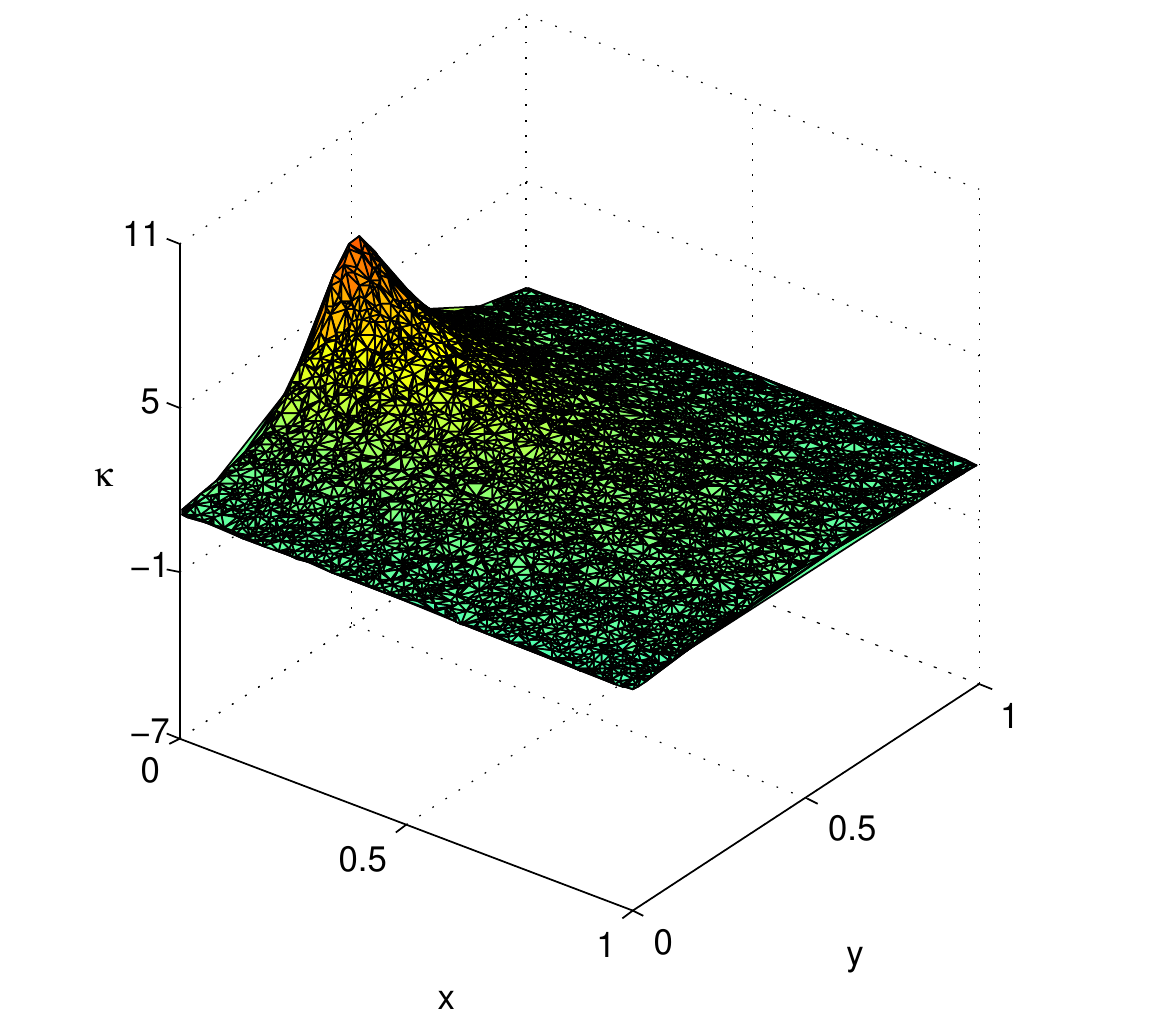}
  \label{subfig:coniccurvtrue}}
  \hfil
  \subfloat[]{\includegraphics[width=0.66\columnwidth]{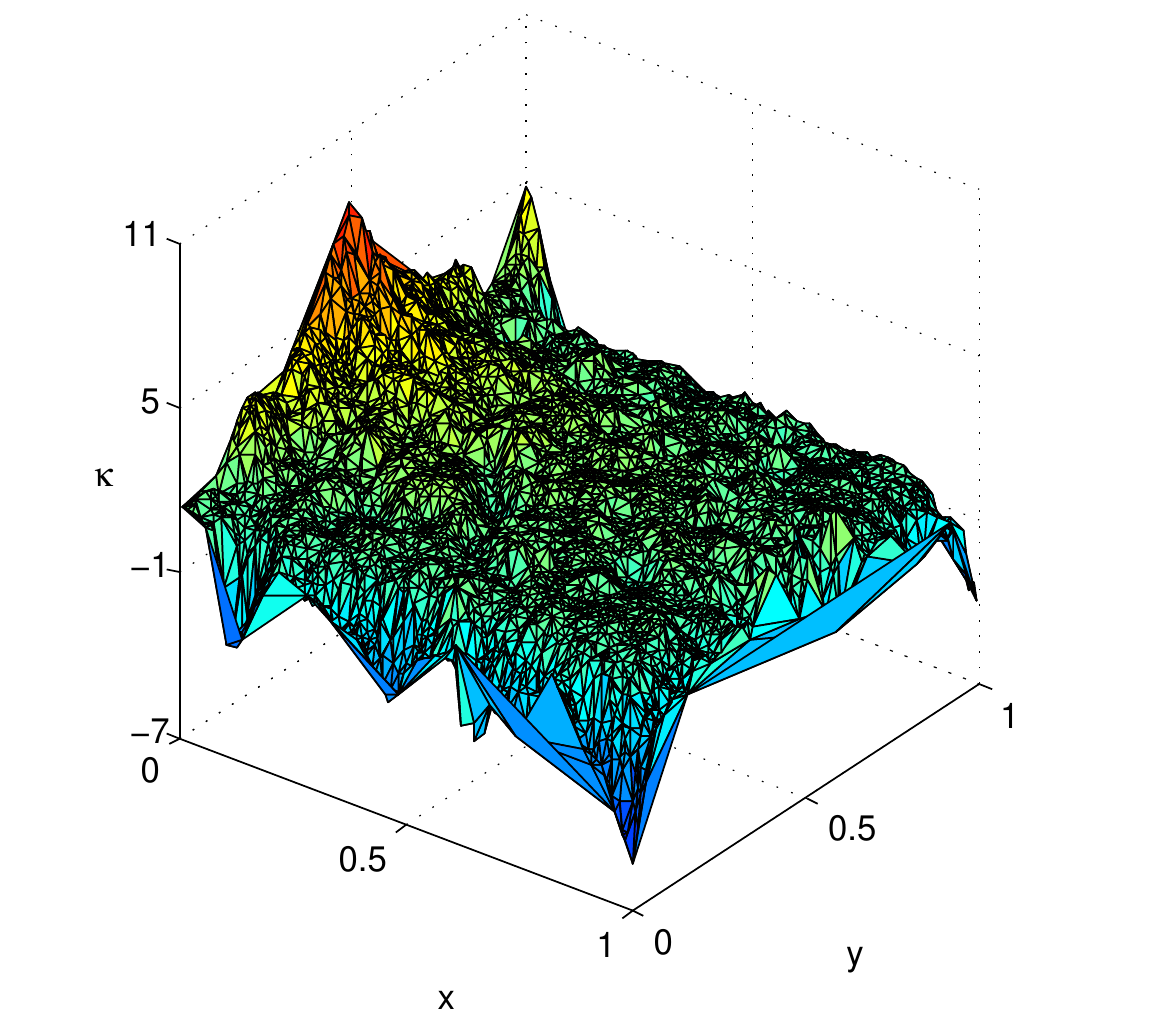}
  \label{subfig:coniccurvgeomaverage}}
  \hfil
  \subfloat[]{\includegraphics[width=0.66\columnwidth]{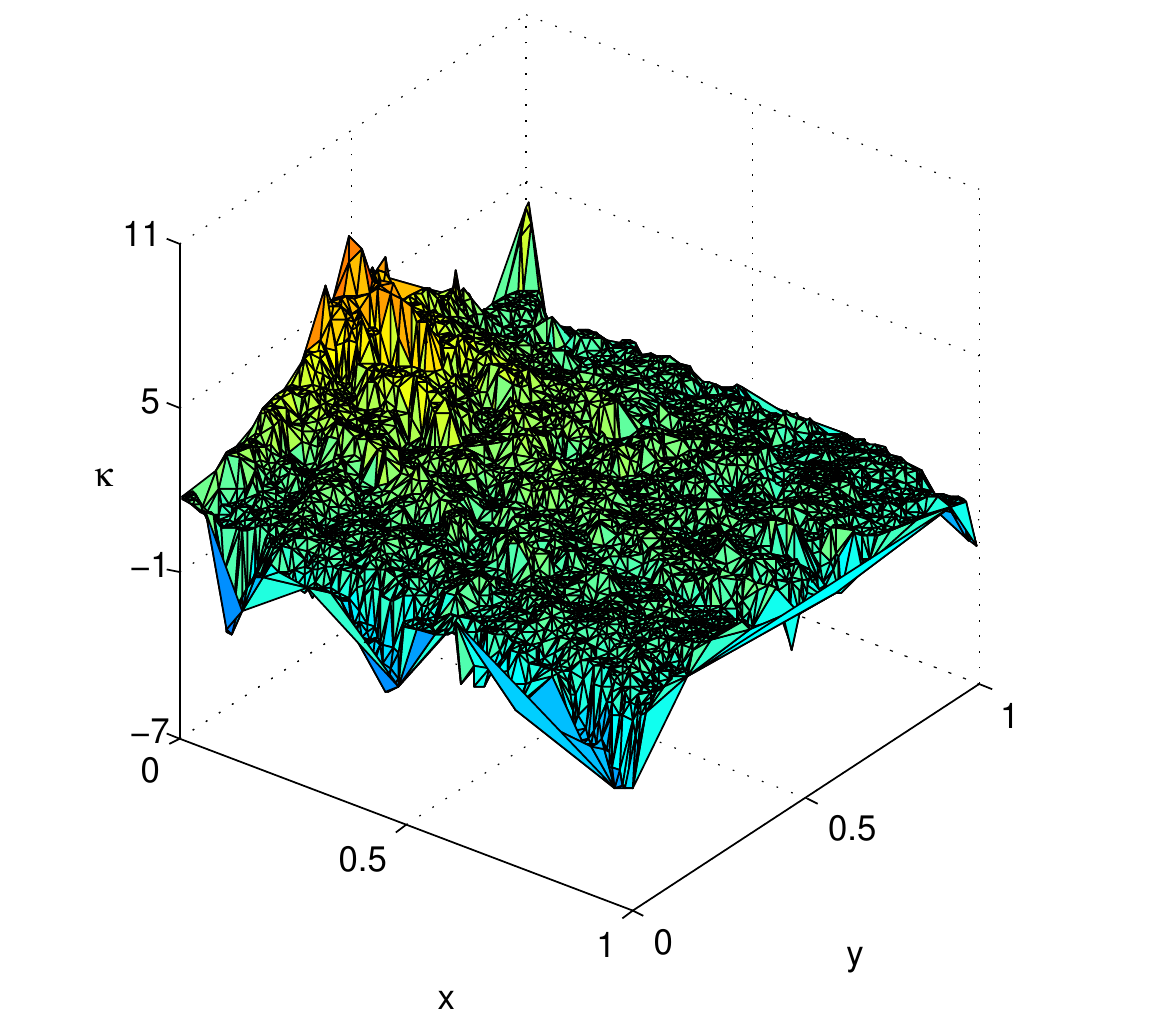}
  \label{subfig:coniccurvgeommedian}}
  \\
  \subfloat[]{\includegraphics[width=0.66\columnwidth]{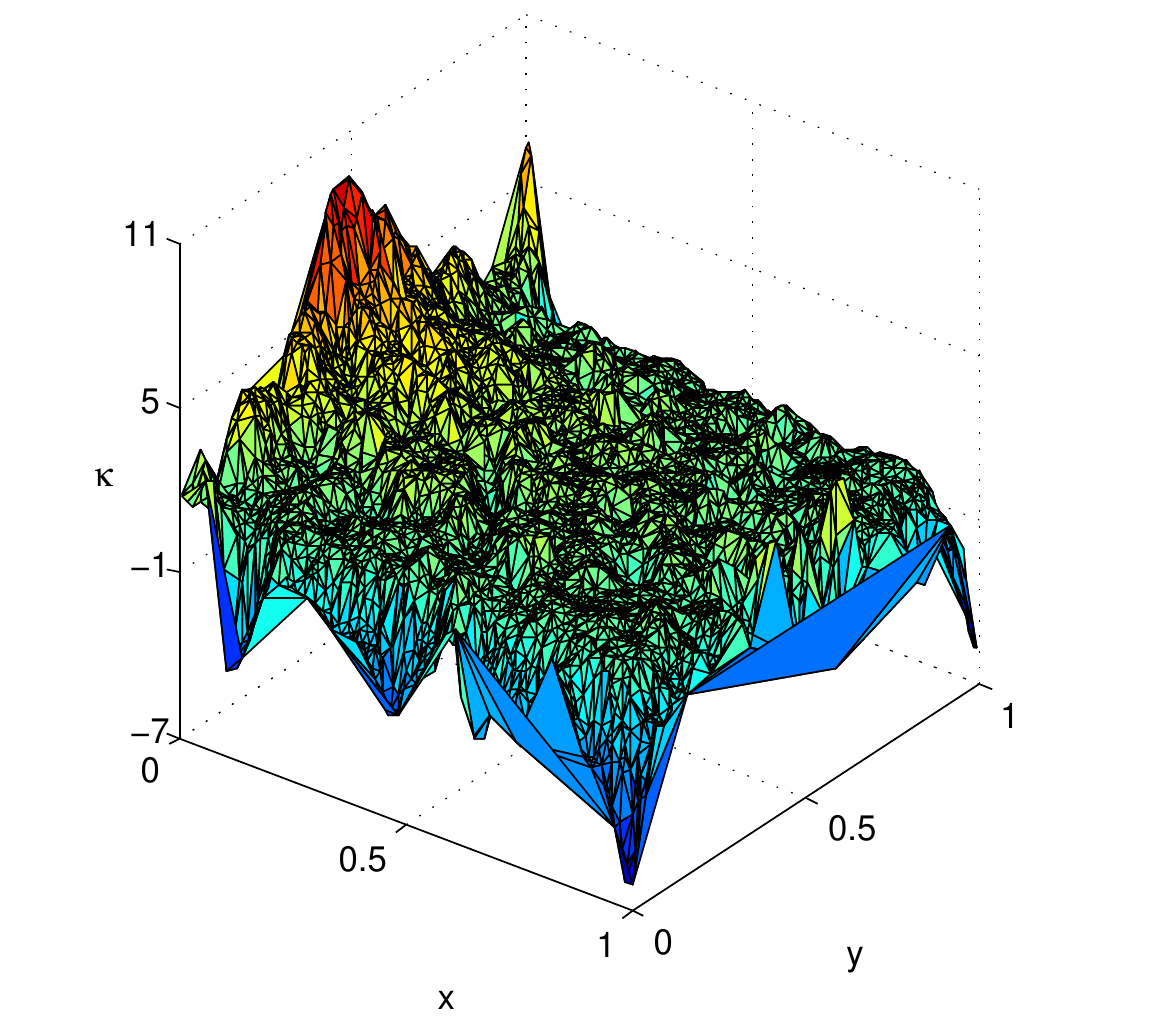}
  \label{subfig:coniccurvgradbasedaverage}}
  \hfil
  \subfloat[]{\includegraphics[width=0.66\columnwidth]{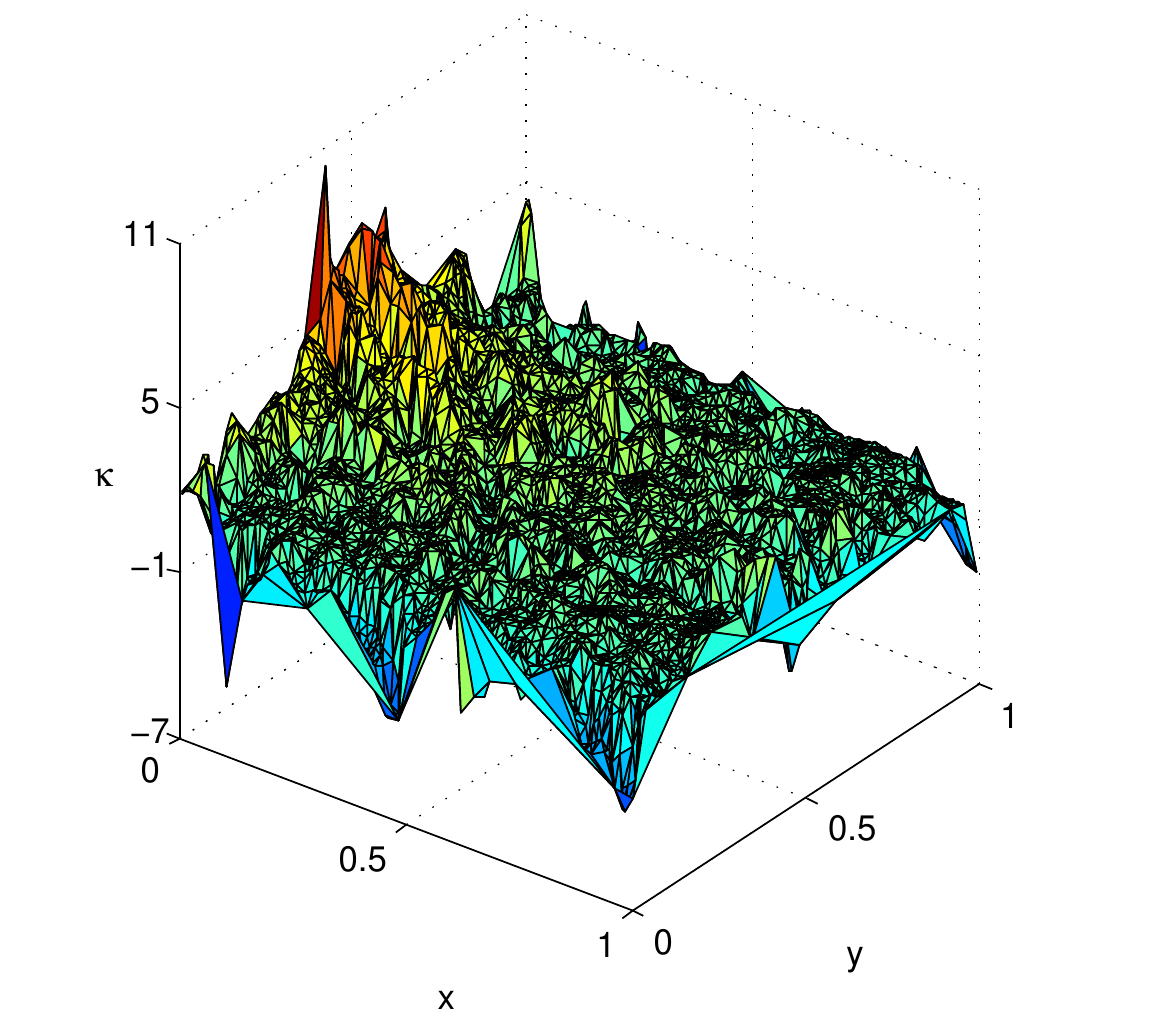}
  \label{subfig:coniccurvgradbasedmedian}}
  \caption{Comparison of curvature approximations for a conic function defined on an RGG. The true curvature is shown in \protect\subref{subfig:coniccurvtrue}. \protect\subref{subfig:coniccurvgeomaverage} and \protect\subref{subfig:coniccurvgeommedian} show the geometric approximation smoothed with average and median filtering respectively, while \protect\subref{subfig:coniccurvgradbasedaverage} and \protect\subref{subfig:coniccurvgradbasedmedian} show the gradient-based approximation smoothed with average and median filtering respectively. In each case, the smoothing filter used for curvature has also been applied to the input gradient values.}
  \label{fig:coniccurvcomparison}
\end{figure*}

Fig. \ref{fig:coniccurvcomparison} presents a comparison of different curvature approximations for a function whose graph corresponds to an elliptical cone, which we will call conic function for short. The form of this conic function is
\begin{equation} \label{eq:conic}
\sqrt{\frac{(x-x_0)^2}{\alpha^2} + \frac{(y-y_0)^2}{\beta^2}},
\end{equation}
where \(\alpha = 0.4\), \(\beta = 0.3\), \(x_0 = -0.25\) and \(y_0 = 0.5\) for Fig. \ref{fig:coniccurvcomparison}. The underlying graph is an RGG with 7000 vertices. Even though all approximations are smoothed, they demonstrate strong, abrupt variations from the true curvature values. This degradation relative to the gradient case is explained by the cascaded nature of curvature approximation. Nonetheless, the overall curvature trend of the conic function is captured well by all approaches.

\begin{figure}
  \centering
  \subfloat[]{\includegraphics[width=0.98\columnwidth]{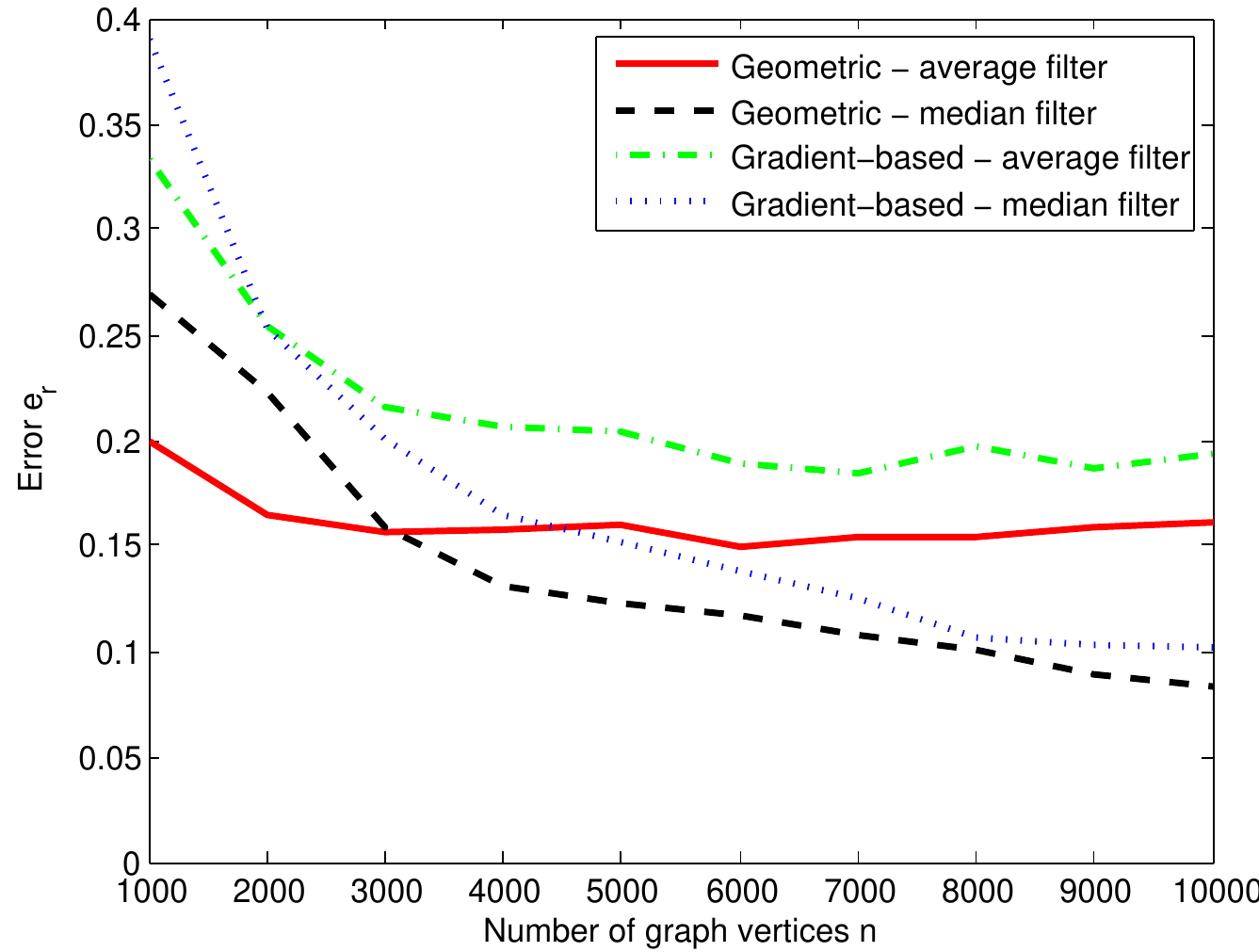}
  \label{subfig:coniccurvrelerrorcomparisonnolocalextremum}}
  \\
  \subfloat[]{\includegraphics[width=0.98\columnwidth]{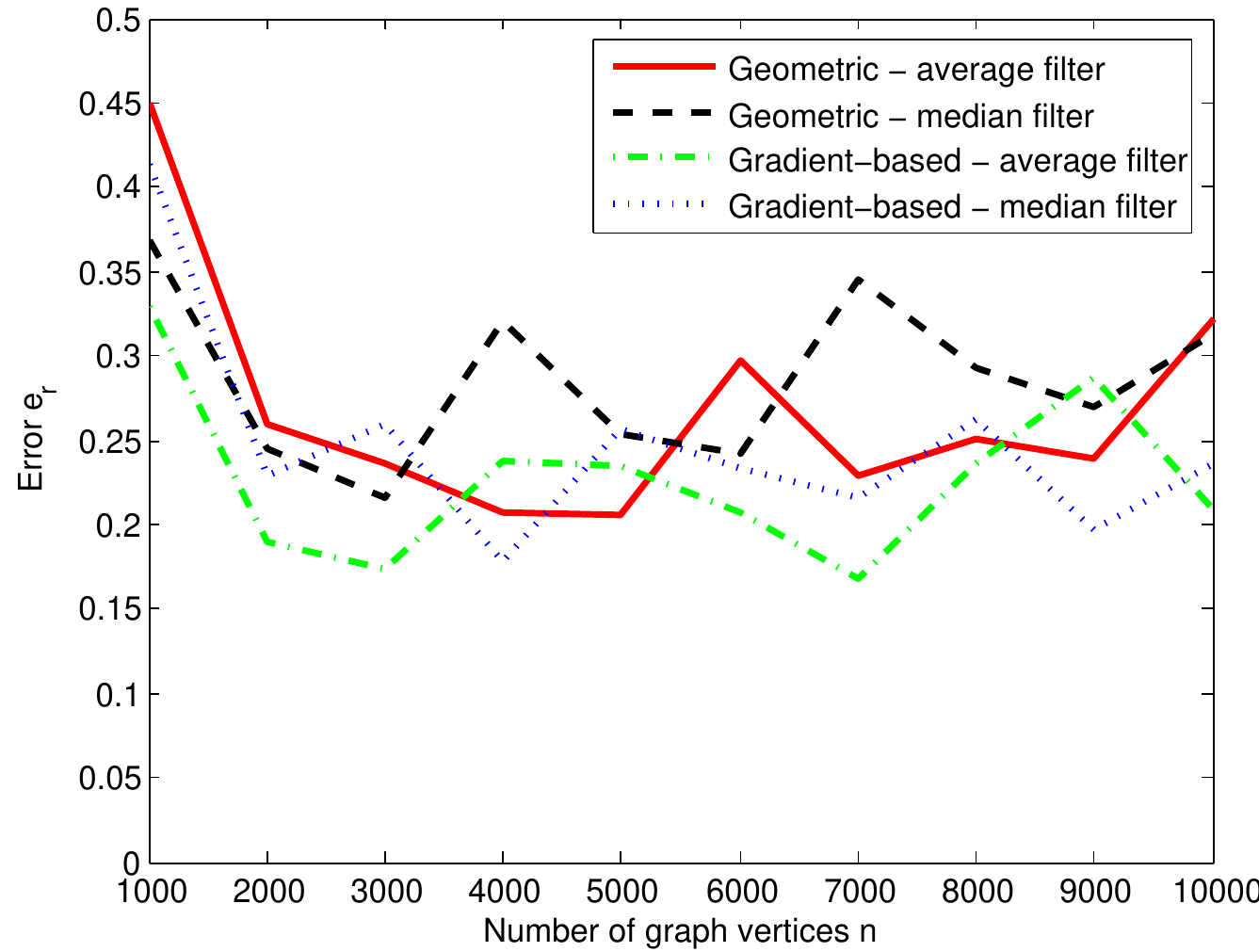}
  \label{subfig:coniccurvrelerrorcomparisonlocalextremum}}
  \caption{Relative error of curvature approximations for a conic function defined on RGGs of increasing size. In \protect\subref{subfig:coniccurvrelerrorcomparisonnolocalextremum}, the conic function does not assume a local extremum in the interior of the graph's region, whereas in \protect\subref{subfig:coniccurvrelerrorcomparisonlocalextremum} it does. All four combinations of type of approximation (geometric or gradient-based) and smoothing filter (average or median) are compared in both cases.}
  \label{fig:coniccurvrelerrorcomparison}
\end{figure}

In Fig. \ref{fig:coniccurvrelerrorcomparison}\subref{subfig:coniccurvrelerrorcomparisonnolocalextremum}, we present the results of an experiment similar to the one in Fig. \ref{fig:Gaussgradrelerrorcomparison}, this time focusing on the curvature of the conic function in \eqref{eq:conic}. The parameters of the function are the same as before. Median filtering appears superior: the median-filtered approximations exhibit lower error than the corresponding average-filtered ones over almost the entire range of graph sizes (except for the smaller sizes). Even more importantly, the relative error of median-filtered curvature is steadily decreasing for increasing graph size both with the geometric and the gradient-based approximation, in contrast to the average-filtered cases, where the error stops decreasing around 4000 vertices. Due to these facts, we use median filtering for smoothing gradient and curvature for the GAC model in the rest of the paper. We also observe that the geometric approximation induces a relatively smaller error than the gradient-based approximation, when the same filter is used.

To emphasize on the importance of the assumptions made for convergence of the curvature approximations in Section~\ref{sec:curvapprox}, we repeat the last experiment setting \(x_0 = 0.5\). This way, the conic function is not differentiable at \((0.5,0.5)\), which lies in the interior of the graphs' region, and therefore the assumptions of Theorems \ref{th:curvgeomapproxconv} and \ref{th:curvgradapproxconv} do not hold necessarily for every vertex of the graphs. In fact, near this point, the true curvature of the level sets approaches infinity. Indeed, the evolution of relative error depicted in Fig. \ref{fig:coniccurvrelerrorcomparison}\subref{subfig:coniccurvrelerrorcomparisonlocalextremum} confirms that all approximations are less accurate and they do not converge in this case.

\subsection{Gaussian Smoothing}

At the initialization stage of the GAC algorithm, one of the tasks is to process the original image function \(I: \mathcal{V} \rightarrow \mathbb{R}^+\) so as to obtain a smoother version of it. This way, the stopping function \(g\) can be computed subsequently, encoding only the predominant edges and ignoring local variations on the graph.

Following \cite{DrMa12}, we employ Gaussian smoothing defined on graphs for this task. The filter is an isotropic 2D Gaussian with standard deviation \(\sigma\):
\begin{equation} \label{eq:Gaussianfilter}
G_{\sigma}(\mathbf{x}) = \frac{1}{2\pi\sigma^2}\exp\left(-\frac{\vecnorm{\mathbf{x}}^2}{2\sigma^2}\right).
\end{equation}
We will denote the smoothed image which is obtained by using such a filter with \(I_{\sigma}\). The authors in \cite{DrMa12} use a simple graph-based convolution of \eqref{eq:Gaussianfilter} with the image to perform smoothing:
\begin{equation} \label{eq:simpleGaussianfiltering}
I_{\sigma}(\mathbf{v}) = \sum_{\mathbf{w} \in \mathcal{V}} I(\mathbf{w}) G_{\sigma}(\mathbf{v} - \mathbf{w}).
\end{equation}
A simple formula is then used to calculate the stopping function \(g\):
\begin{equation} \label{eq:gfunction}
g(\vecnorm{\nabla{I_{\sigma}}}) = \frac{1}{1+\frac{{\vecnorm{\nabla{I_{\sigma}}}}^2}{\lambda^2}}.
\end{equation}

\begin{figure*}
  \centering
  \subfloat[]{\includegraphics[width=0.66\columnwidth]{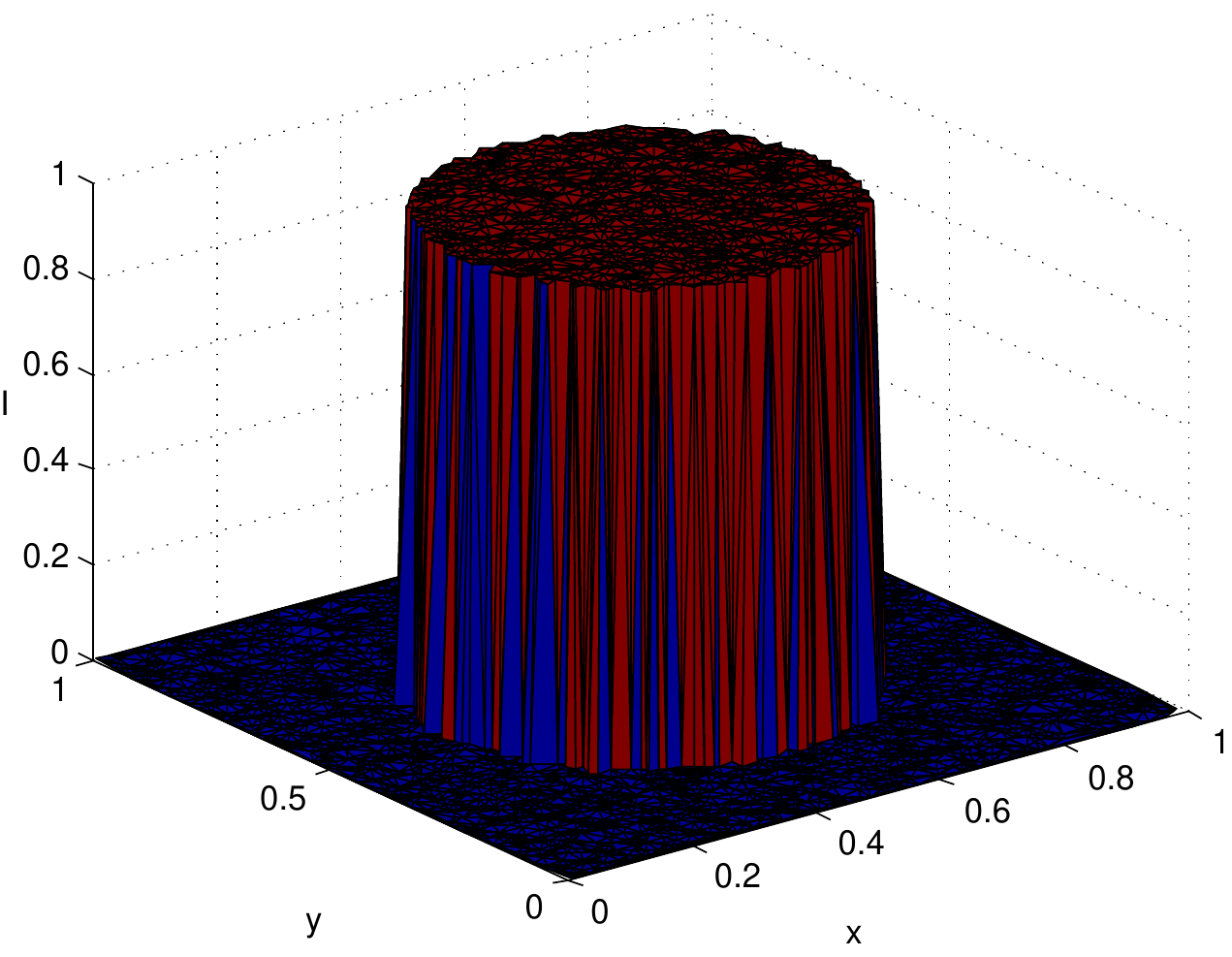}
  \label{subfig:gfunI}}
  \hfil
  \subfloat[]{\includegraphics[width=0.66\columnwidth]{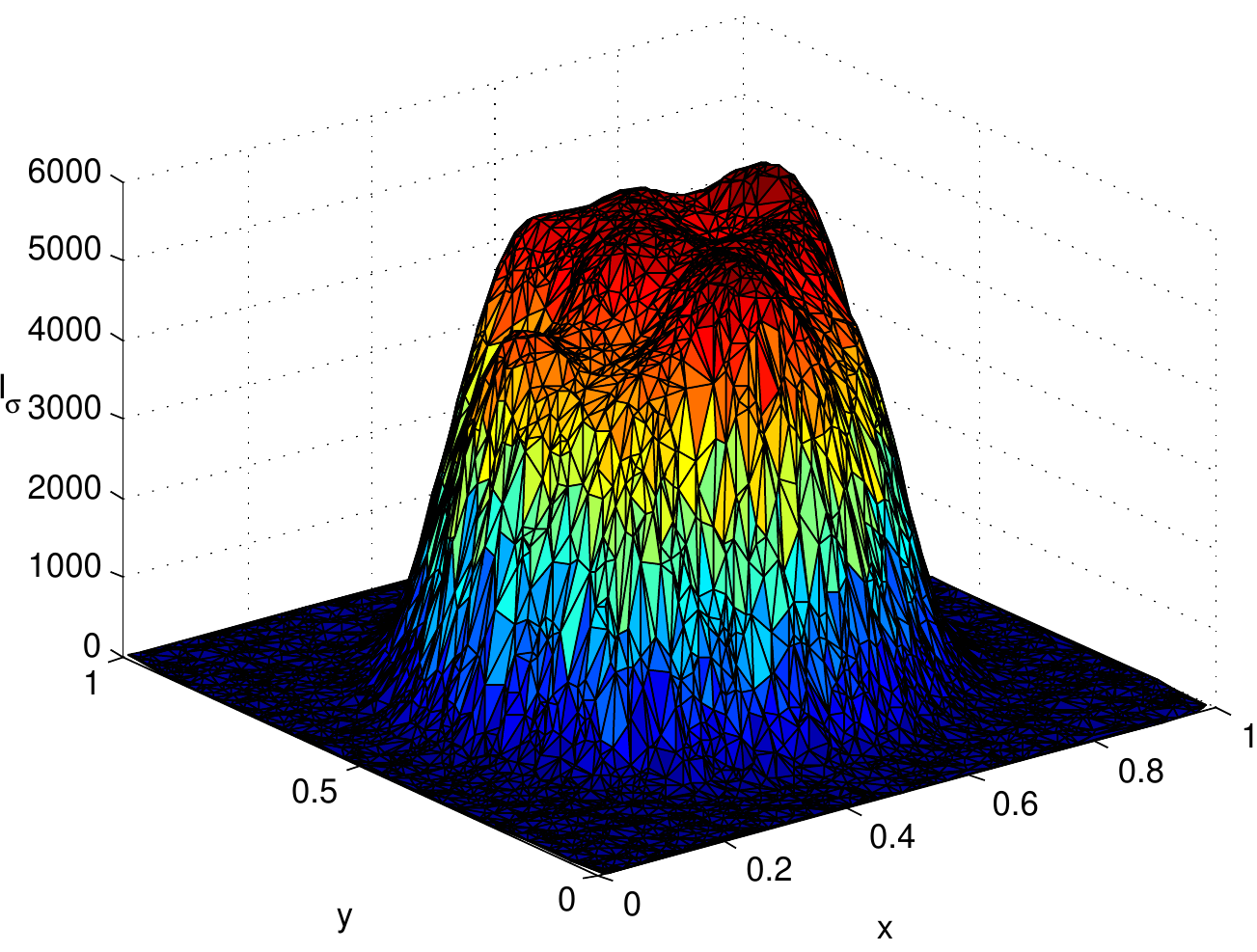}
  \label{subfig:gfunIsigmasimple}}
  \hfil
  \subfloat[]{\includegraphics[width=0.66\columnwidth]{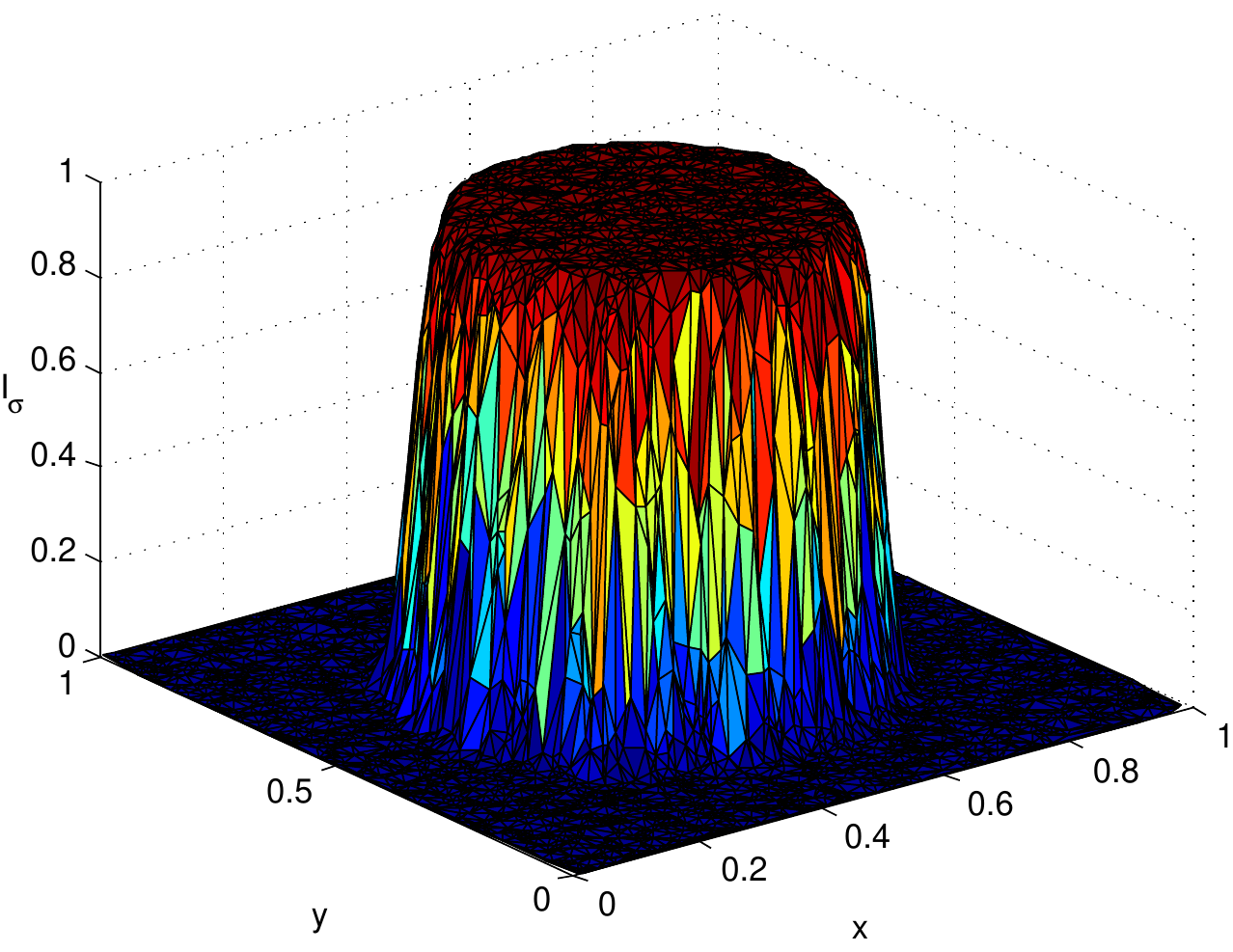}
  \label{subfig:gfunIsigmanorm}}
  \\
  \subfloat[]{\includegraphics[width=0.66\columnwidth]{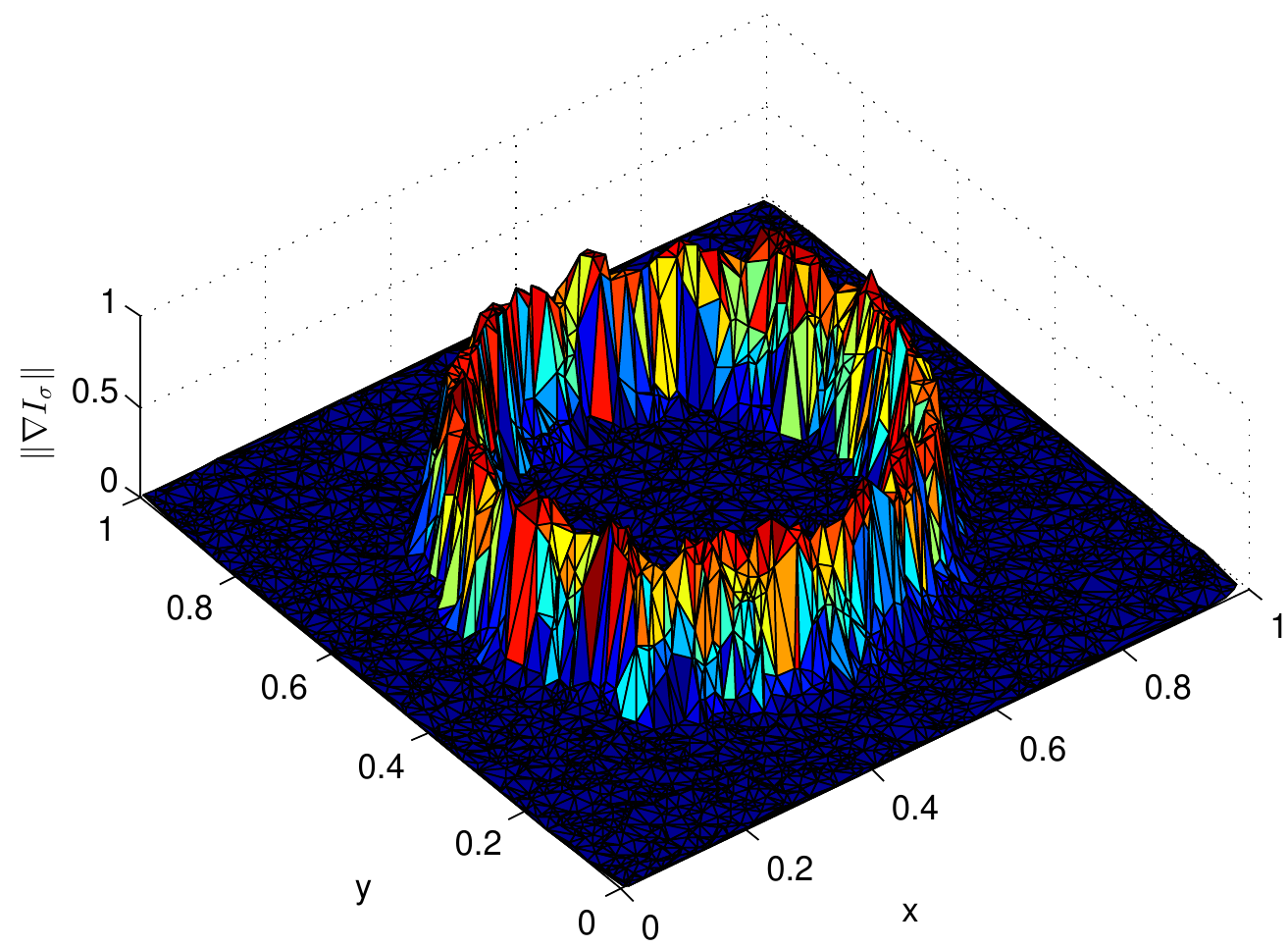}
  \label{subfig:gfunIsigmagradmagnnormderiv}}
  \hfil
  \subfloat[]{\includegraphics[width=0.66\columnwidth]{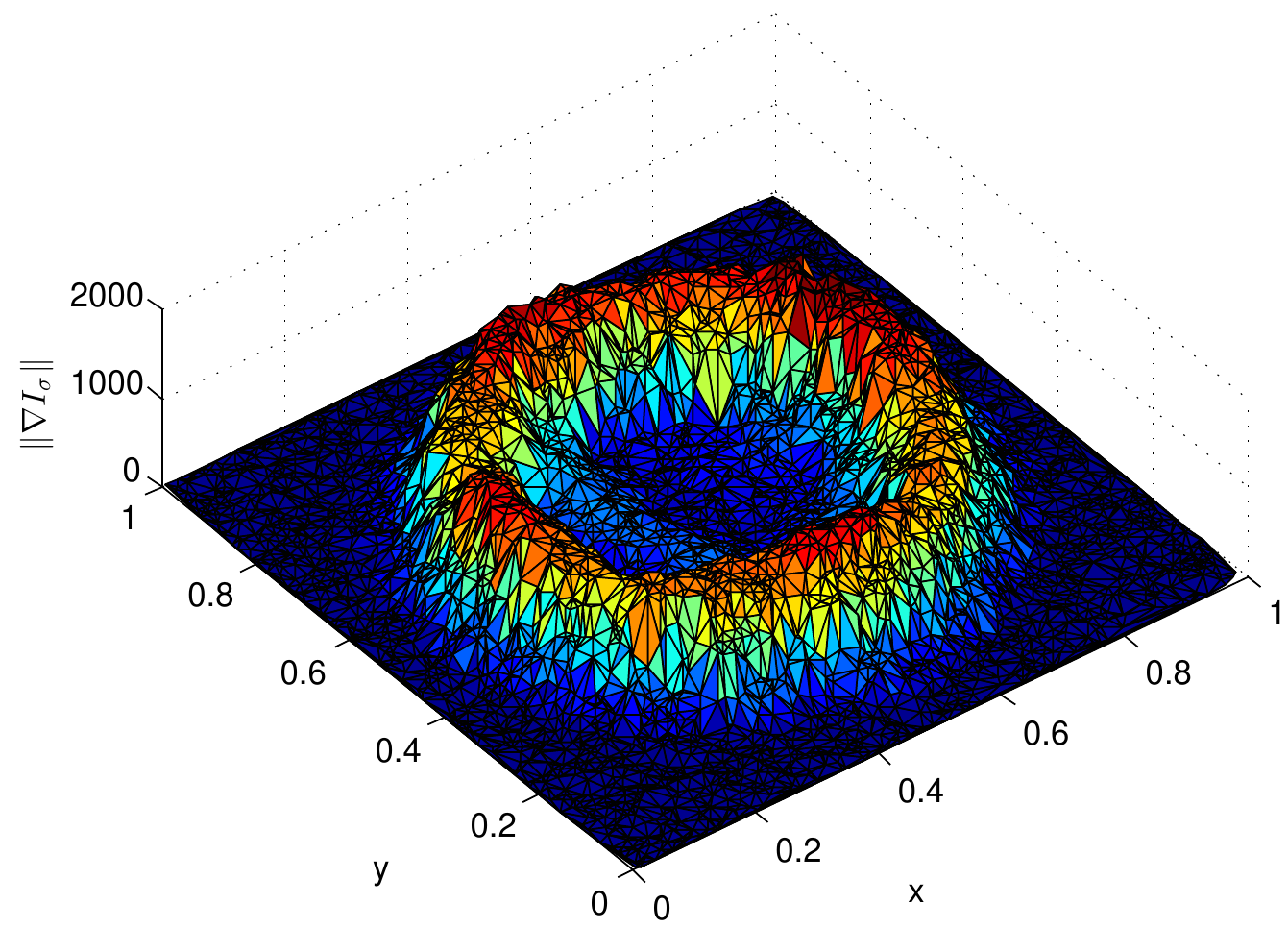}
  \label{subfig:gfunIsigmagradmagnsimple}}
  \hfil
  \subfloat[]{\includegraphics[width=0.66\columnwidth]{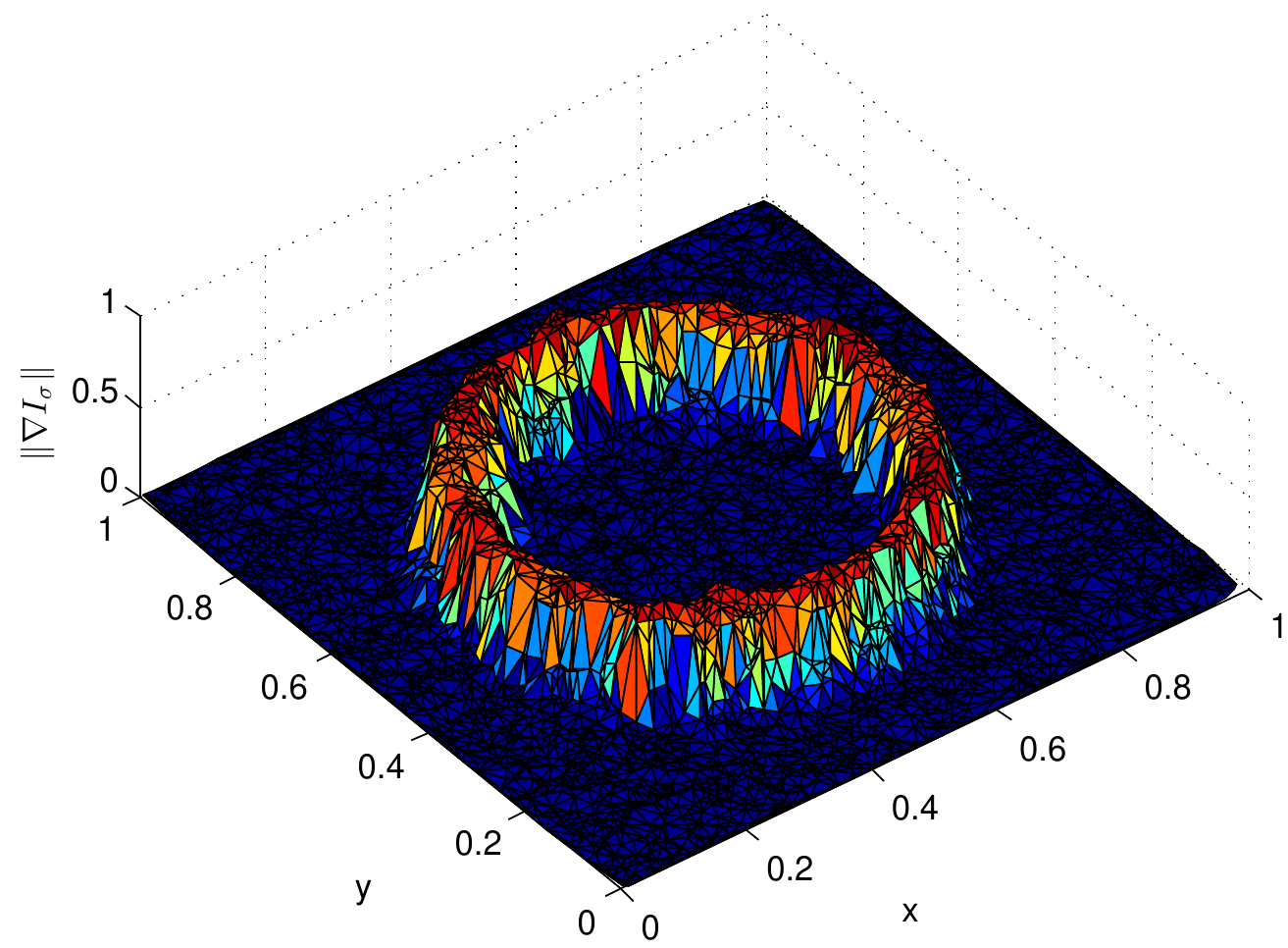}
  \label{subfig:gfunIsigmagradmagnnorm}}
  \\
  \subfloat[]{\includegraphics[width=0.66\columnwidth]{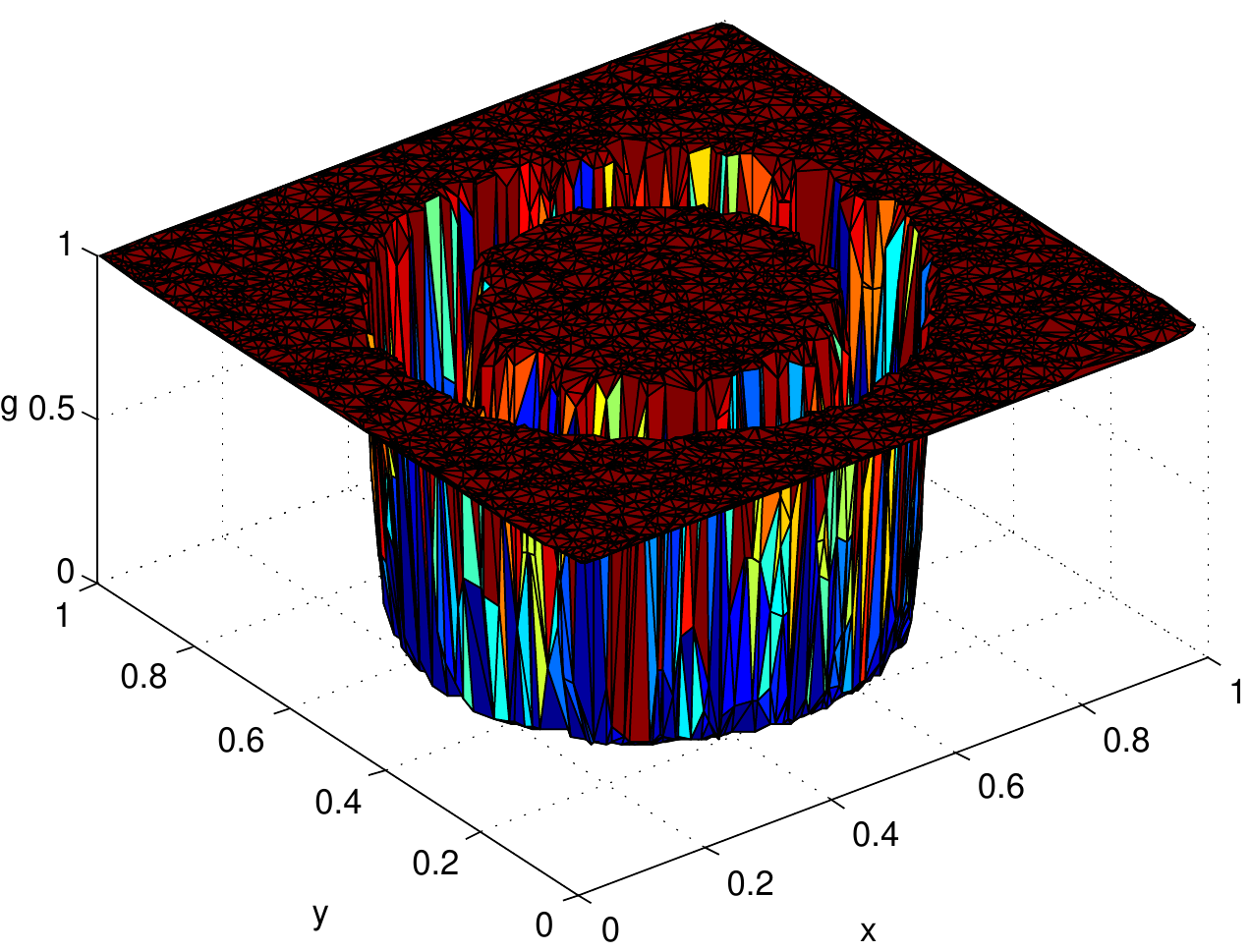}
  \label{subfig:gfunresultnormderiv}}
  \hfil
  \subfloat[]{\includegraphics[width=0.66\columnwidth]{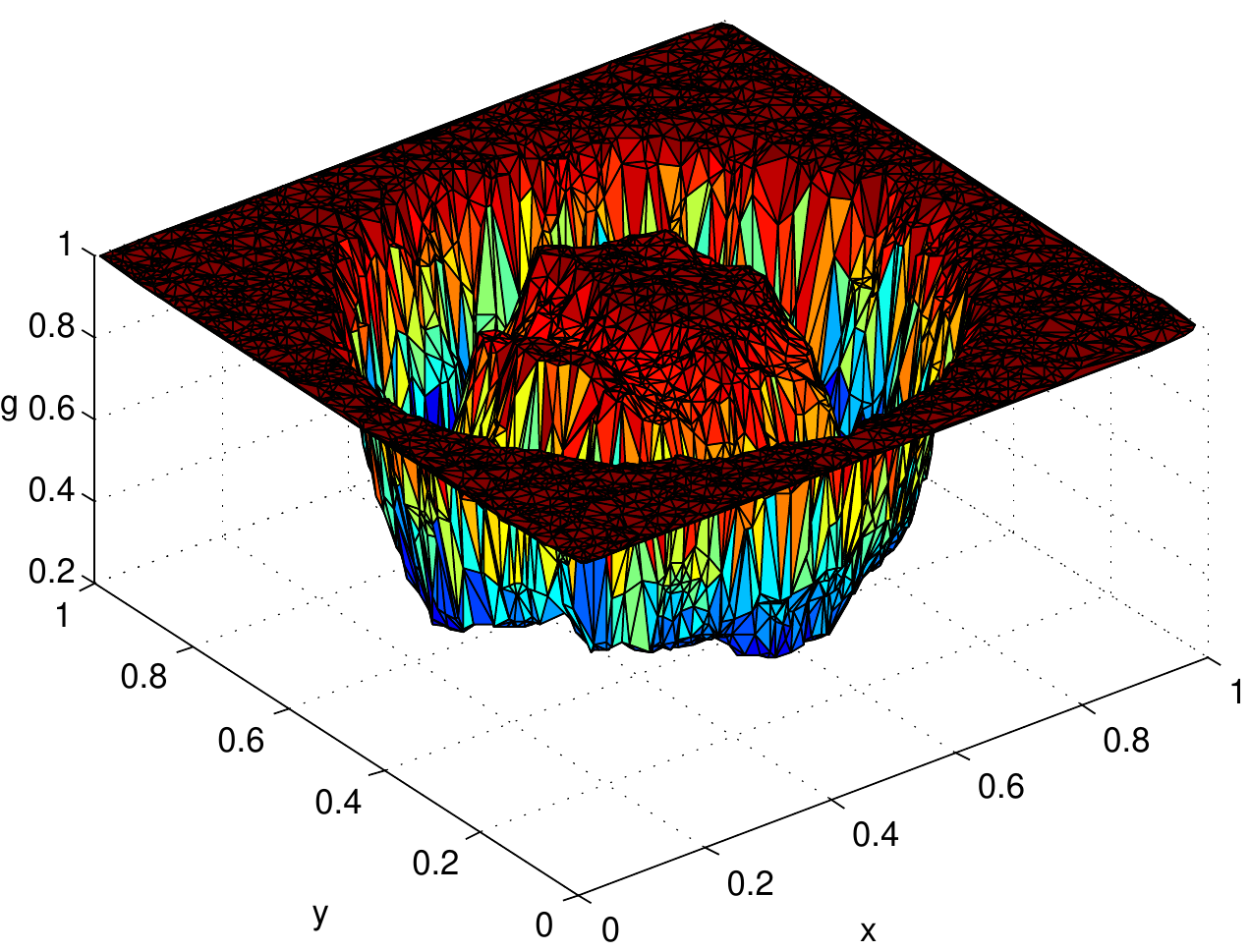}
  \label{subfig:gfunresultsimple}}
  \hfil
  \subfloat[]{\includegraphics[width=0.66\columnwidth]{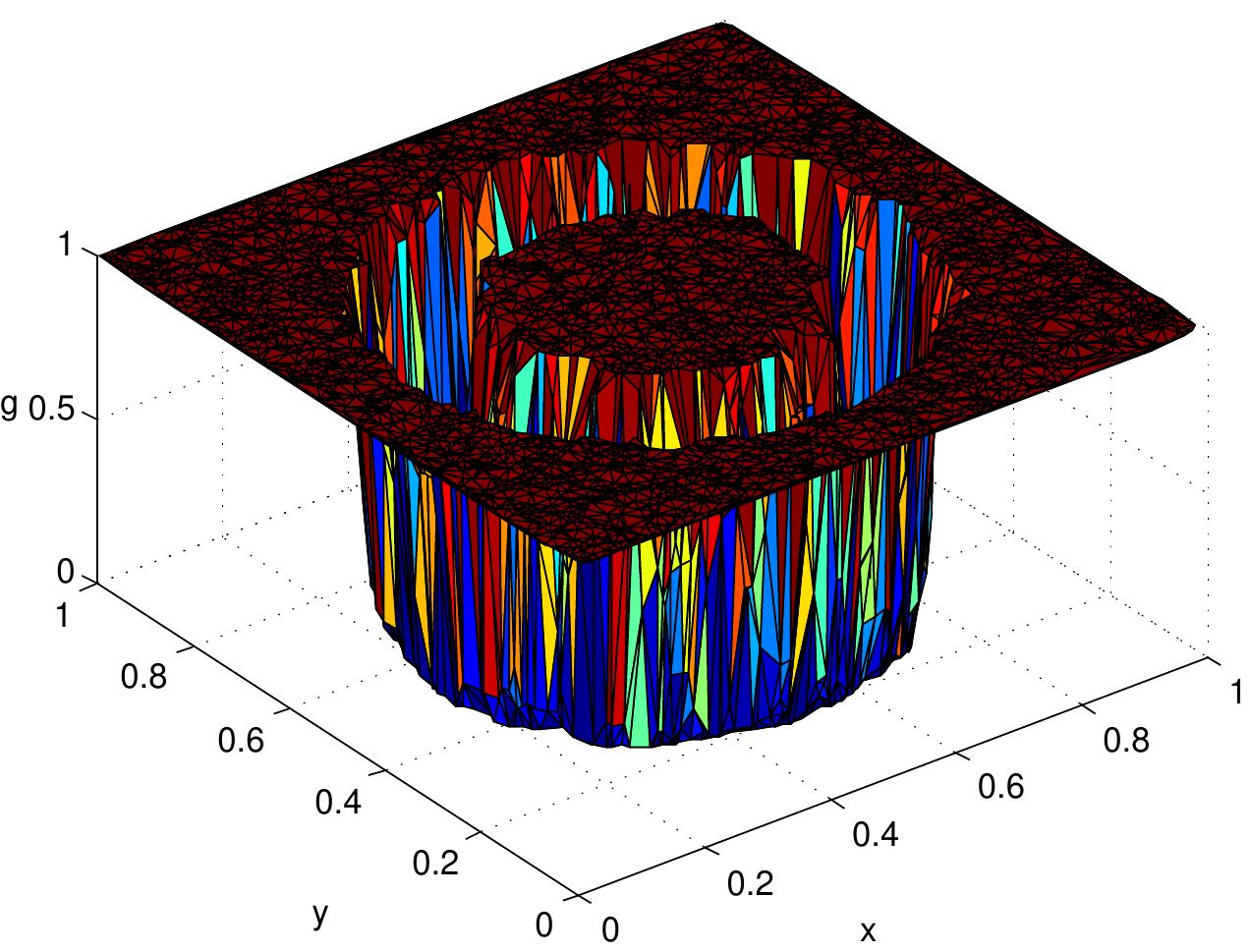}
  \label{subfig:gfunresultnorm}}
  \caption{Comparison of methods for Gaussian smoothing and computation of stopping function. The original image on the graph (a disk) is shown in \protect\subref{subfig:gfunI}. The rest of the figure is organized as follows: the two rightmost plots of the top row contain smoothed versions \(I_{\sigma}\) of the original image, the middle row contains gradient magnitudes of \(I_{\sigma}\) and the bottom row contains \(g\) values based on these gradient magnitudes. The results in the left column pertain to Gaussian derivative filtering with separate normalization using \(\sigma = 0.02\) and \(\lambda = 0.05\), those in the middle column pertain to simple Gaussian filtering using \(\sigma = 0.05\) and \(\lambda = 1000\) and those in the right column correspond to normalized Gaussian filtering with \(\sigma = 0.02\) and \(\lambda = 0.05\). We use the approximation of \eqref{eq:gradmagnmaxabsdiff} to compute the gradient magnitude in \protect\subref{subfig:gfunIsigmagradmagnsimple} and \protect\subref{subfig:gfunIsigmagradmagnnorm}. For all three methods, we filter \(\vecnorm{\nabla{I_{\sigma}}}\) with a median filter before feeding it to the formula for \(g\).}
  \label{fig:gfuncomparison}
\end{figure*}

However, the arbitrary graph setting introduces non-uniformities: in some parts of the graph, the vertices might be distributed more densely than in other parts. This implies that \eqref{eq:simpleGaussianfiltering} will operate counter-intuitively, introducing variations to the smoothed image in regions of the graph where the original intensity function is constant. To demonstrate this behavior, we use a simple binary image of a disk, shown in Fig. \ref{fig:gfuncomparison}\subref{subfig:gfunI}. The result of plain Gaussian filtering of this image is shown in Fig. \ref{fig:gfuncomparison}\subref{subfig:gfunIsigmasimple}. Not only has the range of image values changed, but also the interior of the original disk now exhibits significant variations in intensity values. This shortcoming is propagated to \(\vecnorm{\nabla{I_{\sigma}}}\) and \(g\) values, as we present in Fig. \ref{fig:gfuncomparison}\subref{subfig:gfunIsigmagradmagnsimple} and Fig. \ref{fig:gfuncomparison}\subref{subfig:gfunresultsimple} respectively. There is a deviation of \(g\) values from the ideal value of \(1\) inside the area corresponding to the disk and a variation in these values as well, which means that the gradient of the stopping function is not \(\mathbf{0}\), as it should.

To tackle this issue, we add a normalization term to \eqref{eq:simpleGaussianfiltering} to account for non-uniformities:
\begin{equation} \label{eq:normalizedGaussianfiltering}
I_{\sigma}(\mathbf{v}) = \frac{\displaystyle\sum_{\mathbf{w} \in \mathcal{V}} I(\mathbf{w}) G_{\sigma}(\mathbf{v} - \mathbf{w})}{\displaystyle\sum_{\mathbf{w} \in \mathcal{V}} G_{\sigma}(\mathbf{v} - \mathbf{w})}.
\end{equation}
We term this method normalized Gaussian filtering and show its result for the examined disk image in Fig. \ref{fig:gfuncomparison}\subref{subfig:gfunIsigmanorm}. The smoothed image is now very similar to the output of simple Gaussian filtering in the usual image processing setting with regularly spaced pixels. As a result, the corresponding magnitude of the gradient of \(I_{\sigma}\) and \(g\) function (shown in Fig. \ref{fig:gfuncomparison}\subref{subfig:gfunIsigmagradmagnnorm} and Fig. \ref{fig:gfuncomparison}\subref{subfig:gfunresultnorm} respectively) match our expectations.

An important observation at this point is that in the stopping function computation pipeline, we are rather interested in the smoothed image's gradient than in the smoothed image itself. Since the derivatives of the Gaussian filter have closed analytical forms, it is appealing to exchange the convolution with the gradient operator and convolve the image directly with Gaussian derivatives in order to obtain the gradient of \(I_{\sigma}\). In this case, normalization is not straightforward as in normalized Gaussian filtering: Gaussian derivatives assume both positive and negative values. We circumvent this difficulty by splitting the vertices into two sets, according to the sign of the Gaussian derivative with respect to the processed vertex, and perform separate normalization for each of these sets. This separation can be easily expressed in terms of the vertices' coordinates. If we denote \(\mathbf{v} = (v_1,\,v_2)\), then Gaussian derivative filtering with separate normalization is defined as:
\begin{align} \label{eq:Gaussianderivativefilteringseparatenormalization}
&\nabla{I_{\sigma}(\mathbf{v})} =\nonumber\\
&\left[\begin{array}{c}
\frac{\displaystyle\sum_{\substack{\mathbf{w} \in \mathcal{V}:\\w_1 \geq v_1}} I(\mathbf{w}) \frac{\partial G_{\sigma}(\mathbf{v} - \mathbf{w})}{\partial x}}{\displaystyle\sum_{\substack{\mathbf{w} \in \mathcal{V}:\\w_1 \geq v_1}} \frac{\partial G_{\sigma}(\mathbf{v} - \mathbf{w})}{\partial x}} + \frac{\displaystyle\sum_{\substack{\mathbf{w} \in \mathcal{V}:\\w_1 < v_1}} I(\mathbf{w}) \frac{\partial G_{\sigma}(\mathbf{v} - \mathbf{w})}{\partial x}}{-\displaystyle\sum_{\substack{\mathbf{w} \in \mathcal{V}:\\w_1 < v_1}} \frac{\partial G_{\sigma}(\mathbf{v} - \mathbf{w})}{\partial x}}\\
 \frac{\displaystyle\sum_{\substack{\mathbf{w} \in \mathcal{V}:\\w_2 \geq v_2}} I(\mathbf{w}) \frac{\partial G_{\sigma}(\mathbf{v} - \mathbf{w})}{\partial y}}{\displaystyle\sum_{\substack{\mathbf{w} \in \mathcal{V}:\\w_2 \geq v_2}} \frac{\partial G_{\sigma}(\mathbf{v} - \mathbf{w})}{\partial y}} + \frac{\displaystyle\sum_{\substack{\mathbf{w} \in \mathcal{V}:\\w_2 < v_2}} I(\mathbf{w}) \frac{\partial G_{\sigma}(\mathbf{v} - \mathbf{w})}{\partial y}}{-\displaystyle\sum_{\substack{\mathbf{w} \in \mathcal{V}:\\w_2 < v_2}} \frac{\partial G_{\sigma}(\mathbf{v} - \mathbf{w})}{\partial y}}
\end{array}\right]
\end{align}

The application of Gaussian derivative filtering with separate normalization on the examined image produces the results shown in Fig. \ref{fig:gfuncomparison}\subref{subfig:gfunIsigmagradmagnnormderiv} for \(\vecnorm{\nabla{I_{\sigma}}}\) and Fig. \ref{fig:gfuncomparison}\subref{subfig:gfunresultnormderiv} for \(g\). The quality of the stopping function is at least as satisfactory as in the normalized Gaussian filtering case of Fig. \ref{fig:gfuncomparison}\subref{subfig:gfunresultnorm}. Consequently, both our novel methods for Gaussian smoothing on graphs outperform the simple Gaussian filtering approach and can be readily used in the GAC framework.

\section{Results} \label{sec:results}
Having approximated the various terms of the active contour model in the arbitrary graph setting, we are able to apply the iterative algorithm for object detection on graphs that stems from the relevant PDE. The input comprises a graph \(\mathcal{G} = (\mathcal{V},\,\mathcal{E})\) whose vertices are embedded in \(\mathbb{R}^2\) and a real-valued function \(I:\mathcal{V}\rightarrow\mathbb{R}^+\). The algorithm includes the following steps:
\begin{enumerate}
\item Compute \(g(\vecnorm{\nabla{I_{\sigma}}})\), using either \eqref{eq:normalizedGaussianfiltering} and \eqref{eq:gradmagnmaxabsdiff} or \eqref{eq:Gaussianderivativefilteringseparatenormalization} to compute \(\vecnorm{\nabla{I_{\sigma}}}\). In both cases, median filtering is applied to \(\vecnorm{\nabla{I_{\sigma}}}\) before plugging it into the formula for \(g\). Then, compute the magnitude of \(g\)'s gradient using \eqref{eq:gradmagnmaxabsdiff} and its direction with \eqref{eq:gradgeomapprox}.
\item Choose a subset \(X\) of \(\mathcal{V}\) which \emph{contains} the objects to be detected and initialize the embedding function with the signed distance function from the boundary of \(X\), denoted by \(u_0\). By convention, \(u_0\) is positive inside \(X\).
\item \label{step:iterations} Iterate for \(r\in\mathbb{N}\)
\begin{equation} \label{eq:GACiteration}
u_{r} = u_{r-1} + \Delta{}t((\kappa - c)\vecnorm{\nabla{u_{r-1}}}g + \nabla{g}\cdot\nabla{u_{r-1}})
\end{equation}
until convergence, i.e. until \(u\) has not changed its sign at most vertices for several consecutive iterations. In the difference equation \eqref{eq:GACiteration}, \(\Delta{}t\) and \(c\) are positive constants and \(\kappa\) is the curvature of the level sets of \(u_{r-1}\).
\end{enumerate}

In practice, after each iteration of step \ref{step:iterations} of the algorithm, we smooth \(u_r\) with a median filter before proceeding to the next iteration. The parameters involved in the algorithm are the time step \(\Delta{}t\) of the difference equation, the balloon force constant \(c\), the scale \(\sigma\) of the Gaussian smoothing filter and parameter \(\lambda\) in \(g\)'s formula. Tuning their values depending on the particular input is pivotal in obtaining satisfactory segmentation results. In the following experiments, unless otherwise specified, we set \(\Delta{}t = 0.005\), \(c = 20\), \(\sigma = 0.02\) and \(\lambda = 0.05\).

Another important aspect in applying the active contour model on graphs is the method used to create the graph. The original input often consists only of a set of intensity values at vertex locations, without any information about the edges of the graph. This setting leaves us free to choose the underlying model for the structure of the graph. In our experiments, we used random geometric graphs and Delaunay triangulations (DTs). In other cases, one may have a full image at her disposal; however, the graph framework is still relevant. In particular, sampling the image uniformly at random with much fewer samples than the total number of pixels brings us to the previous setting and at the same time reduces the size of the input compared to the standard image-based active contour framework. An attractive alternative to random sampling is to extract vertex locations via watershed transformation. More specifically, we apply watershed transformation directly to the gradient of the image and place the vertices at the centroids (ultimate erosions) of the resulting segments. This approach leads to far better detection results than randomly sampling the image, as it captures image particularities into the structure of the graph and ``compresses'' the intensity function to the part that is crucial for segmentation.

\begin{figure}
    \centering
    \subfloat[]{\includegraphics[width=0.98\columnwidth]{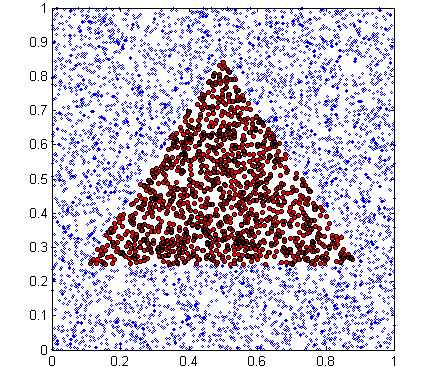}
    \label{subfig:triangleshape}}\\
    \subfloat[]{\includegraphics[width=0.48\columnwidth]{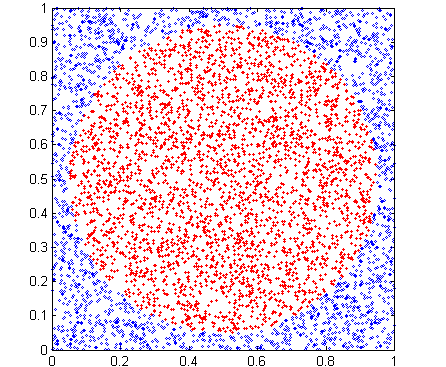}
    \label{subfig:triangleinit}}
    \hfil
    \subfloat[]{\includegraphics[width=0.48\columnwidth]{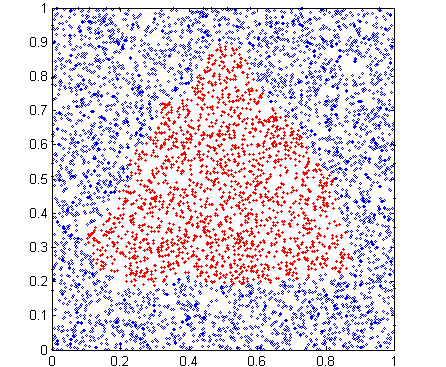}
    \label{subfig:triangleit60}}\\
    \subfloat[]{\includegraphics[width=0.48\columnwidth]{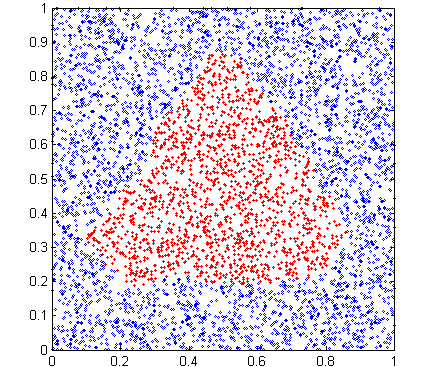}
    \label{subfig:triangleit120}}
    \hfil
    \subfloat[]{\includegraphics[width=0.48\columnwidth]{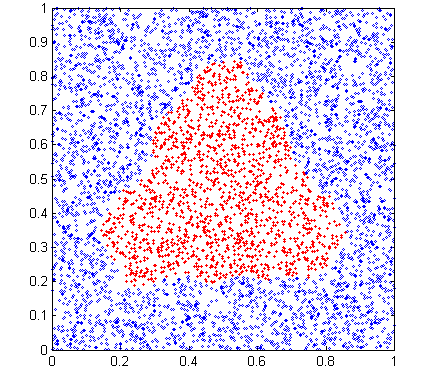}
    \label{subfig:triangleit180}}\\
    \subfloat[]{\includegraphics[width=0.48\columnwidth]{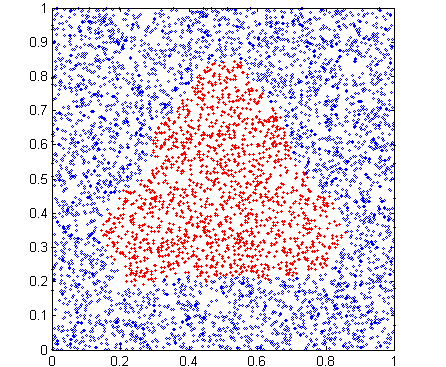}
    \label{subfig:triangleit240}}
    \hfil
    \subfloat[]{\includegraphics[width=0.48\columnwidth]{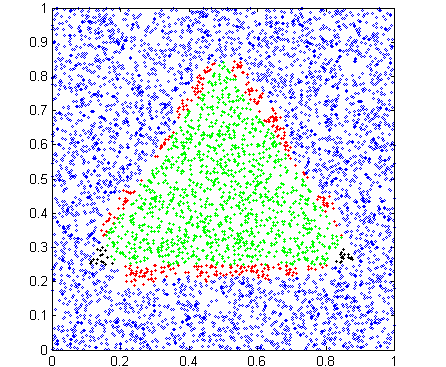}
    \label{subfig:triangleit300detection}}
    \caption{Detection of triangle on a random geometric graph. Edges are omitted for illustration purposes. \protect\subref{subfig:triangleshape} Original triangle on graph \protect\subref{subfig:triangleinit}--\protect\subref{subfig:triangleit240} Instances of active contour evolution at intervals of 60 iterations, with vertices in the contour's interior shown in red and the rest in blue \protect\subref{subfig:triangleit300detection} Final detection result after 300 iterations, using green for true positives, blue for true negatives, red for false positives and black for false negatives.}
    \label{fig:trianglegacsnapshots}
\end{figure}

\begin{figure}
    \centering
    \subfloat[]{\includegraphics[width=0.48\columnwidth]{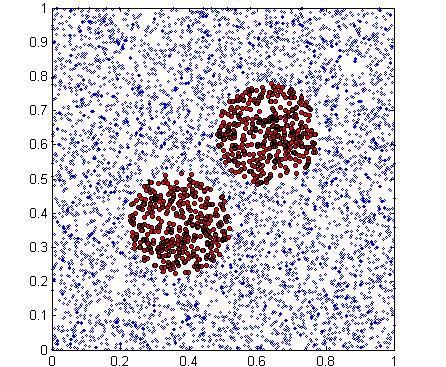}
    \label{subfig:twoclosedisksshape}}
    \hfil
    \subfloat[]{\includegraphics[width=0.48\columnwidth]{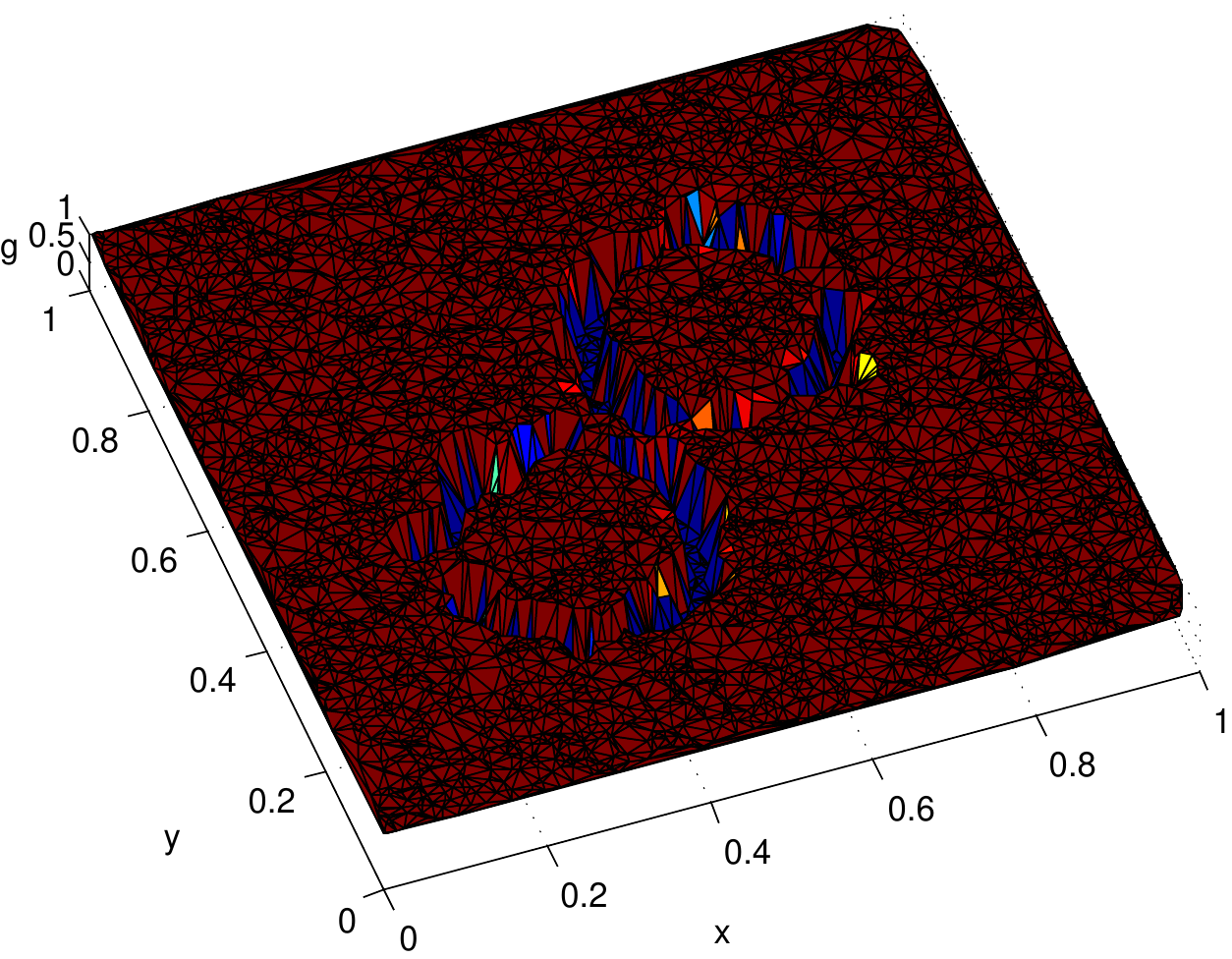}
    \label{subfig:twoclosedisksgfunction}}\\
    \subfloat[]{\includegraphics[width=0.48\columnwidth]{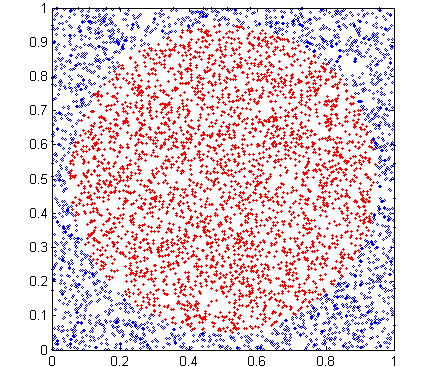}
    \label{subfig:twoclosedisksinit}}
    \hfil
    \subfloat[]{\includegraphics[width=0.48\columnwidth]{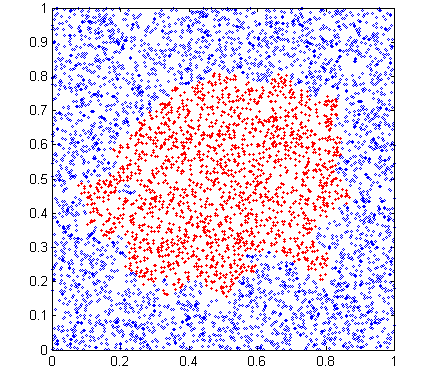}
    \label{subfig:twoclosedisksit40}}\\
    \subfloat[]{\includegraphics[width=0.48\columnwidth]{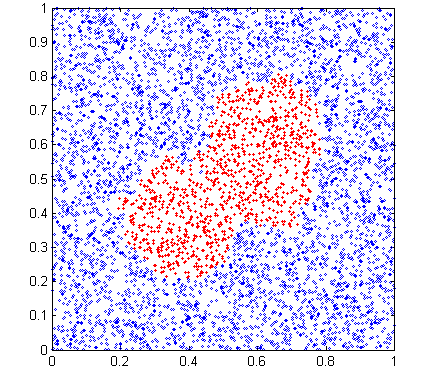}
    \label{subfig:twoclosedisksit80}}
    \hfil
    \subfloat[]{\includegraphics[width=0.48\columnwidth]{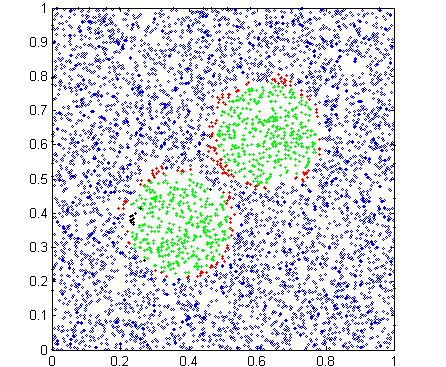}
    \label{subfig:twoclosedisksit120detection}}
    \caption{Detection of two disks on a random geometric graph. \protect\subref{subfig:twoclosedisksshape} Original disks on graph \protect\subref{subfig:twoclosedisksgfunction} \(g\) function \protect\subref{subfig:twoclosedisksinit}--\protect\subref{subfig:twoclosedisksit80} Instances of active contour evolution at intervals of 40 iterations, with vertices in the contour's interior shown in red and the rest in blue \protect\subref{subfig:twoclosedisksit120detection} Final detection result after 120 iterations, using green for true positives, blue for true negatives, red for false positives and black for false negatives.}
    \label{fig:twoclosedisksgacsnapshots}
\end{figure}

First, we apply our method to synthetic data. In particular, we construct RGGs by placing \(n\) vertices uniformly at random in \([0,1]^2\) and using \eqref{eq:radius} to create the edges. We then define binary image functions on the graphs, which model simple shapes. Fig. \ref{fig:trianglegacsnapshots} shows the evolution of the active contour when detecting a triangle, with \(n = 5500\). The final form of the contour in Fig. \ref{fig:trianglegacsnapshots}\subref{subfig:triangleit300detection}, comprising the green and red vertices, does not capture well the sharp corners of the triangle, which is expected due to the isotropic Gaussian smoothing of the image.

A harder benchmark is a non-connected shape, such as the two disks of Fig. \ref{fig:twoclosedisksgacsnapshots}, where \(n = 5500\). In order to allow the contour to change its topology and separate the disks, we need to tune the spatial parameters of the algorithm based on the distance of the objects. Specifically, the smoothed objects' boundaries must be at least two radii apart for \(g\) to assume values close to \(1\) between the objects, which is a necessary condition for separation. Setting \(\sigma = 0.005\), \(\lambda = 0.1\), \(C = 0.45\) and \(c = 40\), this condition is satisfied (Fig. \ref{fig:twoclosedisksgacsnapshots}\subref{subfig:twoclosedisksgfunction}) and the contour is able to distinguish the two disks (Fig. \ref{fig:twoclosedisksgacsnapshots}\subref{subfig:twoclosedisksit120detection}).

\begin{figure*}
  \centering
  \subfloat[]{\includegraphics[width=0.66\columnwidth]{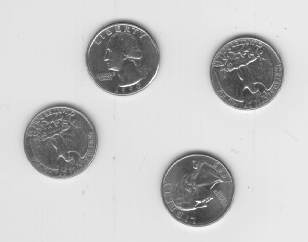}
  \label{subfig:coinsimage}}
  \hfil
  \subfloat[]{\includegraphics[width=0.66\columnwidth]{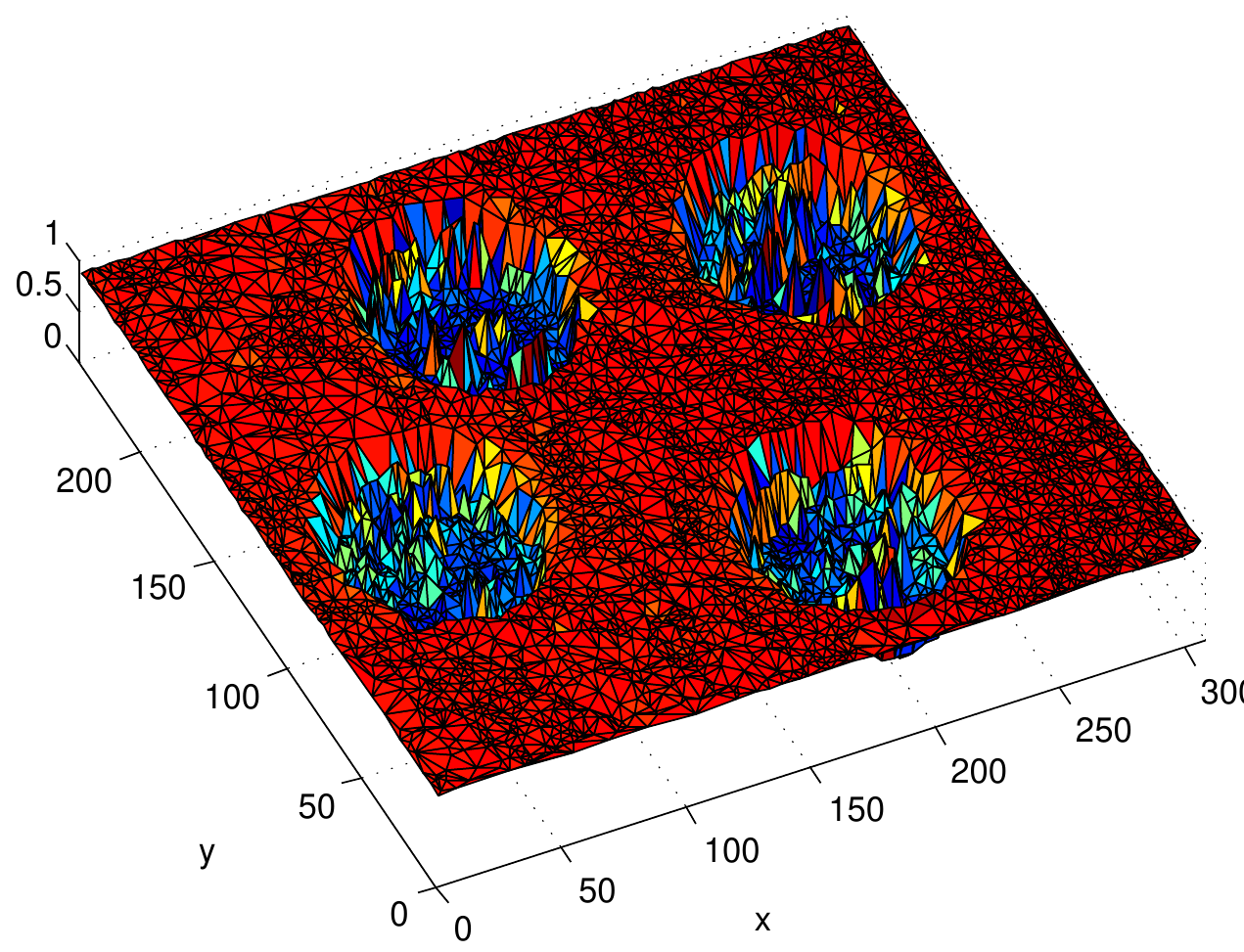}
  \label{subfig:coinsIgraph}}
  \hfil
  \subfloat[]{\includegraphics[width=0.66\columnwidth]{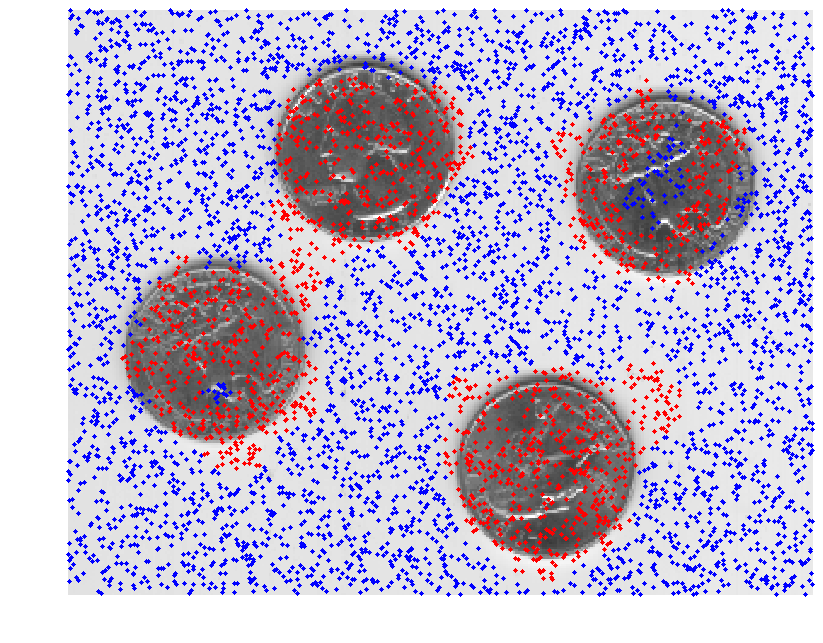}
  \label{subfig:coinsrggranddetection}}\\
  \subfloat[]{\includegraphics[width=0.66\columnwidth]{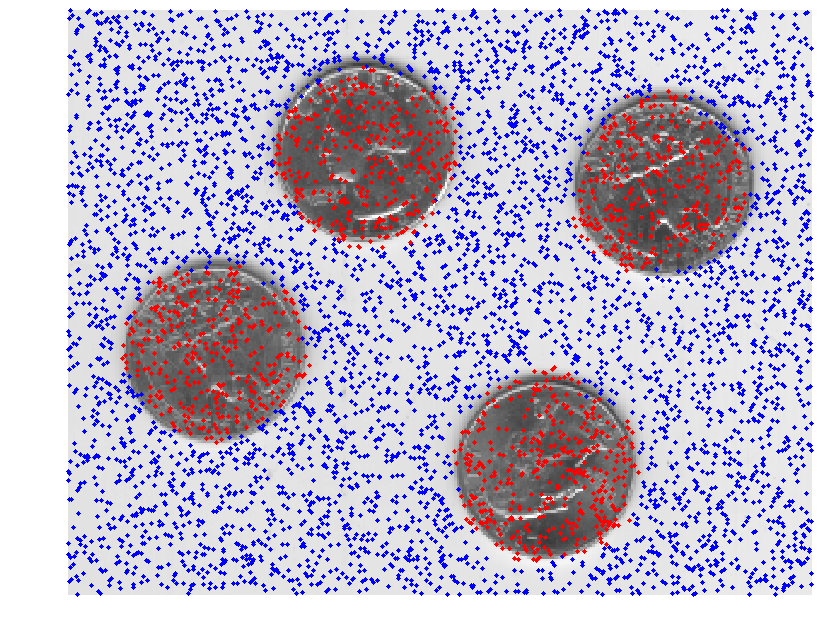}
  \label{subfig:coinsdtranddetection}}
  \hfil
  \subfloat[]{\includegraphics[width=0.66\columnwidth]{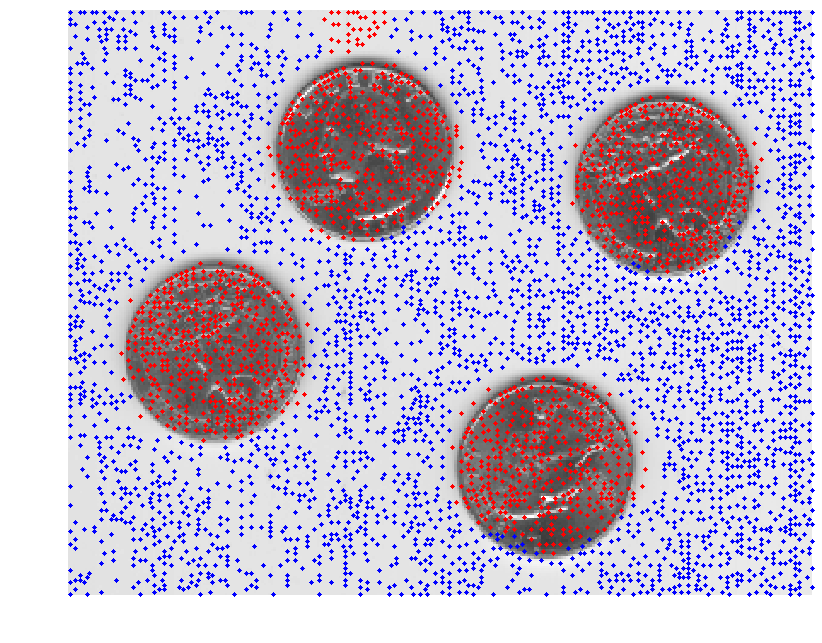}
  \label{subfig:coinsrggwsheddetection}}
  \hfil
  \subfloat[]{\includegraphics[width=0.66\columnwidth]{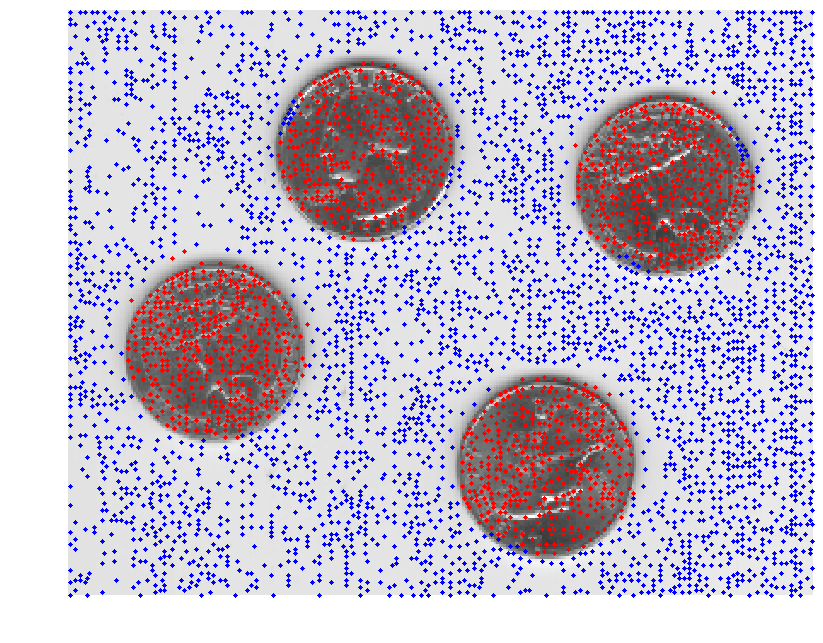}
  \label{subfig:coinsdtwsheddetection}}
  \caption{Detection of grayscale objects on graphs with active contours. \protect\subref{subfig:coinsimage} Full grayscale image with four coins \protect\subref{subfig:coinsIgraph} Intensity function on watershed-based DT \protect\subref{subfig:coinsrggranddetection}--\protect\subref{subfig:coinsdtwsheddetection} Final detection results overlaid on original image, with detected objects shown in red and background in blue. We use \protect\subref{subfig:coinsrggranddetection} randomly placed vertices with random geometric structure, \(\sigma = 0.02\) and \(\lambda = 0.03\), \protect\subref{subfig:coinsdtranddetection} randomly placed vertices with DT structure, \(\sigma = 0.02\) and \(\lambda = 0.03\), \protect\subref{subfig:coinsrggwsheddetection} watershed-placed vertices with random geometric structure, \(\sigma = 0.008\) and \(\lambda = 0.07\), and \protect\subref{subfig:coinsdtwsheddetection} watershed-placed vertices with DT structure, \(\sigma = 0.01\) and \(\lambda = 0.07\). The total number of iterations to obtain the final segmentation result is \protect\subref{subfig:coinsrggranddetection} 2200, \protect\subref{subfig:coinsdtranddetection} 4000, \protect\subref{subfig:coinsrggwsheddetection} 3000 and \protect\subref{subfig:coinsdtwsheddetection} 4000.}
  \label{fig:coinscomparison}
\end{figure*}

Our next experiment involves a full grayscale image with four distinct coins (Fig. \ref{fig:coinscomparison}\subref{subfig:coinsimage}). We make a two-fold comparison, on the one hand between placing the graph's vertices at random or via watershed transformation and on the other hand between using random geometric or DT structure for the edges. Results from the four experiments corresponding to all possible combinations are shown in Fig. \ref{fig:coinscomparison}\subref{subfig:coinsrggranddetection}--\subref{subfig:coinsdtwsheddetection}. To ensure a fair comparison, the number of randomly placed vertices is approximately the same as in the watershed case. In all the experiments, we set \(c = 2\). The most accurate segmentation is achieved with DT and watershed-placed vertices (Fig. \ref{fig:coinscomparison}\subref{subfig:coinsdtwsheddetection}), as the boundaries of the objects are captured very well. Using randomly placed vertices and DT structure, or watershed and random geometric structure, also yields decent results (Fig. \ref{fig:coinscomparison}\subref{subfig:coinsdtranddetection} and \subref{subfig:coinsrggwsheddetection}). On the contrary, the combination of randomly placed vertices and random geometric structure (i.e. a proper RGG) leads to poor segmentation, in which close objects are not separated and others have holes opened in their interior (Fig. \ref{fig:coinscomparison}\subref{subfig:coinsrggranddetection}). Consequently, the DT structure is preferable to the random geometric one and usage of watershed transformation to place vertices when a full image is available is better than random placement.

\begin{figure}
  \centering
  \subfloat[]{\includegraphics[width=0.48\columnwidth]{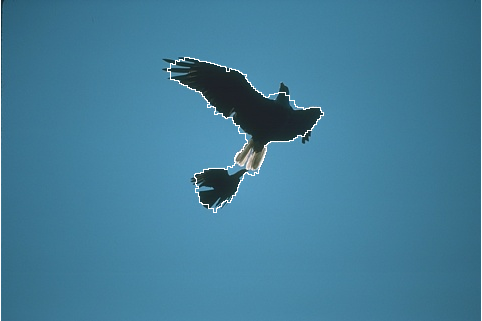}
  \label{subfig:135069}}
  \hfil
  \subfloat[]{\includegraphics[width=0.48\columnwidth]{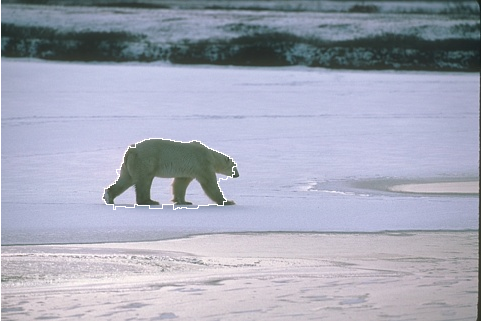}
  \label{subfig:100007}}\\
  \subfloat[]{\includegraphics[width=0.31\columnwidth]{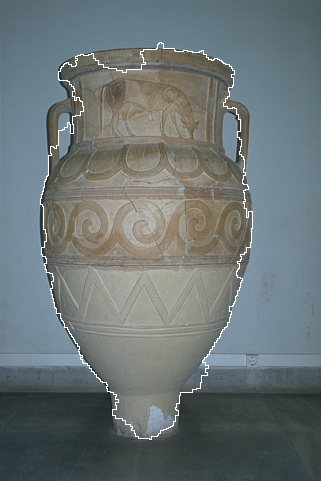}
  \label{subfig:227092}}
  \hfil
  \subfloat[]{\includegraphics[width=0.31\columnwidth]{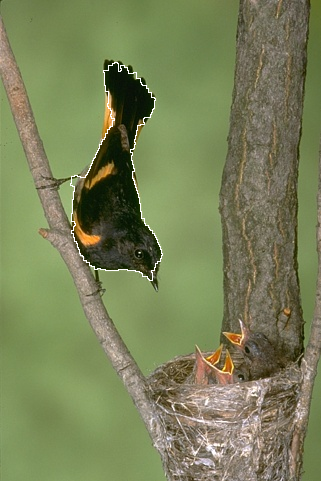}
  \label{subfig:163014}}
  \hfil
  \subfloat[]{\includegraphics[width=0.31\columnwidth]{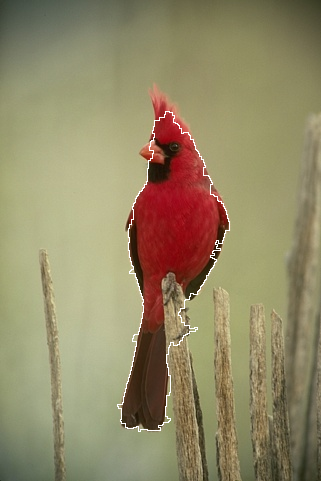}
  \label{subfig:196027}}
  \caption{Segmentation of images from BSDS500 dataset. The detected boundary, marked white, is defined as the boundary of the union of the watershed segments for which the corresponding graph vertex belongs to the final contour's interior. We use \(\sigma = 0.005\), \(\lambda = 0.03\) and \(c = 2\) in all cases except \protect\subref{subfig:163014}, for which we set \(\lambda = 0.05\), and \protect\subref{subfig:196027}, for which we set \(\sigma = 0.01\) and \(\lambda = 0.02\). The total number of iterations to obtain the final segmentation result is \protect\subref{subfig:135069} 3400, \protect\subref{subfig:100007} 2000, \protect\subref{subfig:227092} 2600, \protect\subref{subfig:163014} 4400 and \protect\subref{subfig:196027} 5600.}
  \label{fig:BSDS500}
\end{figure}

We use the combination of watershed-placed vertices and DT structure for the edges to repeat the above experiment for a collection of natural color images coming from the Berkeley Segmentation Dataset BSDS500 \cite{MFTM01}. The images were converted to grayscale for the application of our method. Segmentation results on the images are presented in Fig. \ref{fig:BSDS500}. In general, our algorithm detects the dominant objects in the images successfully, even though background clutter and thin protrusions or concavities of the objects' boundaries may cause minor inaccuracies.

\begin{figure}
    \centering
    \subfloat[]{\includegraphics[width=0.31\columnwidth]{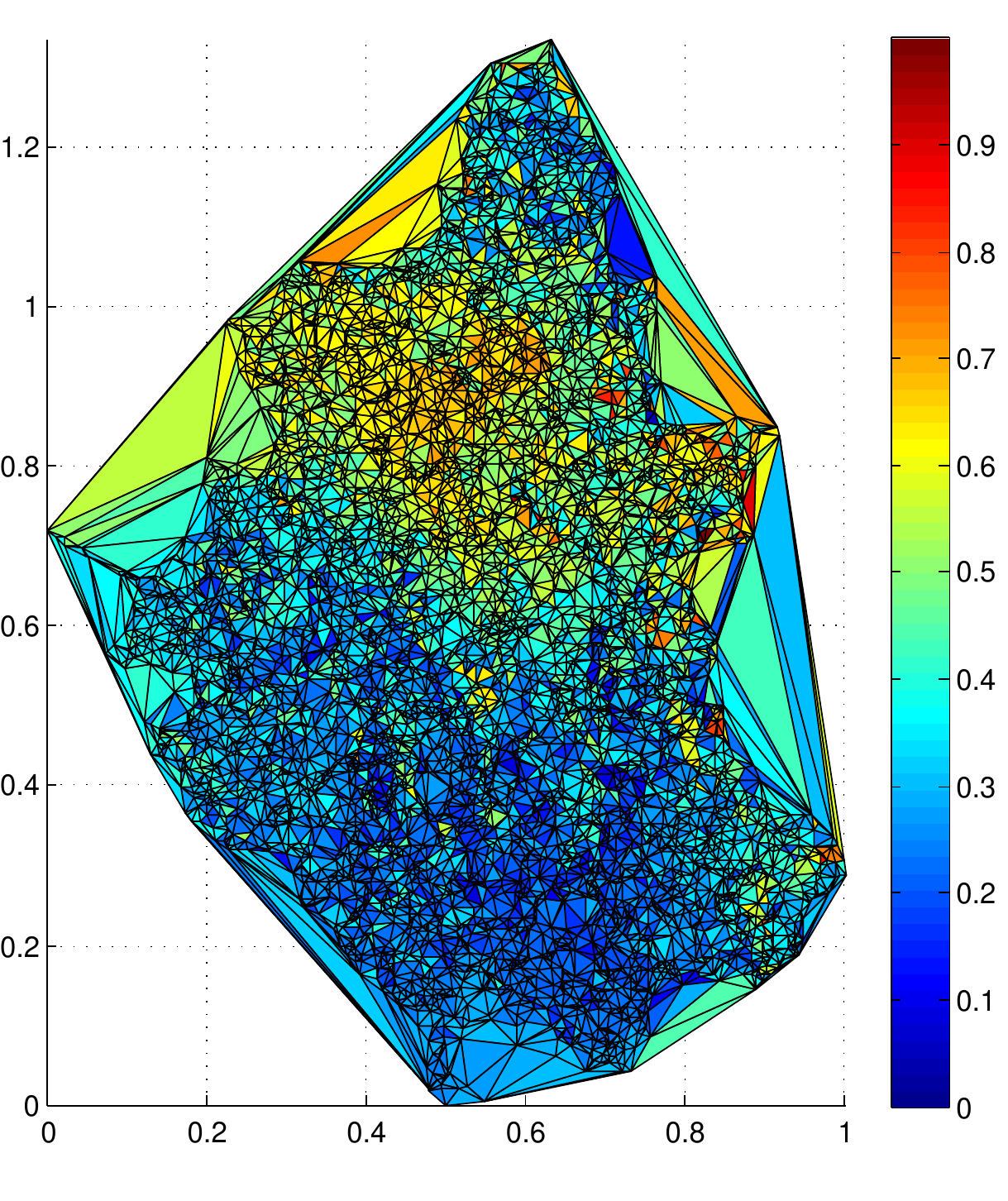}
    \label{subfig:windipirosIgraph}}
    \hfil
    \subfloat[]{\includegraphics[width=0.31\columnwidth]{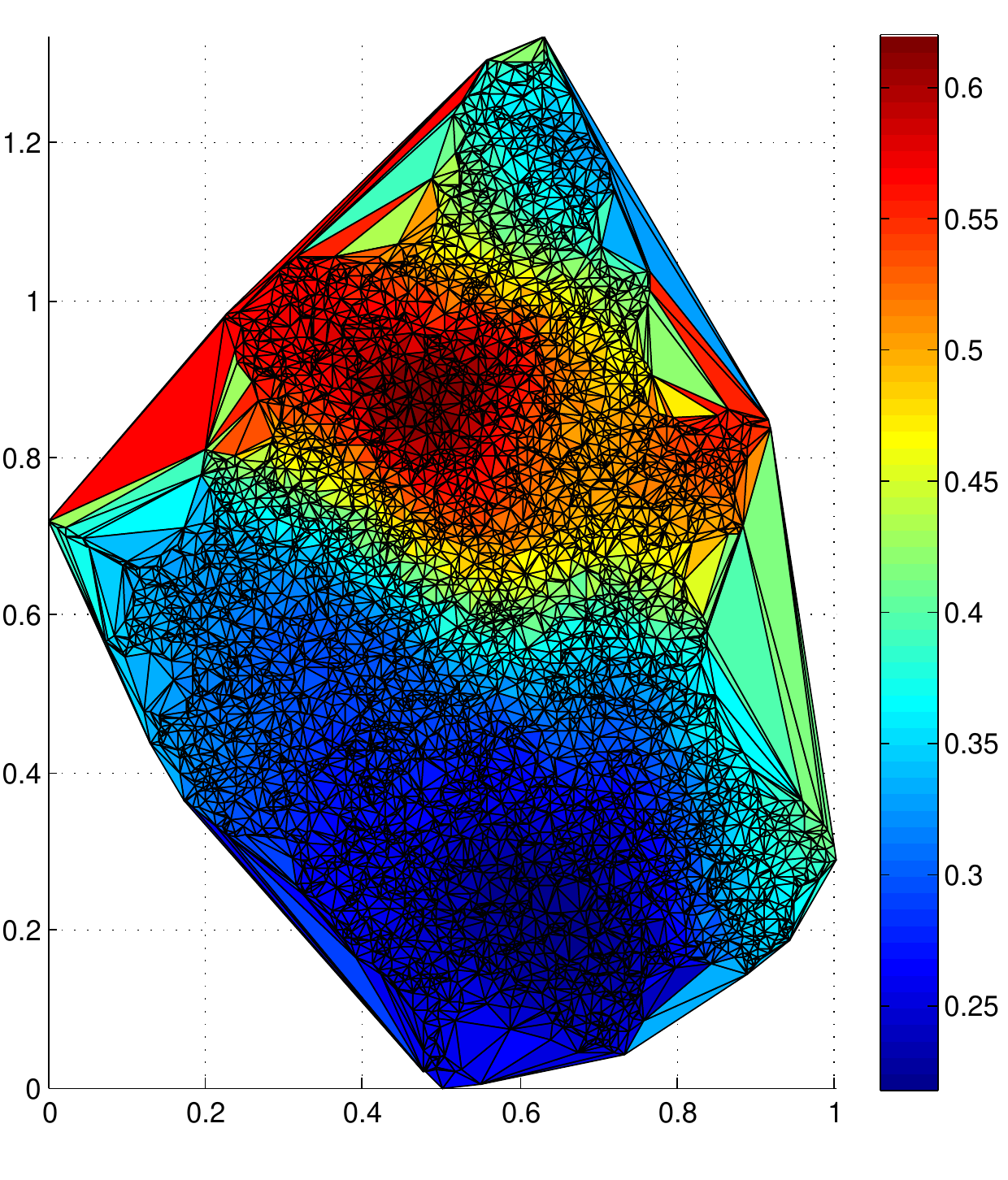}
    \label{subfig:windipirosIsmoothed}}
    \hfil
    \subfloat[]{\includegraphics[width=0.31\columnwidth]{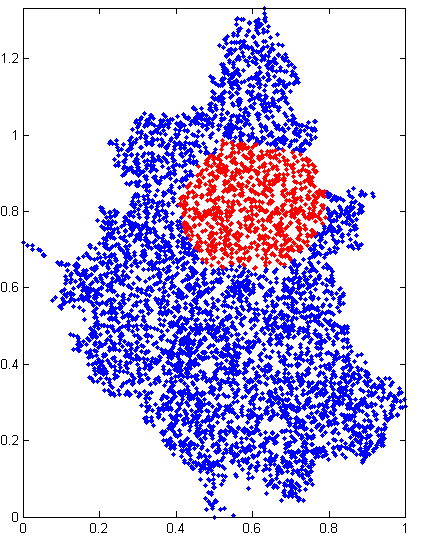}
    \label{subfig:windipirosdetection}}
    \caption{Segmentation of wind speed data on a graph. \protect\subref{subfig:windipirosIgraph} Normalized data on graph \protect\subref{subfig:windipirosIsmoothed} Smoothed wind speed \protect\subref{subfig:windipirosdetection} Final detection result after 1000 iterations, with vertices in the contour's interior shown in red and the rest in blue.}
    \label{fig:windipirosresults}
\end{figure}

\begin{figure}
    \centering
    \subfloat[]{\includegraphics[width=0.48\columnwidth]{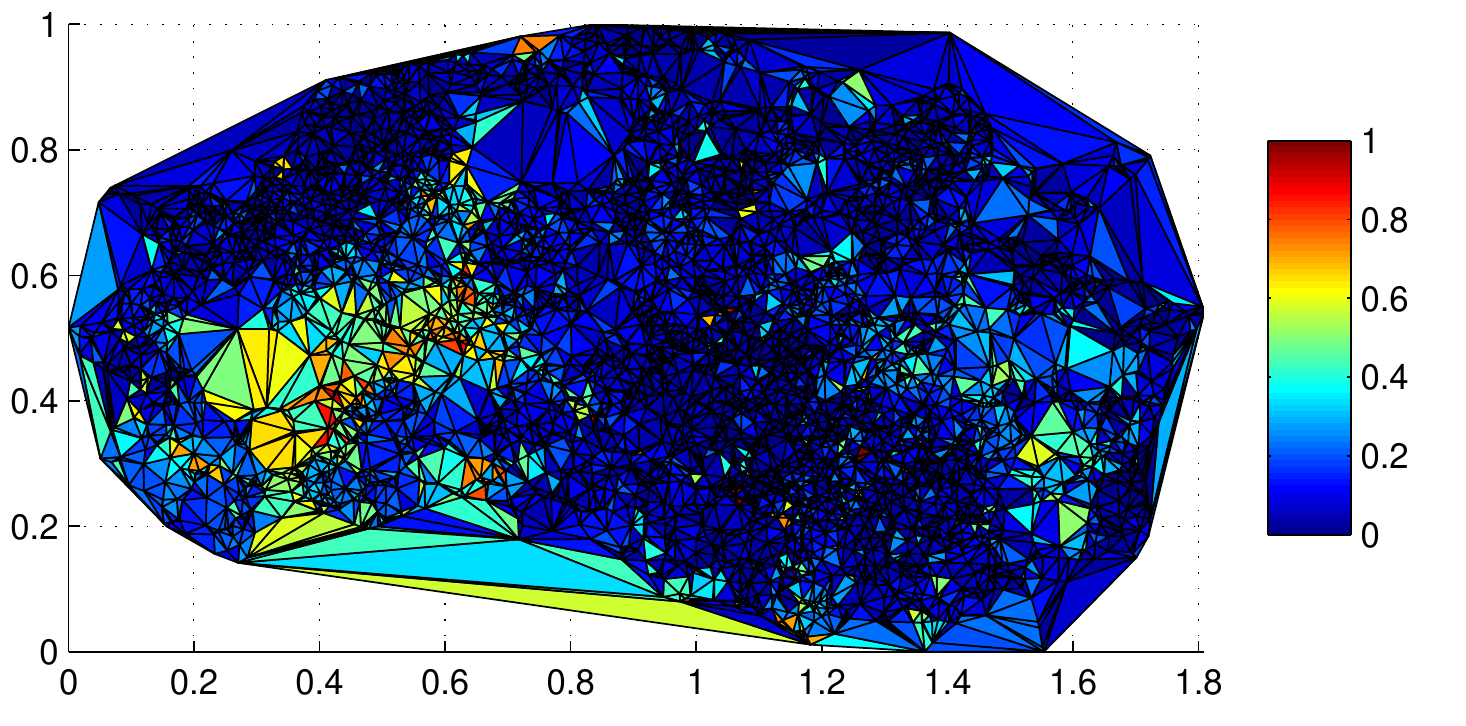}
    \label{subfig:cellularIgraph}}
    \hfil
    \subfloat[]{\includegraphics[width=0.48\columnwidth]{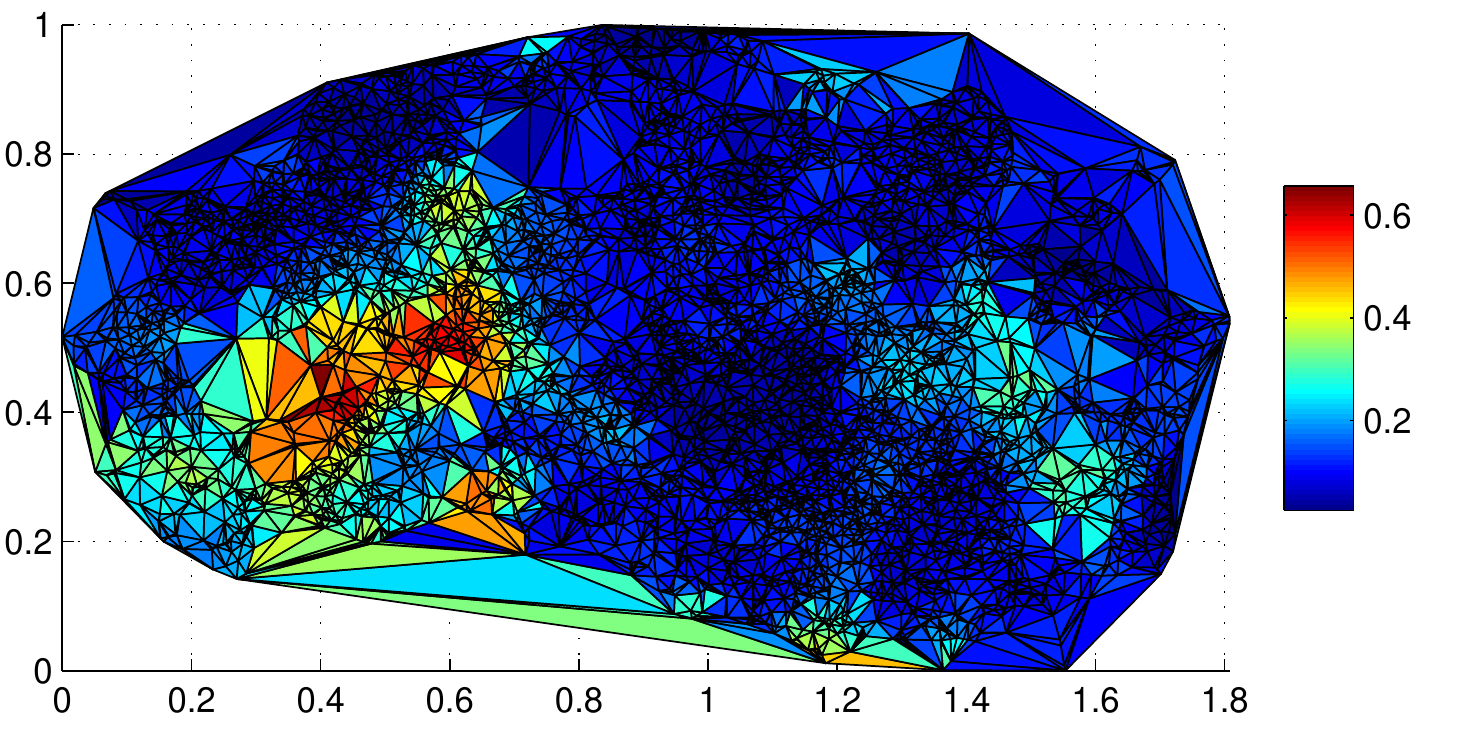}
    \label{subfig:cellularIsmoothed}}\\
    \subfloat[]{\includegraphics[width=0.48\columnwidth]{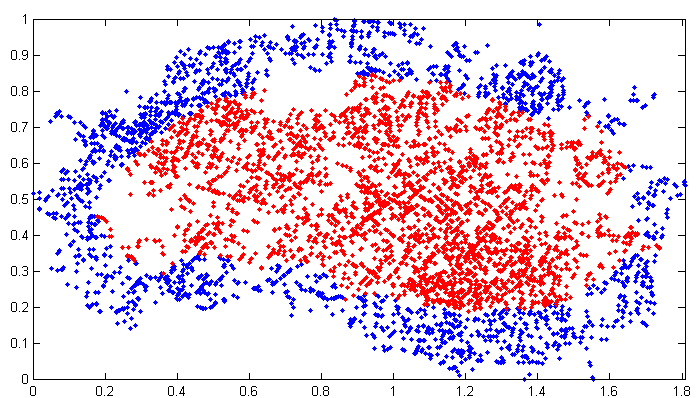}
    \label{subfig:cellularinit}}
    \hfil
    \subfloat[]{\includegraphics[width=0.48\columnwidth]{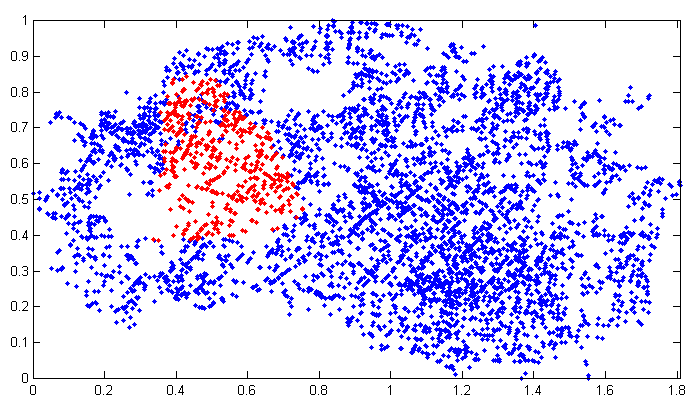}
    \label{subfig:cellulardetection}}
    \caption{Segmentation of signal strength data of a cellular network on a graph. \protect\subref{subfig:cellularIgraph} Normalized data on graph \protect\subref{subfig:cellularIsmoothed} Smoothed signal strength \protect\subref{subfig:cellularinit} Initial contour \protect\subref{subfig:cellulardetection} Final detection result after 40000 iterations, with vertices in the contour's interior shown in red and the rest in blue.}
    \label{fig:cellularresults}
\end{figure}

An interesting application of our method is related to geographical data, where the two spatial coordinates are longitude and latitude and the image function can encode information about any type of real-valued signal defined at the vertices of the graph. Such a signal is the average annual wind speed, which is particularly important for locating regions with high wind power potential. Fig. \ref{fig:windipirosresults} presents the application of our active contour algorithm to average annual wind speed data on a graph constructed as a Delaunay triangulation. We use \(\sigma = 0.05\), \(\lambda = 0.8\) and \(c = 5\) and normalize the coordinates and the values of wind speed. The algorithm detects a cluster that corresponds to a region with relatively uniform and quite high wind speed. Another example regards the signal strength of a cellular network. Fig. \ref{fig:cellularresults} demonstrates the result of our method for such data on a Delaunay triangulation. We again normalize the data and use a very small time step \(\Delta{}t = 10^{-4}\) to guarantee convergence, which requires far more iterations until termination than in the previous experiments. In addition, we set \(\sigma = 0.03\) and \(\lambda = 0.02\). The segmented set of vertices comprises two regions, the southern of which is characterized by an increased signal strength compared to the rest of the graph. Our approach is tailored for geographical data with arbitrary spatial configuration such as the above cases, which grants greater flexibility when collecting measurements.

\section{Conclusion}
In this paper, we introduce approximations of the gradient and curvature terms involved in the level set formulation of active contour evolution models for the case of arbitrary graphs. We examine theoretically the conditions under which these approximations converge in probability to the true value of the operators and, in the case of gradient, the respective rate of convergence, in the limit of large graphs. At an algorithmic level, we propose smoothing filtering to improve accuracy of such approximations and help the partial difference scheme \eqref{eq:GACiteration} converge properly, and provide improved implementations of Gaussian smoothing on graphs which account for potential non-uniformities. We also demonstrate the applicability of active contours on graphs, equipped with our approximations, for segmentation of regular images as well as raw geographical data.

A remaining challenge in our work is related to the curvature term of active contour models. Both the geometric and the gradient-based approximation which are proposed are proved to converge in probability to the true value of curvature at points with nonzero gradient, which is the first result of this kind to the best of our knowledge. However, due to the cascaded approximation that we perform, a larger amount of noise is injected in the approximate values, which is also reflected in the slower convergence observed in our empirical tests compared to gradient approximation. This forces us to take a very small step in time in some cases when updating the embedding function, which leads to much slower convergence of the active contour algorithm. To overcome this difficulty, a deeper analysis of the curvature term needs to be accomplished, ideally establishing asymptotic bounds on the respective approximation error similar to our bound for the gradient approximation error.

Our theoretical analysis considers random geometric graphs, whose definition simplifies convergence proofs for our approximations. However, judging from segmentation quality, more regular graph structures, such as Delaunay triangulations, lead to more accurate results. Therefore, an interesting extension of our work is the theoretical study of our approximations on graphs formed as Delaunay triangulations. Furthermore, our initial Gaussian smoothing of the image corresponds to isotropic diffusion, which blurs predominant edges and rounds corners, like in Fig. \ref{fig:trianglegacsnapshots}. More faithful preservation of edges in the final segmentation can be ensured by defining anisotropic smoothing on graphs.

\section*{Appendix}

\begin{IEEEproof}[Proof of Theorem \ref{th:curvgeomapproxconv}]
Due to differentiability of \(u\), Theorem \ref{th:gradapproxconv} implies that the difference between the geometric gradient approximation \(\nabla{u_g(w)}\) of \eqref{eq:gradgeomapprox} and the true value of the gradient \(\nabla{u(w)}\) converges in probability to zero for every neighbor \(w\) of \(v\):
\[\nabla{u_g(w)}-\nabla{u(w)} \xrightarrow{\prob} \mathbf{0}.\]
We show that a consequence of this convergence is that \(\mathbf{F}_g(w)-\mathbf{F}(w) \xrightarrow{\prob} \mathbf{0}\). The law of total probability implies that for every \(\epsilon > 0\)
\begin{align*}
&\prob\left(\vecnorm{\mathbf{F}_g(w)-\mathbf{F}(w)} > \epsilon\right)\\
&{\leq}\;\prob\left(\nabla{u(w)} = \mathbf{0}\right)+\prob\left(\left.\nabla{u_g(w)} = \mathbf{0}\right|\nabla{u(w)} \neq \mathbf{0}\right)\\
&{+}\:\prob\left(\left.\vecnorm{\mathbf{F}_g(w)-\mathbf{F}(w)} > \epsilon\right|\nabla{u(w)} \neq \mathbf{0}\text{ and }\nabla{u_g(w)} \neq \mathbf{0}\right).
\end{align*}

We examine the terms on the right-hand side of the above inequality, showing that all three of them converge to \(0\), which guarantees that their sum converges to \(0\) as well. Since \(\vecnorm{\nabla{u(v)}} > 0\), it holds for the first term that
\begin{align*}
&\prob\left(\nabla{u(w)} = \mathbf{0}\right) \leq \prob\left(\vecnorm{\nabla{u(w)}} < \frac{\vecnorm{\nabla{u(v)}}}{2}\right)\\
&{=}\;\prob\left(\vecnorm{\nabla{u(v)}}-\vecnorm{\nabla{u(w)}} > \frac{\vecnorm{\nabla{u(v)}}}{2}\right)\\
&{\leq}\;\prob\left(\vecnorm{\nabla{u(w)}-\nabla{u(v)}} > \frac{\vecnorm{\nabla{u(v)}}}{2}\right) \to 0,
\end{align*}
where the last limit is due to \(\nabla{u(w)} \xrightarrow{\prob} \nabla{u(v)}\), as \(\mathbf{w} \xrightarrow{\prob} \mathbf{v}\) and \(\nabla{u}\) is continuous. For the second term, the sum property of convergence in probability yields
\[\nabla{u_g(w)}-\nabla{u(w)}+\nabla{u(w)} = \nabla{u_g(w)} \xrightarrow{\prob} \nabla{u(v)}\]
and hence this term can be bounded as follows:
\begin{align*}
&\prob\left(\nabla{u_g(w)} = \mathbf{0}\right) \leq \prob\left(\vecnorm{\nabla{u_g(w)}} < \frac{\vecnorm{\nabla{u(v)}}}{2}\right)\\
&{=}\;\prob\left(\vecnorm{\nabla{u(v)}}-\vecnorm{\nabla{u_g(w)}} > \frac{\vecnorm{\nabla{u(v)}}}{2}\right)\\
&{\leq}\;\prob\left(\vecnorm{\nabla{u_g(w)}-\nabla{u(v)}} > \frac{\vecnorm{\nabla{u(v)}}}{2}\right) \to 0.
\end{align*}
In the third term, it is given that \(\vecnorm{\nabla{u_g(w)}} \geq \delta > 0\) for some \(\delta\). We write
\begin{align*}
&\mathbf{F}_g(w)-\mathbf{F}(w) = \frac{\nabla{u_g(w)}\vecnorm{\nabla{u(w)}}-\nabla{u(w)}\vecnorm{\nabla{u_g(w)}}}{\vecnorm{\nabla{u_g(w)}}\vecnorm{\nabla{u(w)}}}\\
&{=}\;(\vecnorm{\nabla{u(w)}}(\nabla{u_g(w)}-\nabla{u(w)})\\
&{+}\:\nabla{u(w)}(\vecnorm{\nabla{u(w)}}-\vecnorm{\nabla{u_g(w)}}))/(\vecnorm{\nabla{u_g(w)}}\vecnorm{\nabla{u(w)}}).
\end{align*}
The triangle inequality can be applied to the above expression:
\begin{align*}
&\vecnorm{\mathbf{F}_g(w)-\mathbf{F}(w)}\\
&{\leq}\;\left(\vphantom{\Bigl\lvert\vecnorm{\nabla{u(w)}}-\vecnorm{\nabla{u_g(w)}}\Bigr\rvert}\vecnorm{\nabla{u(w)}}\vecnorm{\nabla{u_g(w)}-\nabla{u(w)}}\right.\\
&{+}\:\left.\vecnorm{\nabla{u(w)}}\Bigl\lvert\vecnorm{\nabla{u(w)}}-\vecnorm{\nabla{u_g(w)}}\Bigr\rvert\right)\\
&{/}\:(\vecnorm{\nabla{u_g(w)}}\vecnorm{\nabla{u(w)}})\\
&{\leq}\;\frac{2\vecnorm{\nabla{u_g(w)}-\nabla{u(w)}}}{\vecnorm{\nabla{u_g(w)}}} \leq \frac{2\vecnorm{\nabla{u_g(w)}-\nabla{u(w)}}}{\delta}.
\end{align*}
Therefore, given that \(\nabla{u(w)} \neq \mathbf{0}\) and \(\nabla{u_g(w)} \neq \mathbf{0}\), it holds for every \(\epsilon > 0\) that
\begin{align*}
&\prob\left(\vecnorm{\mathbf{F}_g(w)-\mathbf{F}(w)} > \epsilon\right)\\
&{\leq}\;\prob\left(\frac{2\vecnorm{\nabla{u_g(w)}-\nabla{u(w)}}}{\delta} > \epsilon\right)\\
&{=}\;\prob\left(\vecnorm{\nabla{u_g(w)}-\nabla{u(w)}} > \frac{\epsilon\delta}{2}\right) \to 0.
\end{align*}
As a result,
\begin{equation} \label{eq:Fgconv}
\mathbf{F}_g(w)-\mathbf{F}(w) \xrightarrow{\prob} \mathbf{0}.
\end{equation}

Using the closed form \eqref{eq:Ia} for \(I_a(w_i)\) and the product and sum properties of convergence in probability, we obtain
\begin{equation} \label{eq:FgminusFofwiintconv}
\int_{C_a(w_i)} \left(\mathbf{F}_g(w_i)-\mathbf{F}(w_i)\right)\cdot\mathbf{n}\,d\ell \xrightarrow{\prob} 0.
\end{equation}
In addition, the law of total probability implies that for every \(\epsilon > 0\)
\begin{align*}
&\prob\left(\left|\int\limits_{C_a(w_i)} \left(\mathbf{F}(w_i)-\mathbf{F}\right)\cdot\mathbf{n}\,d\ell\right| > \epsilon\right)\\
&{\leq}\;\prob\left(\mathbf{F}\text{ not continuous at some point on }C_a(w_i)\right)\\
&{+}\:\prob\left(\left|\int\limits_{C_a(w_i)} \left(\mathbf{F}(w_i)-\mathbf{F}\right)\cdot\mathbf{n}\,d\ell\right| > \epsilon|\mathbf{F}\text{ cont. on }C_a(w_i)\right).
\end{align*}
Since \(\rho(n) \in o\left(1\right)\), it holds that \(C_a(w_i) \xrightarrow{\prob} \{\mathbf{v}\}\). Moreover, \(\mathbf{F}\) is continuous at \(\mathbf{v}\), yielding
\[\prob\left(\mathbf{F}\text{ not continuous at some point on }C_a(w_i)\right) \to 0.\]
For the second term on the right-hand side of the previous inequality, due to the given continuity of \(\mathbf{F}\) on the arc \(C_a(w_i)\) and the fact that the length of this arc \(d(v,w_i)\Delta\phi(w_i) \xrightarrow{\prob} 0\) and \(\mathbf{w}_i \in C_a(w_i)\), it follows that
\begin{align*}
&\prob\left(\left|\int\limits_{C_a(w_i)} \left(\mathbf{F}(w_i)-\mathbf{F}\right)\cdot\mathbf{n}\,d\ell\right| > \epsilon|\mathbf{F}\text{ cont. on }C_a(w_i)\right) \to\\
&0.
\end{align*}
Consequently, the following convergence result is obtained:
\begin{equation} \label{eq:FofwiminusFintconv}
\int_{C_a(w_i)} \left(\mathbf{F}(w_i)-\mathbf{F}\right)\cdot\mathbf{n}\,d\ell \xrightarrow{\prob} 0.
\end{equation}
We combine \eqref{eq:FgminusFofwiintconv} and \eqref{eq:FofwiminusFintconv} using the sum property of convergence in probability into
\begin{equation} \label{eq:FgofwiminusFintconv}
\int_{C_a(w_i)} \left(\mathbf{F}_g(w_i)-\mathbf{F}\right)\cdot\mathbf{n}\,d\ell = I_a(w_i)-\int_{C_a(w_i)} \mathbf{F}\cdot\mathbf{n}\,d\ell \xrightarrow{\prob} 0.
\end{equation}

The next step is to show that
\begin{equation} \label{eq:FgminusFnormmeanconv}
\frac{\mathbf{F}_g(w_i)+\mathbf{F}_g(w_{i+1})}{\vecnorm{\mathbf{F}_g(w_i)+\mathbf{F}_g(w_{i+1})}}-\frac{\mathbf{F}(w_i)+\mathbf{F}(w_{i+1})}{\vecnorm{\mathbf{F}(w_i)+\mathbf{F}(w_{i+1})}} \xrightarrow{\prob} \mathbf{0}.
\end{equation}
Again, we employ the law of total probability to arrive at the following inequality for every \(\epsilon > 0\):
\begin{align*}
&\prob\left(\vecnorm{\frac{\mathbf{F}_g(w_i)+\mathbf{F}_g(w_{i+1})}{\vecnorm{\mathbf{F}_g(w_i)+\mathbf{F}_g(w_{i+1})}}-\frac{\mathbf{F}(w_i)+\mathbf{F}(w_{i+1})}{\vecnorm{\mathbf{F}(w_i)+\mathbf{F}(w_{i+1})}}} > \epsilon\right)\\
&{\leq}\;\prob\left(\mathbf{F}(w_i)+\mathbf{F}(w_{i+1}) = \mathbf{0}\right)\\
&{+}\:\prob\left(\mathbf{F}_g(w_i)+\mathbf{F}_g(w_{i+1}) = \mathbf{0}\left|\mathbf{F}(w_i)+\mathbf{F}(w_{i+1}) \neq \mathbf{0}\right.\right)\\
&{+}\:\prob\left(\vecnorm{\frac{\mathbf{F}_g(w_i)+\mathbf{F}_g(w_{i+1})}{\vecnorm{\mathbf{F}_g(w_i)+\mathbf{F}_g(w_{i+1})}}-\frac{\mathbf{F}(w_i)+\mathbf{F}(w_{i+1})}{\vecnorm{\mathbf{F}(w_i)+\mathbf{F}(w_{i+1})}}} > \epsilon\right.\\
&\left.\left|\vphantom{\vecnorm{\frac{\mathbf{F}_g(w_i)+\mathbf{F}_g(w_{i+1})}{\vecnorm{\mathbf{F}_g(w_i)+\mathbf{F}_g(w_{i+1})}}-\frac{\mathbf{F}(w_i)+\mathbf{F}(w_{i+1})}{\vecnorm{\mathbf{F}(w_i)+\mathbf{F}(w_{i+1})}}}}\mathbf{F}(w_i)+\mathbf{F}(w_{i+1}) \neq \mathbf{0}\text{ and }\mathbf{F}_g(w_i)+\mathbf{F}_g(w_{i+1}) \neq \mathbf{0}\right.\right).
\end{align*}
Similarly to previous parts of the proof, it will be shown that each of the terms on the right-hand side of the above inequality converges to \(0\). As \(\mathbf{w}_i \xrightarrow{\prob} \mathbf{v}\) and \(\mathbf{w}_{i+1} \xrightarrow{\prob} \mathbf{v}\), the continuity of \(\mathbf{F}\) at \(\mathbf{v}\) yields \(\mathbf{F}(w_i) \xrightarrow{\prob} \mathbf{F}(v)\) and \(\mathbf{F}(w_{i+1}) \xrightarrow{\prob} \mathbf{F}(v)\). The sum property of convergence in probability then implies that
\[\mathbf{F}(w_i)+\mathbf{F}(w_{i+1}) \xrightarrow{\prob} 2\mathbf{F}(v).\]
Since \(\vecnorm{F(v)} = 1\), we get
\begin{align*}
&\prob\left(\mathbf{F}(w_i)+\mathbf{F}(w_{i+1}) = \mathbf{0}\right)\\
&{\leq}\;\prob\left(\vecnorm{\mathbf{F}(w_i)+\mathbf{F}(w_{i+1})} < \vecnorm{\mathbf{F}(v)}\right)\\
&{=}\;\prob\left(2\vecnorm{\mathbf{F}(v)}-\vecnorm{\mathbf{F}(w_i)+\mathbf{F}(w_{i+1})} > \vecnorm{\mathbf{F}(v)}\right)\\
&{\leq}\;\prob\left(\vecnorm{\mathbf{F}(w_i)+\mathbf{F}(w_{i+1})-2\mathbf{F}(v)} > \vecnorm{\mathbf{F}(v)}\right) \to 0.
\end{align*}
In the second term, it is given that \(\vecnorm{\mathbf{F}(w_i)+\mathbf{F}(w_{i+1})} > 0\). Furthermore, \eqref{eq:Fgconv} holds both for \(w_i\) and \(w_{i+1}\), which implies that
\[\mathbf{F}_g(w_i)+\mathbf{F}_g(w_{i+1})-\mathbf{F}(w_i)-\mathbf{F}(w_{i+1}) \xrightarrow{\prob} \mathbf{0}.\]
Thus, the conditional probability
\begin{align*}
\prob&\left(\mathbf{F}_g(w_i)+\mathbf{F}_g(w_{i+1}) = \mathbf{0}\right)\\
\leq\prob&\left(\vecnorm{\mathbf{F}_g(w_i)+\mathbf{F}_g(w_{i+1})} < \frac{\vecnorm{\mathbf{F}(w_i)+\mathbf{F}(w_{i+1})}}{2}\right)\\
\leq\prob&\left(\vphantom{\frac{\vecnorm{\mathbf{F}(w_i)\mathbf{F}(w_{i+1})}}{2}}\vecnorm{\mathbf{F}_g(w_i)+\mathbf{F}_g(w_{i+1})-\mathbf{F}(w_i)-\mathbf{F}(w_{i+1})}\right.\\
&{>}\;\left.\frac{\vecnorm{\mathbf{F}(w_i)+\mathbf{F}(w_{i+1})}}{2}\right) \to 0.
\end{align*}
In the third term, it is given that \(\vecnorm{\mathbf{F}_g(w_i)+\mathbf{F}_g(w_{i+1})} \geq \delta > 0\) for some \(\delta\). As a result, exactly the same steps as in the proof of \eqref{eq:Fgconv} can be used to bound for every \(\epsilon > 0\) the conditional probability
\begin{align*}
&\prob\left(\vecnorm{\frac{\mathbf{F}_g(w_i)+\mathbf{F}_g(w_{i+1})}{\vecnorm{\mathbf{F}_g(w_i)+\mathbf{F}_g(w_{i+1})}}-\frac{\mathbf{F}(w_i)+\mathbf{F}(w_{i+1})}{\vecnorm{\mathbf{F}(w_i)+\mathbf{F}(w_{i+1})}}} > \epsilon\right)\\
&{\leq}\;\prob\left(\frac{2\vecnorm{\mathbf{F}_g(w_i)+\mathbf{F}_g(w_{i+1})-\mathbf{F}(w_i)-\mathbf{F}(w_{i+1})}}{\delta} > \epsilon\right)\\
&{=}\;\prob\left(\vecnorm{\mathbf{F}_g(w_i)+\mathbf{F}_g(w_{i+1})-\mathbf{F}(w_i)-\mathbf{F}(w_{i+1})} > \frac{\epsilon\delta}{2}\right)\\
&{\to}\;0.
\end{align*}
Consequently, all three terms converge to \(0\), which means that their sum also converges to \(0\), leading to \eqref{eq:FgminusFnormmeanconv}.

Afterwards, we define the start and end points of the line segment \(C_l(w_i)\), \(\mathbf{s}_i\) and \(\mathbf{t}_i\), as shown in Fig. \ref{fig:SFvectors}, and prove that
\begin{equation} \label{eq:FnodesminusFsegendsnormmeanconv}
\frac{\mathbf{F}(w_i)+\mathbf{F}(w_{i+1})}{\vecnorm{\mathbf{F}(w_i)+\mathbf{F}(w_{i+1})}}-\frac{\mathbf{F}(\mathbf{s}_i)+\mathbf{F}(\mathbf{t}_i)}{\vecnorm{\mathbf{F}(\mathbf{s}_i)+\mathbf{F}(\mathbf{t}_i)}} \xrightarrow{\prob} \mathbf{0}.
\end{equation}
An intermediate result for the proof of \eqref{eq:FnodesminusFsegendsnormmeanconv} is
\begin{equation} \label{eq:FnodesminusFsegendsconv}
\mathbf{F}(w_i)+\mathbf{F}(w_{i+1})-\mathbf{F}(\mathbf{s}_i)-\mathbf{F}(\mathbf{t}_i) \xrightarrow{\prob} \mathbf{0}.
\end{equation}
To show this, we start from the following application of the law of total probability for every \(\epsilon > 0\):
\begin{align*}
&\prob\left(\vecnorm{\mathbf{F}(w_i)-\mathbf{F}(\mathbf{s}_i)} > \epsilon\right)\\
&{\leq}\;\prob\left(\nabla{u(w_i)} = \mathbf{0}\right)+\prob\left(\nabla{u(\mathbf{s}_i)} = \mathbf{0}\left|\nabla{u(w_i)} \neq \mathbf{0}\right.\right)\\
&{+}\:\prob\left(\vecnorm{\mathbf{F}(w_i)-\mathbf{F}(\mathbf{s}_i)} > \epsilon\left|\nabla{u(w_i)} \neq \mathbf{0}\text{ and }\nabla{u(\mathbf{s}_i)} \neq \mathbf{0}\right.\right).
\end{align*}
It has already been shown that \(\prob\left(\nabla{u(w_i)} = \mathbf{0}\right) \to 0\). For the second term on the right-hand side, \(\mathbf{w}_i-\mathbf{s}_i \xrightarrow{\prob} \mathbf{0}\), since \(\vecnorm{\mathbf{w}_i-\mathbf{s}_i} \leq d(v,w_i)\Delta\phi(w_i) \xrightarrow{\prob} 0\). Taking into account the fact that \(\mathbf{w}_i \xrightarrow{\prob} \mathbf{v}\), we get \(\mathbf{s}_i \xrightarrow{\prob} \mathbf{v}\). The continuity of \(\nabla{u}\) yields \(\nabla{u(\mathbf{s}_i)} \xrightarrow{\prob} \nabla{u(v)}\). Therefore, we can bound the second term as follows:
\begin{align*}
&\prob\left(\nabla{u(\mathbf{s}_i)} = \mathbf{0}\right) \leq \prob\left(\vecnorm{\nabla{u(\mathbf{s}_i)}} < \frac{\vecnorm{\nabla{u(v)}}}{2}\right)\\
&{\leq}\;\prob\left(\vecnorm{\nabla{u(\mathbf{s}_i)-\nabla{u(v)}}} > \frac{\vecnorm{\nabla{u(v)}}}{2}\right) \to 0.
\end{align*}
In the third term, it is given that \(\vecnorm{\nabla{u(w_i)}} \geq \delta > 0\) for some \(\delta\). We thus follow the same steps as in the proof of \eqref{eq:Fgconv} to bound the respective conditional probability:
\begin{align*}
&\prob\left(\vecnorm{\mathbf{F}(w_i)-\mathbf{F}(\mathbf{s}_i)} > \epsilon\right)\\
&{\leq}\;\prob\left(\frac{2\vecnorm{\nabla{u(w_i)-\nabla{u(\mathbf{s}_i)}}}}{\delta} > \epsilon\right)\\
&{=}\;\prob\left(\vecnorm{\nabla{u(w_i)-\nabla{u(\mathbf{s}_i)}}} > \frac{\epsilon\delta}{2}\right) \to 0\;\forall \epsilon > 0,
\end{align*}
where the limit in the last line holds due to the fact that \(\nabla{u(w_i)}-\nabla{u(\mathbf{s}_i)} \xrightarrow{\prob} \mathbf{0}\). As a result, it holds that \(\mathbf{F}(w_i)-\mathbf{F}(\mathbf{s}_i) \xrightarrow{\prob} \mathbf{0}\) and in a similar fashion \(\mathbf{F}(w_{i+1})-\mathbf{F}(\mathbf{t}_i) \xrightarrow{\prob} \mathbf{0}\). The sum property of convergence in probability directly implies \eqref{eq:FnodesminusFsegendsconv} from the above two results.

To prove \eqref{eq:FnodesminusFsegendsnormmeanconv}, we use the law of total probability to write
\begin{align}
&\prob\left(\vecnorm{\frac{\mathbf{F}(w_i)+\mathbf{F}(w_{i+1})}{\vecnorm{\mathbf{F}(w_i)+\mathbf{F}(w_{i+1})}}-\frac{\mathbf{F}(\mathbf{s}_i)+\mathbf{F}(\mathbf{t}_i)}{\vecnorm{\mathbf{F}(\mathbf{s}_i)+\mathbf{F}(\mathbf{t}_i)}}} > \epsilon\right)\nonumber\\
&{\leq}\;\prob\left(\mathbf{F}(w_i)+\mathbf{F}(w_{i+1}) = \mathbf{0}\right)\nonumber\\
&{+}\:\prob\left(\mathbf{F}(\mathbf{s}_i)+\mathbf{F}(\mathbf{t}_i) = \mathbf{0}\left|\mathbf{F}(w_i)+\mathbf{F}(w_{i+1}) \neq \mathbf{0}\right.\right)\nonumber\\
&{+}\:\prob\left(\vecnorm{\frac{\mathbf{F}(w_i)+\mathbf{F}(w_{i+1})}{\vecnorm{\mathbf{F}(w_i)+\mathbf{F}(w_{i+1})}}-\frac{\mathbf{F}(\mathbf{s}_i)+\mathbf{F}(\mathbf{t}_i)}{\vecnorm{\mathbf{F}(\mathbf{s}_i)+\mathbf{F}(\mathbf{t}_i)}}} > \epsilon\right.\nonumber\\
&\left.\left|\vphantom{\vecnorm{\frac{\mathbf{F}(w_i)+\mathbf{F}(w_{i+1})}{\vecnorm{\mathbf{F}(w_i)+\mathbf{F}(w_{i+1})}}-\frac{\mathbf{F}(\mathbf{s}_i)+\mathbf{F}(\mathbf{t}_i)}{\vecnorm{\mathbf{F}(\mathbf{s}_i)+\mathbf{F}(\mathbf{t}_i)}}}}\mathbf{F}(w_i)+\mathbf{F}(w_{i+1}) \neq \mathbf{0}\text{ and }\mathbf{F}(\mathbf{s}_i)+\mathbf{F}(\mathbf{t}_i) \neq \mathbf{0}\right.\right).\label{eq:ProbFnodesminusFsegendsineq}
\end{align}
We have already proved
\[\prob\left(\mathbf{F}(w_i)+\mathbf{F}(w_{i+1}) = \mathbf{0}\right) \to 0.\]
For the second term on the right-hand side, it is given that \(\vecnorm{\mathbf{F}(w_i)+\mathbf{F}(w_{i+1})} > 0\). We bound the conditional probability
\begin{align*}
\prob&\left(\mathbf{F}(\mathbf{s}_i)+\mathbf{F}(\mathbf{t}_i) = \mathbf{0}\right)\\
\leq\prob&\left(\vecnorm{\mathbf{F}(\mathbf{s}_i)+\mathbf{F}(\mathbf{t}_i)} < \frac{\vecnorm{\mathbf{F}(w_i)+\mathbf{F}(w_{i+1})}}{2}\right)\\
\leq\prob&\left(\vphantom{\frac{\vecnorm{\mathbf{F}(w_i)+\mathbf{F}(w_{i+1})}}{2}}\vecnorm{\mathbf{F}(w_i)+\mathbf{F}(w_{i+1})-\mathbf{F}(\mathbf{s}_i)-\mathbf{F}(\mathbf{t}_i)}\right.\\
&{>}\;\left.\frac{\vecnorm{\mathbf{F}(w_i)+\mathbf{F}(w_{i+1})}}{2}\right) \to 0,
\end{align*}
where the last limit is due to \eqref{eq:FnodesminusFsegendsconv}. Last, we focus on the third term, where it is given that \(\vecnorm{\mathbf{F}(w_i)+\mathbf{F}(w_{i+1})} \geq \delta > 0\) for some \(\delta\). Taking the same steps as in the proof of \eqref{eq:Fgconv} to bound the third term, we obtain
\begin{align*}
&\prob\left(\vecnorm{\frac{\mathbf{F}(w_i)+\mathbf{F}(w_{i+1})}{\vecnorm{\mathbf{F}(w_i)+\mathbf{F}(w_{i+1})}}-\frac{\mathbf{F}(\mathbf{s}_i)+\mathbf{F}(\mathbf{t}_i)}{\vecnorm{\mathbf{F}(\mathbf{s}_i)+\mathbf{F}(\mathbf{t}_i)}}} > \epsilon\right)\\
&{\leq}\;\prob\left(\frac{2\vecnorm{\mathbf{F}(w_i)+\mathbf{F}(w_{i+1})-\mathbf{F}(\mathbf{s}_i)-\mathbf{F}(\mathbf{t}_i)}}{\delta} > \epsilon\right)\\
&{=}\;\prob\left(\vecnorm{\mathbf{F}(w_i)+\mathbf{F}(w_{i+1})-\mathbf{F}(\mathbf{s}_i)-\mathbf{F}(\mathbf{t}_i)} > \frac{\epsilon\delta}{2}\right)\\
&{\to}\;0\;\forall \epsilon > 0.
\end{align*}
The sum of all three terms on the right-hand side of \eqref{eq:ProbFnodesminusFsegendsineq} converges to \(0\) as each of them converges to \(0\), which implies \eqref{eq:FnodesminusFsegendsnormmeanconv}.

For the line segment \(C_l(w_i)\), we define the event
\[B = \exists \mathbf{r} \in C_l(w_i):\;\mathbf{F}(\mathbf{r}) = \frac{\mathbf{F}(\mathbf{s}_i)+\mathbf{F}(\mathbf{t}_i)}{\vecnorm{\mathbf{F}(\mathbf{s}_i)+\mathbf{F}(\mathbf{t}_i)}}\]
and show that
\begin{equation} \label{eq:Probintermvaluesegendsconv}
\prob\left(B\right) \to 1.
\end{equation}
Let us consider the complement of \(B\). The law of total probability implies that
\begin{align*}
\prob\left(B^c\right) \leq &\prob\left(\mathbf{F}\text{ not continuous at some point on }C_l(w_i)\right)\\
&{+}\:\prob\left(\left.B^c\right|\mathbf{F}\text{ continuous on }C_l(w_i)\right).
\end{align*}
Since \(C_l(w_i) \xrightarrow{\prob} \{\mathbf{v}\}\) and \(\mathbf{F}\) is continuous at \(\mathbf{v}\), it holds for the first term on the right-hand side that
\[\prob\left(\mathbf{F}\text{ not continuous at some point on }C_l(w_i)\right) \to 0.\]
In the second term, it is given that \(\mathbf{F}\) is continuous on \(C_l(w_i)\). We note that \(\mathbf{F}\) is a circular quantity, since \(\vecnorm{\mathbf{F}} = 1\). Therefore, if we define the event
\[D = \exists \mathbf{q} \in C_l(w_i):\;\mathbf{F}(\mathbf{q}) = -\frac{\mathbf{F}(\mathbf{s}_i)+\mathbf{F}(\mathbf{t}_i)}{\vecnorm{\mathbf{F}(\mathbf{s}_i)+\mathbf{F}(\mathbf{t}_i)}},\]
the intermediate value theorem for \(\mathbf{F}\) on \(C_l(w_i)\) implies that \(B \cup D = \Omega\). It is straightforward that
\[\vecnorm{-\frac{\mathbf{F}(\mathbf{s}_i)+\mathbf{F}(\mathbf{t}_i)}{\vecnorm{\mathbf{F}(\mathbf{s}_i)+\mathbf{F}(\mathbf{t}_i)}}-\mathbf{F}(\mathbf{s}_i)} \geq \sqrt{2}.\]
However, since it holds for every \(\mathbf{p} \in C_l(w_i)\) that \(\vecnorm{\mathbf{p}-\mathbf{v}} \leq \rho(n)\), we have \(\mathbf{p} \xrightarrow{\prob} \mathbf{v}\). It was previously shown that \(\mathbf{s}_i \xrightarrow{\prob} \mathbf{v}\), so the continuity of \(\mathbf{F}\) at \(\mathbf{v}\) yields \(\mathbf{F}(\mathbf{p}) \xrightarrow{\prob} \mathbf{F}(v)\) and \(\mathbf{F}(\mathbf{s}_i) \xrightarrow{\prob} \mathbf{F}(v)\). It follows that
\[\prob\left(\vecnorm{\mathbf{F}(\mathbf{p})-\mathbf{F}(\mathbf{s}_i)} \geq \epsilon\right) \to 0\;\forall \mathbf{p} \in C_l(w_i)\;\forall \epsilon > 0.\]
If we consider the point \(\mathbf{q}\) which appears in the expression of event \(D\), we obtain
\[\prob(D) \leq \prob\left(\exists \mathbf{q} \in C_l(w_i):\;\vecnorm{\mathbf{F}(\mathbf{q})-\mathbf{F}(\mathbf{s}_i)} \geq \sqrt{2}\right) \to 0.\]
As a result, when it is given that \(\mathbf{F}\) is continuous on \(C_l(w_i)\), it holds that
\[\prob\left(B^c\right) = 1 - \prob(B) \leq \prob(D) \to 0.\]
Since both terms of the sum that bounds \(\prob\left(B^c\right)\) converge to \(0\), so does \(\prob\left(B^c\right)\), leading to \eqref{eq:Probintermvaluesegendsconv}. Thus, there exists a sequence of random variables \(\left\{\mathbf{r}_n\right\}\) such that every term of the sequence belongs to \({\left(C_l(w_i)\right)}_n\) and
\begin{equation} \label{eq:Fsegendsnormmeanminusintermvalueconv}
\frac{\mathbf{F}(\mathbf{s}_i)+\mathbf{F}(\mathbf{t}_i)}{\vecnorm{\mathbf{F}(\mathbf{s}_i)+\mathbf{F}(\mathbf{t}_i)}}-\mathbf{F}(\mathbf{r}_n) \xrightarrow{\prob} \mathbf{0}.
\end{equation}

For simplicity, we omit the index \(n\) from \(\mathbf{r}_n\) and write \(\mathbf{r}\) instead, as we have done for all the rest of the sequences of random variables that are treated in the proof. For this \(\mathbf{r}\), we combine \eqref{eq:FgminusFnormmeanconv}, \eqref{eq:FnodesminusFsegendsnormmeanconv} and \eqref{eq:Fsegendsnormmeanminusintermvalueconv} through summation into
\begin{equation} \label{eq:FgnormmeanminusFintermvalueconv}
\frac{\mathbf{F}_g(w_i)+\mathbf{F}_g(w_{i+1})}{\vecnorm{\mathbf{F}_g(w_i)+\mathbf{F}_g(w_{i+1})}}-\mathbf{F}(\mathbf{r}) \xrightarrow{\prob} \mathbf{0}.
\end{equation}
Taking into account the closed form \eqref{eq:Il} for \(I_l(w_i)\) and the product and sum properties of convergence in probability, we get
\begin{equation} \label{eq:FgnormmeanminusFintermvalueintconv}
\int_{C_l(w_i)} \left(\frac{\mathbf{F}_g(w_i)+\mathbf{F}_g(w_{i+1})}{\vecnorm{\mathbf{F}_g(w_i)+\mathbf{F}_g(w_{i+1})}}-\mathbf{F}(\mathbf{r})\right)\cdot\mathbf{n}\,d\ell \xrightarrow{\prob} 0.
\end{equation}

Additionally, the law of total probability yields for every \(\epsilon > 0\)
\begin{align*}
&\prob\left(\left|\int\limits_{C_l(w_i)} \left(\mathbf{F}(\mathbf{r})-\mathbf{F}\right)\cdot\mathbf{n}\,d\ell\right| > \epsilon\right)\\
&{\leq}\;\prob\left(\mathbf{F}\text{ not continuous at some point on }C_l(w_i)\right)\\
&{+}\:\prob\left(\left|\int\limits_{C_l(w_i)} \left(\mathbf{F}(\mathbf{r})-\mathbf{F}\right)\cdot\mathbf{n}\,d\ell\right| > \epsilon|\mathbf{F}\text{ cont. on }C_l(w_i)\right).
\end{align*}
It has been shown that the first term on the right-hand side converges to \(0\). The second term, which is conditioned on \(\mathbf{F}\) being continuous on \(C_l(w_i)\), also converges to \(0\), since the length of \(C_l(w_i)\), which is equal to \(|d(v,w_{i+1})-d(v,w_i)|\), converges to \(0\) almost surely. Consequently, we obtain
\begin{equation} \label{eq:FintermvalueminusFintconv}
\int_{C_l(w_i)} \left(\mathbf{F}(\mathbf{r})-\mathbf{F}\right)\cdot\mathbf{n}\,d\ell \xrightarrow{\prob} 0.
\end{equation}
The summation of \eqref{eq:FgnormmeanminusFintermvalueintconv} and \eqref{eq:FintermvalueminusFintconv} leads to
\begin{equation} \label{eq:FgnormmeanminusFintconv}
I_l(w_i)-\int_{C_l(w_i)}\mathbf{F}\cdot\mathbf{n}\,d\ell \xrightarrow{\prob} 0.
\end{equation}

Due to the fact that \eqref{eq:FgofwiminusFintconv} and \eqref{eq:FgnormmeanminusFintconv} hold for every \(i \in \{1,\,\dots,\,N\}\), the sum property of convergence in probability implies \eqref{eq:curvgeomapproxconvnumer}.

Finally, our construction of region \(S(v)\) guarantees that
\[S(v) \xrightarrow{\text{a.s.}} \{\mathbf{v}\}\]
because \(\rho(n) \in o\left(1\right)\). Thus, it holds
\[\frac{\oint\limits_{\Gamma(S(v))}^{} \mathbf{F}\cdot\mathbf{n}\,d\ell}{\left|S(v)\right|} \xrightarrow{\text{a.s.}} \kappa(v).\]
The last result, together with \eqref{eq:curvgeomapproxconvnumer}, implies that our approximation \eqref{eq:curvgeomapprox} converges in probability to the true value of the curvature, after making use of the triangle inequality.
\end{IEEEproof}

\bibliographystyle{IEEEtran}
\bibliography{IEEEabrv,SDM16}

\end{document}